\documentclass[10pt,twocolumn,letterpaper]{article}

\usepackage{cvpr}              
\usepackage[dvipsnames]{xcolor}

\definecolor{cvprblue}{rgb}{0.21,0.49,0.74}
\usepackage[pagebackref,breaklinks,colorlinks,citecolor=cvprblue]{hyperref}
\usepackage{url}

\usepackage[utf8]{inputenc} 
\usepackage[T1]{fontenc}    
\usepackage{booktabs}       
\usepackage{amsfonts}       
\usepackage{nicefrac}       
\usepackage{microtype}      
\usepackage{xcolor}         
\usepackage{amsthm}
\usepackage{amsmath}
\usepackage{amssymb}
\usepackage{dsfont}
\usepackage{amsthm}
\usepackage{thm-restate}
\usepackage{graphicx}
\usepackage{blindtext}
\usepackage{multirow}
\usepackage{gensymb}
\usepackage{float}
\usepackage{caption}
\usepackage{amssymb}
\usepackage{pifont}
\usepackage{hyperref}
\usepackage{textcomp}

\newcommand{\xmark}{\ding{55}}%
\newcommand{\dc}[1]{\multicolumn{2}{c|}{#1}}

\def\paperID{2714} 
\def\confName{CVPR}
\def\confYear{2024}

\title{
Outdoor Scene Extrapolation with \\
Hierarchical Generative Cellular Automata
\vspace{-0.1cm}
}

\author{
Dongsu Zhang\textsuperscript{1\;*}
\quad\quad
Francis Williams\textsuperscript{2\;}
\quad\quad
Zan Gojcic\textsuperscript{2\;}
\quad\quad
Karsten Kreis\textsuperscript{2\;}
\vspace{0.15cm}
\\
Sanja Fidler\textsuperscript{2,3,4}
\quad\quad
Young Min Kim\textsuperscript{1}
\quad\quad
Amlan Kar\textsuperscript{2, 3, 4}
\vspace{0.1cm}
\\
{\small\textsuperscript{1}Seoul National University \quad \textsuperscript{2}NVIDIA \quad \textsuperscript{3}Vector Institute \quad \textsuperscript{4}University of Toronto\vspace{0.15cm}}
}

\begin{document}

\twocolumn[{
\renewcommand\twocolumn[1][]{#1}
\maketitle
\begin{center}
\vspace{-3em}
    \centering
        \captionsetup{type=figure}
        \includegraphics[width=0.93\linewidth]{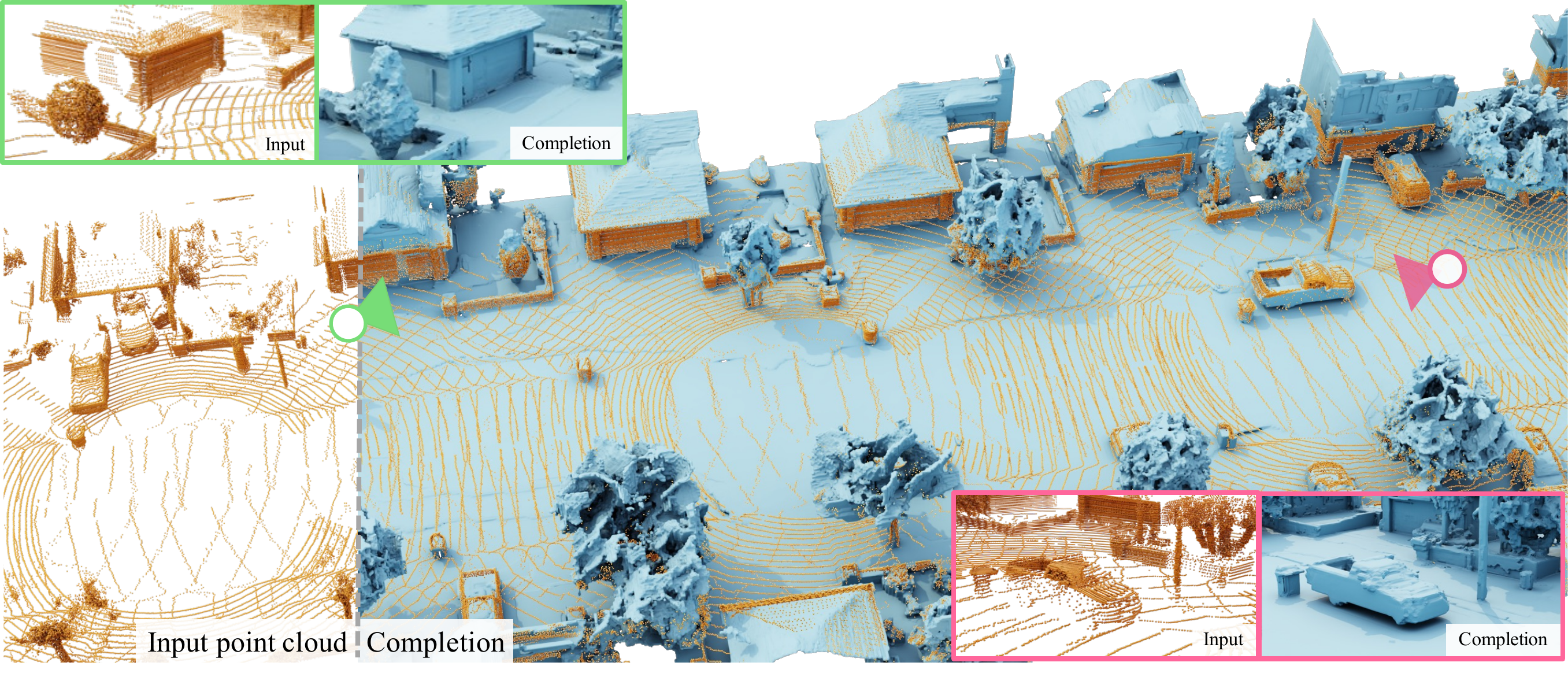}
        \vspace{-1em}
        \captionof{figure}{
            Geometry generation from hGCA (blue) from five accumulated LiDAR scans (yellow spheres) on real-world Waymo-open dataset. 
            hGCA is a conditional 3D generative model that can generate geometry beyond occlusions (vehicles, facades) and input field of view (roofs, trees, poles), from sparse and noisy LiDAR scans.
            Our method is also spatially scalable, completing this whole scene (120 meters) at high resolution on a single 24GB GPU without additional tricks.
            \vspace{-0.5em}
        }
        \label{fig:teaser}
\end{center}
}]

\setlength{\abovedisplayskip}{2pt}
\setlength{\belowdisplayskip}{3pt}

\begin{abstract}
\vspace{-2em}
We aim to generate fine-grained 3D geometry from large-scale sparse LiDAR scans, abundantly captured by autonomous vehicles (AV). 
Contrary to prior work on AV scene completion, we aim to extrapolate fine geometry from unlabeled and beyond spatial limits of LiDAR scans, taking a step towards generating realistic, high-resolution simulation-ready 3D street environments. 
We propose hierarchical Generative Cellular Automata (hGCA), a spatially scalable conditional 3D generative model, which grows geometry recursively with local kernels following~\cite{zhang2021learning, zhang2022probabilistic}, in a coarse-to-fine manner, equipped with a light-weight planner to induce global consistency.
Experiments on synthetic scenes show that hGCA generates plausible scene geometry with higher fidelity and completeness compared to state-of-the-art baselines.
Our model generalizes strongly from sim-to-real, qualitatively outperforming baselines on the Waymo-open dataset. 
We also show anecdotal evidence of the ability to create novel objects from real-world geometric cues even when trained on limited synthetic content.
More results and details can be found on our \href{https://research.nvidia.com/labs/toronto-ai/hGCA/}{project page}.

\let\thefootnote\relax\footnote{\textsuperscript{*} Work started during Dongsu's internship at NVIDIA}
\end{abstract}

\vspace{-3em}
\section{Introduction}
How can we scalably build large-scale, diverse and realistic digital worlds for applications in simulation for autonomous vehicles (AV) or gaming and entertainment? 
Manually authoring a realistic scene requires significant effort in creating individual objects and positioning them in realistic spatial configurations. 
Procedural models are a promising alternative which back recent AAA games such as No Man's Sky.
However, authoring procedural models of objects and environments are usually time consuming manual tasks.
Densely scanning the world is now an increasingly popular and more scalable option, using Neural Radiance Field (NeRF) based approaches.
However, these reconstruction methods typically don't capture content beyond what is observed. 
Sparse LiDAR scans from autonomous vehicles -- a by-product of their development and deployment -- also provide cues to the geometry of street environments in the world. 
Our work aims to use these sparse LiDAR scans as input to a conditional 3D generative model that learns to extrapolate plausible high-resolution scene geometry. 

Prior work in the domain has focused on semantic scene completion (SSC) from a single LiDAR scan~\cite{scpnet, JS3CNet}, using accumulated sequential LiDAR scans with labeled semantic classes as supervision~\cite{SemanticKITTI}. 
This is useful for AV perception to learn to expect 3D semantic occupancy beyond instantaneous observations. 
However, using such accumulated scans as supervision typically results in outputs unsuitable for simulation, since they have low-resolution geometry and suffer from heavy occlusions, exacerbated by scans being taken from a single drive through a dynamic scene~\cite{SemanticKITTI, Waymo-Open}. 
Moreover, typical LiDAR scanners in AV have a restricted height range which prohibits learning to generate scene geometry beyond this limit in SSC. 
From sparse LiDAR scans, we instead aim to generate high-resolution scene geometry and go beyond the LiDAR range (Fig.~\ref{fig:teaser}),  to take a step towards simulation ready scene geometry. 
To differentiate from the task of semantic scene completion (SSC), we name our task outdoor scene extrapolation. However, for ease of expression, we use the terms completion, generation, and extrapolation interchangeably through the rest of the paper. 
We train and evaluate on synthetic scenes which allow fine and complete geometric supervision while maintaining the ability to complete geometry from real LiDAR scans. 
We use a conditional 3D generative model, which is more suited to this challenging inverse problem, as opposed to prior SSC methods that typically use discriminative autoencoder.

We propose a spatially scalable 3D generative model of geometry, with a two-stage hierarchical coarse-to-fine formulation, called hGCA. 
hGCA builds on top of the recent Generative Cellular Automata (GCA) framework~\cite{zhang2021learning, zhang2022probabilistic}, which is a 3D generative model that recursively applies local kernels to incrementally grow geometry from a sparse set of active cells.
GCA was shown to perform competitively with state-of-the-art for geometry completion from dense indoor scans. The sparsity and locality of GCA allows spatial scalability. However, we find that naively applying GCA for fine geometry extrapolation on large outdoor scenes from sparse LiDAR leads to performance deterioration stemming from a lack of global context and the need to use a large number of recursive steps, the latter motivating our coarse-to-fine approach. 
To introduce global context, hGCA's coarse stage uses a GCA conditioned on features from a light-weight bird's eye view \textit{planner} to generate scene geometry in a low-resolution voxel grid, without losing spatial scalability.
The second stage synthesizes finer details with cGCA~\cite{zhang2022probabilistic}, generating high resolution voxels augmented with local implicit functions that allow promoting the output to a 3D mesh.

We train on synthetic street scenes, using data from the CARLA simulator~\cite{carla}, and a city asset from Turbosquid, using simulated LiDAR scans as input. On synthetic scenes, hGCA outperforms state-of-the-art SSC and indoor scene completion methods on multiple metrics for geometry extrapolation. 
Quantitatively evaluating 3D generative models in the real world is challenging. Qualitatively, we observe that hGCA shows strong sim-to-real generalization to real LiDAR scans compared to prior work, generating more complete and higher fidelity geometry, demonstrated on the Waymo-open dataset~\cite{Waymo-Open}. We also demonstrate with examples that despite being trained on limited synthetic content, hGCA can generate some novel content beyond its training data, by taking geometric cues from input LiDAR scans.
\begin{figure*}[t!]
    \vspace{-2em}
    \centering
    \includegraphics[width=0.85\textwidth]{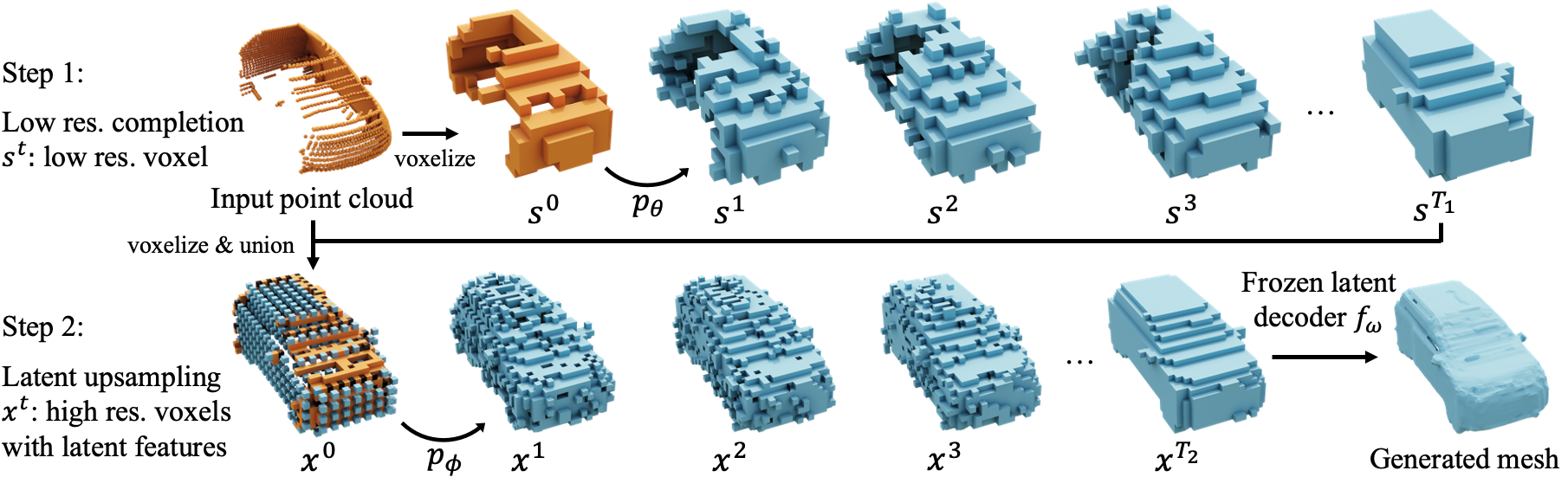}
    \vspace{-1em}
    \caption{
        Overview of our method.
        Given several LiDAR scans, our method generates low resolution completion $s^{T_1}$ using a GCA attached with a planner that adds global consistency.
        Then given $s^{T_1}$ and the input, we upsample the completion using a cGCA into high resolution voxel with a local latent $x^{T_2}$ and decode it to obtain the final generated mesh.
        \vspace{-2em}
        }
        \label{fig:overview}
\end{figure*}

\section{Related Work}
\label{sec:related_works}
\textbf{3D Shape Completion.}
Earlier works~\cite{dai2017complete, yuan2018pcn} on data-driven 3D shape completion regressed a single shape from partial 3D observation using deep neural networks.~\cite{gu2020weakly} learn to complete 3D shapes using partial geometry supervision coming from LiDAR scans. Multiple works~\cite{song2016sscnet, dai2018scancomplete, dai2020sgnn} tackle completion of indoor scenes from dense RGB-D scans.~\cite{dai2018scancomplete, dai2020sgnn} proposed a hierarchical coarse-to-fine approach for fine-grained completion of indoor scenes. Recent works~\cite{chabra2020deep_local_shapes, chiyu2020local, Peng2020ECCV} have employed deep implicit fields to learn to generate continuous surfaces~\cite{Park_2019_CVPR, Occupancy_Networks,chen2018implicit_decoder, Michalkiewicz2019deep_level_sets}. We take inspiration from 
these using a coarse-to-fine approach and local implicit functions
at the finest level.
Most prior works on outdoor scene completion focus on semantic scene completion (SSC)~\cite{SemanticKITTI, Pan2020SemanticPOSSAP, cheng2021s3cnet, JS3CNet, scpnet} for autonomous vehicles (AV), i.e. completing semantic voxel occupancy given a single LiDAR scan using accumulated sequential point clouds with semantic labels for supervision.
JS3CNet~\cite{JS3CNet} proposes a novel point-voxel interaction module for better feature extraction and SCPNet~\cite{scpnet} utilizes student-teacher distillation from a multi-frame input teacher, and improves network design without any dowsampling modules. While the works show suitable results for AV perception, the methods produce low resolution geometry and suffer from occlusions arising from supervision, deficient for simulation. We show superior performance to both indoor scene completion and SSC methods for AV scene geometry extrapolation on synthetic data, as well as qualitatively improved sim-to-real generalization on the Waymo-open~\cite{Waymo-Open} dataset.

\textbf{3D Generative models}~\cite{wu2016learning,henzler2019platonicgan,gao2022get3d,zeng2022lion,yang2019pointflow, cai2020shapegf}, have typically focused on synthesizing single objects, leveraging GANs~\cite{wu2020msc, arora2021}, and diffusion based generative models~\cite{Zhou2021ICCV, zeng2022lion}. Recent methods in text-to-3D generative models~\cite{poole2022dreamfusion}, have shown impressive results in generating novel shapes and small scenes.~\cite{zyrianov2022lidargen, xiong2023learning} learn generative models of LiDAR scans demonstrating scene level point cloud synthesis in autonomous driving. Inspired by the cellular automaton~\cite{von1966theory}, the GCA~\cite{zhang2021learning, zhang2022probabilistic} framework can generate multimodal completions for both objects and indoor scenes. GCA scales to scenes as it recursively applies local kernels to grow a sparse set of active cells in its generative process. This resembles diffusion models~\cite{sohl2015nonequil, bordes2017infusion,ho2020denoising} where samples are generated with a recursive learned denoising kernel, and Neural Cellular Automata~\cite{mordvintsev2020growing}. We find that the locality of GCA fails to capture global context, generating artifacts in large scenes. hGCA extends GCA to capture global context and efficiently generate fine geometry.

\section{Hierarchical Generative Cellular Automata} \label{sec:hgca}
Given LiDAR scans captured from an ego-vehicle, the task is to generate complete scene geometry, including regions beyond the LiDAR range or away from the street.
To efficiently handle expansive scales of outdoor scenes with fine detail (Fig.~\ref{fig:teaser}), hierarchical Generative Cellular Automata (hGCA) proposes a conditional generative model in a two-step, coarse-to-fine manner as shown in Fig.~\ref{fig:overview}. 
The first step of hGCA extrapolates the scene in a low resolution voxel representation using a model based on Generative Cellular Automata (GCA)~\cite{zhang2021learning}; a sparse, local and hence spatially scalable generative model, which we briefly introduce in Sec.~\ref{sec:gca} for completeness.
However, the local generation of GCA can introduce artifacts in extrapolating larges scenes beyond sensor measurements.
We propose to induce global context into GCA by jointly training a light-weight bird's eye view encoder, called \emph{planner} (Sec.~\ref{sec:planner}).
Then we transform the coarse geometry into high-resolution continuous scene geometry using local implicit functions~\cite{zhang2022probabilistic} (Sec.~\ref{sec:upsampling}).
Together, the proposed method can create large outdoor scenes with spatial scalability, global consistency, higher fidelity from sparse, partial real-world scans.

\subsection{Background: Generative Cellular Automata}\label{sec:gca}
\textbf{Generative process.} 
GCA recursively grows an incomplete shape to completion as illustrated in step 1 of Fig.~\ref{fig:overview}, by locally updating occupancies around the current shape. 
GCA represents shapes as sparse voxel  occupancies, $s=\{(c, o_c)|c \in \mathbb{Z}^3, o_c \in \{0, 1\} \}$, where $o_c$ indicates binary occupancy of a voxel / cell with its coordinates $c$. In the following text, we use voxel and cell interchangeably.
Given an observed, incomplete state $s^0$, it generates a completed state $s^T$ by recursively sampling $s^{1:T}$:
\begin{equation}
    s^{t + 1} \sim p_\theta(\cdot|s^t), \label{eq:transition_kernel}
\end{equation}
where $T$ is a predefined number of transition steps and $p_\theta$ is a local transition kernel with parameters $\theta$.
The transition kernel uses a U-Net~\cite{ronneberger2015unet} architecture using sparse convolutions~\cite{graham18} \ie, the convolution only processes occupied cells for efficiency.
The transition kernel $p_\theta$ is computed \emph{locally} on the neighborhood of the occupied cells, $\mathcal{N}(s^t)=\{ c' \in \mathbb{Z}^3 \mid  d(c, c') \leq r, o_c = 1, c\in \mathbb{Z}^3 \}$, \ie, cells within a radius $r$ from current occupied cells under a distance metric $d$.
For efficient sampling, the transition kernel is computed for each cell in $\mathcal{N}(s^t)$ independently,
\begin{equation}
    p(s^{t + 1} | s^{t}) 
    = \prod_{c \in \mathcal{N}(s^t)} p_\theta(o_c| s^t), \label{eq:transition_kernel_factorization}
\end{equation}
\begin{equation}
    p_\theta(o_c|s^t) = Ber(\lambda_{\theta, c}), \label{eq:transition_kernel_factorized}
\end{equation}
where $p_\theta(o_c| s^t)$ is a Bernoulli variable with mean $\lambda_{\theta, c}$ estimated by the neural network for cell $c$, given $s_t$.
This sparse and local generative process of GCA allows more spatial scalability over traditional encoder-decoder methods that process whole scenes at once. 
In this work, we use a variant of GCA where the transition kernel $p_\theta$ is conditioned on both the initial state $s^0$ and the current state $s^t$, improving conditioning on the input $s^0$~\cite{zhang2022probabilistic}.

\textbf{Training GCA.} 
GCA is trained with infusion training~\cite{bordes2017infusion} to converge to a desired shape $s^{\text{gt}}$, given $s^0$ and $s^{\text{gt}}$.
Infusion training supervises the transition kernel $p_\theta(s^{t+1}|s^t)$ at each step, by defining an infusion kernel
\begin{equation}
    q^t_\theta(\tilde{s}^{t + 1}| \tilde{s}^t, s^{\text{gt}}) = \prod_{c \in \mathcal{N}(\tilde{s}^t)} q^t_\theta(o_c| \tilde{s}^t, s^{\text{gt}})
\end{equation}
where the infusion kernel $q^t_\theta(\tilde{s}^{t + 1}| \tilde{s}^t, s^{\text{gt}})$ is factorized per cell as in Eq.~\ref{eq:transition_kernel_factorization}.
The infusion kernel for a single cell $c$,
\begin{equation}
 q^t_\theta(o_c|\tilde{s}^t, s^{\text{gt}}) = Ber( (1 - \alpha^t) \lambda_{\theta, c} + \alpha^t \mathds{1}[c \in s^{\text{gt}}])
\end{equation}
is a Bernoulli variable with its mean defined as the estimated occupancy probability $\lambda_{\theta,c}$ \emph{infused} with target $s^{\text{gt}}$ with weight $\alpha^t = \alpha_1 t + \alpha_0 \mid \alpha^t \in [0,1]$, which increases linearly with $t$.

For training input $s^0$ to generate $s^{\text{gt}} \sim s^T$, intermediate infusion states $\tilde{s}^{1:T}$ are sampled from the infusion kernel $q_\theta(\tilde{s}^{t+ 1}|\tilde{s}^t, s^{\text{gt}})$ recursively.
For each sampled infusion state $\tilde{s}^t$, GCA is trained with binary cross entropy loss against the ground-truth $s^{\text{gt}}$ on the neighborhood $\mathcal{N}(\tilde{s}^t)$ by minimizing: 
\begin{equation}
    \mathcal{L_{\text{GCA}}} = -\sum_{c \in \mathcal{N}(\tilde{s}^t)} \sum_{o_c \in \{0, 1\}} \mathds{1}[o_c = o_{c,s^{\text{gt}}}] \log p_\theta(o_c|\tilde{s}^t), \label{eq:gca_loss}
\end{equation}
where $o_{c, s^{\text{gt}}} \in \{0, 1\}$ is the occupancy of cell $c$ for ground truth shape $s^{\text{gt}}$. 
We refer to~\cite{zhang2022probabilistic} for theoretical foundation of the loss function.

\subsection{Planner} \label{sec:planner}
\begin{figure}[t]
    \vspace{-1em}
    \centering
    \includegraphics[width=0.9\linewidth]{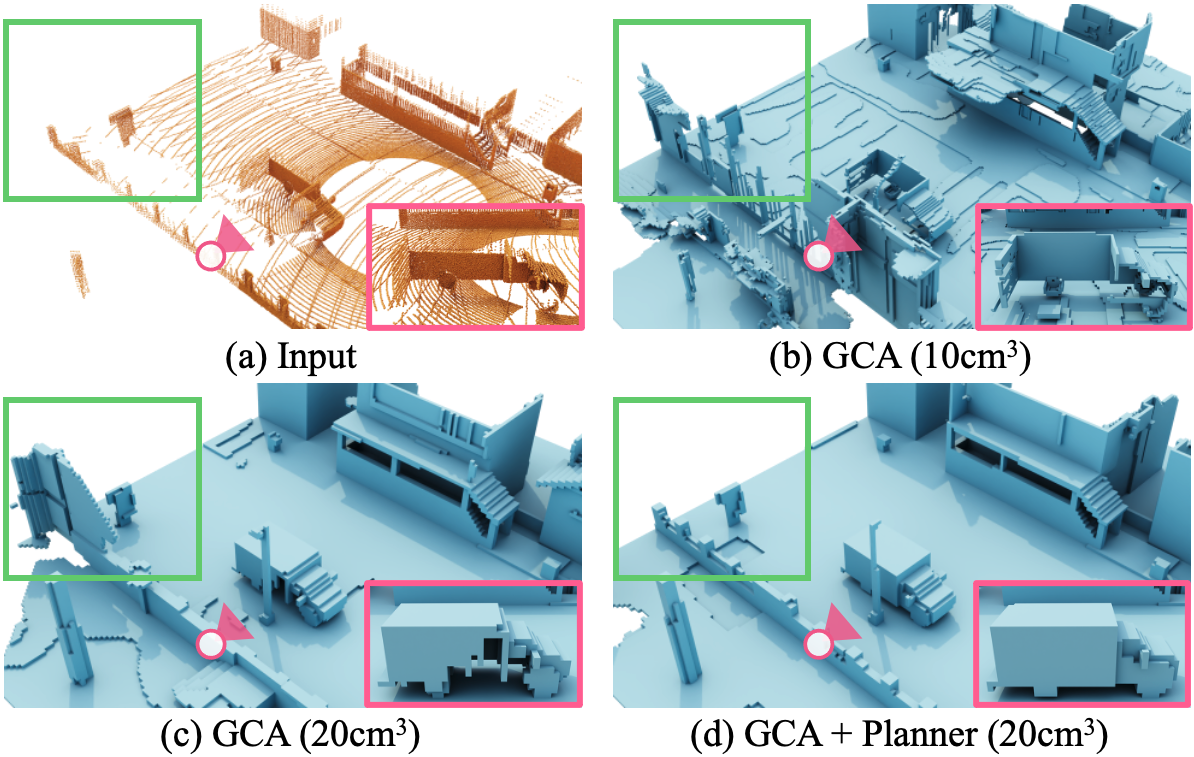}
    \vspace{-1em}
    \caption{
        Left: (a) Input LiDAR scans. 
        (b), (c) GCA completion in $\text{10cm}^3$ and $\text{20cm}^3$ voxel resolution. 
        (d) GCA + planner completion in 20cm voxel resolution.
        GCA is local and often cannot capture the global context, generating imperfect completions (pink box) or artifacts (green box).
        \vspace{-1em}
        }
        \label{fig:planner_example}
\end{figure}

While GCA has been shown to complete small indoor scenes, we find that it lacks global consistency on extrapolating large-scale scenes.
Take the fence in green box in left of Fig.~\ref{fig:planner_example} for instance.
The walls of buildings are generated inconsistently by GCA, at both low and high resolution, showing symptoms of lack of global consistency. 
This issue is exacerbated in sim-to-real inference shown in Fig.~\ref{fig:qual_real}.
We hypothesize that while GCA's sparse and recursive kernel brings scalability, it cannot maintain global context both spatially and temporally. Spatially, the sparse convolutions deliver information only through occupied cells, making it difficult to observe wide spatial context without immediate connection. Moreover, the Markov transition kernel transmits no other memory except binary occupancy between transitions, thus inhibiting long-range "planning" across transition steps.

Hence, we introduce a light-weight \emph{planner} module that provides the global context of the scene into GCA, while it maintains the recursive local operations.
Specifically, we provide the consistent bird's eye view (BEV) features to GCA kernels, independent of time step $t$.
The features are trained to plan ahead and predict very low-resolution, yet dense, final occupancy from the initial state $s^0$.

\begin{figure}[t]
    \vspace{-1em}
    \centering
    \includegraphics[width=0.9\linewidth]{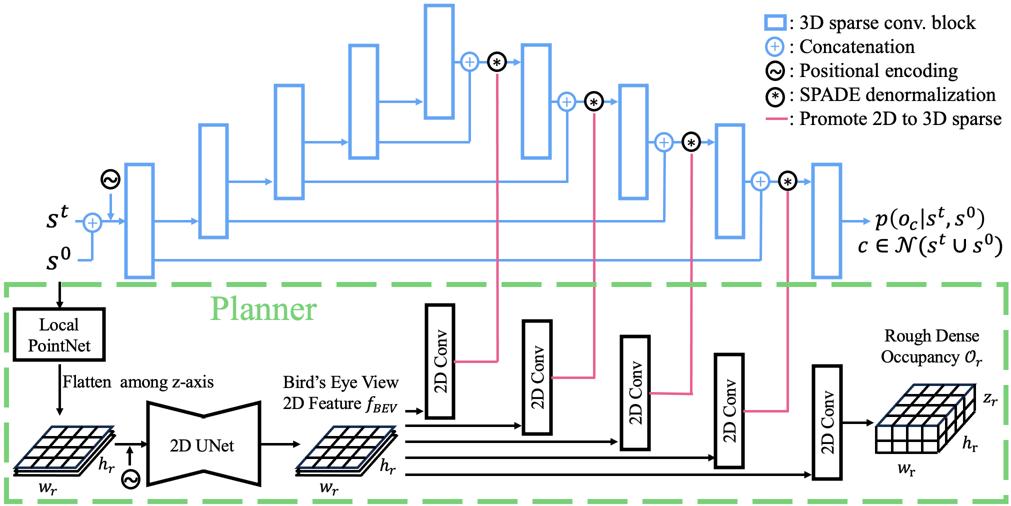}
    \vspace{-1em}
    \caption{
        Illustration of GCA attached with planner module.
        }
        \label{fig:planner_architecture}
    \vspace{-2em}
\end{figure}

\textbf{BEV features.}
The planner module is depicted in Fig.~\ref{fig:planner_architecture} inside the green box. 
We first voxelize the input point cloud to initial state $s^0$ and transform it to $h_r \times w_r$ bird's eye view (BEV) image.
Akin to PointPillars~\cite{lang2019pointpillar}, each pixel on the BEV image contains the feature extracted from the voxels within the vertical `pillar.'
Each pillar aggregates $3 \times 3 \times z_{max}$ voxels ($z$ is the up-axis, $z_{max}$ is the maximum voxels along $z$ axis) of the original voxel grid.
After the $x$ and $y$ coordinates of occupied voxels are converted into offsets from the pillar center and $z$ coordinate is normalized by $z_{max}$, a local PointNet~\cite{qi2016pointnet} processes them to obtain a feature for the corresponding pixel in the low-resolution BEV image.
We further add 2D positional encoding~\cite{vaswani2017attention} to encode relative position, and pass them through a dense 2D UNet~\cite{ronneberger2015unet} to obtain \emph{global} BEV features $f_{BEV}$.

\textbf{Training GCA with BEV features.}
As shown in Fig.~\ref{fig:planner_architecture}, we use 2D convolutions on $f_{BEV}$ to provide the global guidance (shown in pink) in the decoder layers of the sparse UNet of the GCA kernel (shown in blue).
Inspired by the spatial conditioning mechanism in SPADE~\cite{park2019SPADE}, the 2D convolutions compute a mean and variance per pillar. The means and variances per pillar are added and multiplied to the 3D sparse features falling within the pillar, effectively de-normalizing them.
To ensure $f_{BEV}$ contains necessary information to plan geometry, we apply an auxiliary guidance loss. Specifically, $f_{BEV}$ is trained to decode low-resolution 3D occupancy $\mathcal{O}_r$ of shape $(h_r, w_r, z_r)$ (we typically use $z_r=4$, implying voxels of $2$ meter height) with a 2D convolution layer.
It is supervised with a cross-entropy loss,
\begin{equation}
    \mathcal{L_{\text{BEV}}} = CE(\mathcal{O}_r, \mathcal{O}_r^\text{gt}), \label{eq:bev_loss}
\end{equation}
where $\mathcal{O}_r^\text{gt}$ is the ground truth coarse occupancy in the resolution of $\mathcal{O}_r$.
The final loss is a weighted combination of Eq.~\ref{eq:gca_loss} and~\ref{eq:bev_loss} with weight $\beta$,
\begin{equation}
    \mathcal{L} = \mathcal{L}_\text{GCA} + \beta \mathcal{L}_\text{BEV}.
    \label{eq:final_loss}
\end{equation}

\subsection{Upsampling to Continuous Geometry} \label{sec:upsampling}

Most prior works in semantic scene completion target AV perception, and  it suffices to predict occupancy at 20$\text{cm}^3$ voxel resolution given the LiDAR scan.
In contrast, hGCA can create high-resolution surfaces that are suitable for content creation.
Given the low-resolution generation of the scene from the previous step, hGCA generates voxels with latent vectors for local implicit functions~\cite{chabra2020deep_local_shapes, chiyu2020local, zhang2022probabilistic} in a higher resolution.
hGCA's hierarchical generation achieves efficiency in both space and time complexity for the large-scale under-constrained problem, disentangling geometry completion and upsampling into separate steps.

\textbf{Generative process.} 
We utilize a continuous version of GCA, named cGCA~\cite{zhang2022probabilistic}, as our generative model for conditional up-sampling into an implicit representation.
cGCA extends GCA to generate continuous surface, using an augmented state $x$ of each cell $c$, adding a local implicit latent feature $z_c$, i.e. $x = \{(c, o_c, z_c) | c \in \mathbb{Z}^3, o_c \in \{0, 1\}, z_c \in \mathbb{R}^K\}$, where $K$ is the dimension of the latent feature.
The local implicit latent features $z_c$ are in the latent space of a pre-trained auto-encoder, as in~\cite{zhang2022probabilistic}.
The encoder $g_\xi$ encodes coordinate-distance input pairs to $x$ and the decoder $f_\omega(x)$ decodes any point in $\mathbb{R}^3$ into an unsigned distance to surface. 
We additionally double the voxel resolution (\ie, 10$\text{cm}^3$) from our low resolution completion (20$\text{cm}^3$ voxels).
While one could theoretically obtain continuous surface even with implicit latent vectors in low-resolution voxels, we observe improved shape fidelity when using finer resolution voxels. 

As in GCA, a state $x^{t + 1}$ is sampled at each transition step $x^{t+1} \sim p_{\phi}(x^{t + 1} | x^t, x^0)$, where $x^0$ is the initial state.
We set coordinates in the sparse tensor $x^0$ to be the union of the input LiDAR scans and our low resolution completion $s^{T_1}$, all provided in a finer voxel resolution.
For cells $c$ that belong to the input point cloud, we set their latent feature $z_c$ using the pretrained encoder $g_\xi$.
We set the initial features $z_c$ to zeros for cells $c$ in $x^0$ that come solely from $s^{T_1}$.
After recursively sampling $T_2$ steps from $p_{\phi}$, the final state $x^{T_2}$ is decoded into a distance function using the pretrained decoder $f_\omega$, yielding an output mesh.
Further details regarding cGCA are in~\cite{zhang2022probabilistic}.

\textbf{Training.}
The training for upsampling is similar to that of GCA, derived for a continuous case.
Specifically, the training operates in the augmented state $x$, mapped using the encoder $g_\xi$ from coordinate-distance pairs.
For the initial state $x_0$, we use the ground truth low-resolution voxels $s^{gt}$ instead of the generated output from the first stage $s^{T_1}$.
This enforces the upsampler to solely learn to upsample, and the two stages are trained independently and therefore efficiently. 
If we use the outputs $s^{T_1}$, the stochasticity may lead to inconsistent supervision and training instability. We refer the readers to the Appendix for further details.


\begin{table*}[t!]
    \vspace{-3em}
    \centering
    \resizebox{\linewidth}{!}{
        \begin{tabular}{l|c|cc|c|c|cc|c|c|cc|c|cc|c|c|cc|c|c|cc|c}
            \toprule
            & & \multicolumn{11}{c|}{5 scans} &  \multicolumn{11}{c}{10 scans}  \\
            \midrule
            & & \multicolumn{4}{c|}{CARLA} &  \multicolumn{7}{c|}{Karton City} & \multicolumn{4}{c|}{CARLA} &  \multicolumn{7}{c}{Karton City} \\
            \midrule
            \multirow{2}{*}{Method} & \multirow{2}{*}{\shortstack[l]{Represe\\-ntation}} & \multicolumn{3}{c|}{High LiDAR ReSim} & \multirow{2}{*}{IoU} &  \multicolumn{3}{c|}{High LiDAR ReSim} & \multirow{2}{*}{IoU} & \multicolumn{3}{c|}{Street CD} & \multicolumn{3}{c|}{High LiDAR ReSim} & \multirow{2}{*}{IoU} &  \multicolumn{3}{c|}{High LiDAR ReSim} & \multirow{2}{*}{IoU} & \multicolumn{3}{c}{Street CD}  \\
            & & min. $\downarrow$ & avg. $\downarrow$ & TMD $\uparrow$ & & min. $\downarrow$ & avg. $\downarrow$ & TMD $\uparrow$ & & min. $\downarrow$ & avg. $\downarrow$ & TMD $\uparrow$ & min. $\downarrow$ & avg. $\downarrow$ & TMD $\uparrow$ & & min. $\downarrow$ & avg. $\downarrow$ & TMD $\uparrow$ & & min. $\downarrow$ & avg. $\downarrow$ & TMD $\uparrow$\\
            \midrule
            ConvOcc & implicit & \dc{15.52} & - & 13.40 & \dc{10.35} & - & 25.64 & \dc{17.13} & -  & \dc{14.62} & - & 13.74 & \dc{9.25} & - & 26.54 & \dc{15.13} & - \\
            SCPNet & $\text{20cm}^3$ & \dc{5.77} & - & 49.82 & \dc{4.82} & - & 68.53 & \dc{3.64} & -  & \dc{5.47} & - & 52.49 & \dc{4.28} & - & 72.48 & \dc{3.14} & - \\
            \multirow{2}{*}{JS3CNet} & $\text{20cm}^3$ & \dc{6.58} & - & 51.02 & \dc{5.28} & - & 63.76 & \dc{3.92} & -  & \dc{6.29} & - & 53.39 & \dc{5.02} & - & 65.68 & \dc{3.26} & - \\
            & $\text{10cm}^3$ & \dc{6.64} & - & 43.46 & \dc{3.86} & - & 70.28 & \dc{2.30} & -  & \dc{4.99} & - & 46.76 & \dc{3.46} & - & 73.11 & \dc{1.92} & - \\
            SG-NN & $\text{10cm}^3$ & \dc{5.06} & - & 50.76 & \dc{4.06} & - & 70.18 & \dc{2.61} & -  & \dc{4.53} & - & 54.29 & \dc{3.42} & - & 73.58 & \dc{2.04} & - \\
            \multirow{2}{*}{GCA} & $\text{20cm}^3$ & 5.58 & 5.83 & 1.45 & 50.91 & 3.95 & 4.03 & 0.61 & 74.95 & 2.87 & 3.34 & 1.16 & 5.30 & 5.54 & 1.36 & 54.40  & 3.79 & 3.85 & 0.45 & 78.04 & 2.52 & 2.73 & 0.77 \\
            & $\text{10cm}^3$ & 5.66 & 6.17 & 2.25 & 44.26 & 3.93 & 4.10 & 1.16 & 68.23 & 3.38 & 4.16 & \textbf{2.28} & 5.13 & 5.52 & 1.96 & 48.26 & 3.28 & 3.40 & 0.83 & 72.38 & 2.58 & 3.11 & \textbf{1.45} \\ 
            cGCA & implicit & 7.04 & 7.59 & 3.19 & 35.43 & 4.29 & 4.42 & 1.23 & 59.79 & 2.49 & 3.36 & 2.00 & 6.84 & 7.43 & 2.97 & 36.17 & 4.00 & 4.09 & 0.91 & 63.56 & 1.65 & 2.02 & 0.92 \\
            \midrule
            \multirow{2}{*}{hGCA} & $\text{10cm}^3$ & 4.60 & 4.72 & 0.80 & \textbf{53.84} & \textbf{3.20} & 3.25 & 0.51 & \textbf{75.97} & 2.09 & 2.27 & 0.64 & 4.38 & 4.48 & 0.78 & \textbf{56.85} & 2.97 & 3.01 & 0.41 & \textbf{79.63} & 1.79 & 1.89 & 0.54 \\
            & implicit & \textbf{4.53} & 4.65 & 0.92 & 52.17 & \textbf{3.20} & 3.25 & 0.56 & 70.45 & \textbf{1.85} & 2.02 & 0.51 & \textbf{4.30} & 4.40 & 0.88 & 54.68 & \textbf{2.95} & 2.99 & 0.45 & 73.38 & \textbf{1.55} & 1.65 & 0.42 \\
            \midrule
            \dc{input} & \dc{6.33} & - & 34.43 & \dc{6.83} & - & 38.32 & \dc{5.63} & -  & \dc{5.46} & - & 40.42 & \dc{5.45} & - & 47.71 & \dc{4.99} & -  \\
            \bottomrule
        \end{tabular}%
        }
        \vspace{-1em}
        \caption{
    	Quantitative results on CARLA and Karton City with 5 and 10 scans given as input.
        All results except IoU are multiplied by 10 in meter scale.
        LiDAR Resim and Street CD evaluates the fidelity of completion and TMD measures the diversity of generation.
        High LiDAR Resim uses high elevation LiDAR to evaluate the extrapolation.
        IoU is computed with ground truth geometry.
        \vspace{-2em}
    }
    \label{table:scan_5_main}
   
\end{table*}

\section{Experiments}
We evaluate hGCA on street scene generation given LiDAR scans captured from AVs.
We assume registered sequential scans \ie relative poses between captures are known.
We train and evaluate on synthetic street scenes from CARLA~\cite{carla} and Turbosquid \footnote{\label{turbosquidfootnote}\href{https://www.turbosquid.com/3d-models/3d-karton-city-2-model-1196110}{Clickable link to asset on turbosquid.com}} against state-of-the-art methods in Sec.~\ref{sec:exp_synthetic} for scene extrapolation quality, using ground truth geometry. 
We use synthetic LiDAR scans to train, matching the LiDAR scan pattern from Waymo-Open~\cite{Waymo-Open}.
In Sec.~\ref{sec:exp_real}, we test generalization abilities of hGCA on real LiDAR scans from Waymo-Open~\cite{Waymo-Open} and on novel objects, unseen in training. Lastly, in Sec.~\ref{sec:ablation_planner}, we investigate the planner module.
We provide further analysis in Appendix.

\textbf{Datasets.}
1) \textbf{Karton City} is a synthetic city comprised of 20 city blocks, obtained from the Turbosquid 3D asset marketplace\textsuperscript{\ref{turbosquidfootnote}}. 
We split 20 blocks into train/val/test splits and re-combine 4 blocks in each split randomly per scene. We simulate parked cars by placing car assets from ShapeNet~\cite{shapenet2015}. 
2) \textbf{CARLA~\cite{carla}} is an open source driving simulator with diverse environments. We use 5/1/1 towns as train/val/test split with randomly placed static vehicles. We simulate random ego-vehicle trajectories on synthetic data for training and evaluation, detailed in the Appendix.
3) \textbf{Waymo-open}~\cite{Waymo-Open} is a real-world AV dataset with registered LiDAR scans, used here to demonstrate sim-to-real generalization qualitatively. We discuss issues with quantitatively evaluating 3D generative models on real data in Sec.~\ref{sec:exp_real} and the Appendix.

\textbf{Evaluation Metrics}.
We evaluate our generated scenes using three metrics to capture various aspects of scene extrapolation.
1) \textbf{High LiDAR ReSim} evaluates geometry fidelity, focusing on regions visible from the street. 
It measures the chamfer distance (CD) between two LiDAR scans from poses distant from the center, one from the ground truth (GT) mesh, and the other from the completed scene.
For this metric, we add high elevation angles to the LiDAR sensor to evaluate generation \emph{beyond} maximum input height. 
The metric (deliberately) avoids evaluating on inconsistent geometry in interior walls of buildings in GT, which are invisible to LiDAR from the street (green boxes in  Fig.~\ref{fig:qual_synthetic}).
Following evaluation in semantic scene completion ~\cite{SemanticKITTI}, we compute 2) \textbf{IoU} at $\text{20cm}^3$ voxel resolution, for all voxels visible to the high elevation LiDAR (to measure beyond input height) from all novel sampled poses on the ego-vehicle trajectory, also used in LiDAR ReSim. In contrast to the point-wise High LiDAR ReSim metric, IoU evaluates rough occupancy of large scene context, independently of the 3D representation used in generated geometry. 
Additionally, we propose 3) \textbf{Street CD} to include evaluation on geometry completely occluded from the ego-trajectory, such as the sidewalk side of parked cars. 
On Karton City dataset, where the scene is a simple crossroad junction, we compute Chamfer distance between the generated geometry against GT, only on the objects on the main street.
Due to the simplicity and abundance of flat ground in Karton City, we remove it from evaluation by simple height thresholding.
For generative models, we generate $k=3$ generations and measure minimum and average metrics to account for stochasticity, and also measure Total Mutual Difference (TMD)~\cite{wu_2020_ECCV} to capture generation diversity.

\textbf{Baselines.}
We compare hGCA with state-of-the-art AV semantic scene completion methods (JS3CNet~\cite{JS3CNet}, SCPNet~\cite{scpnet}), indoor scene completion methods (SG-NN~\cite{dai2020sgnn}, ConvOcc~\cite{Peng2020ECCV}), and generative models based on GCA (GCA~\cite{zhang2021learning}, cGCA~\cite{zhang2022probabilistic}), training all models from scratch. We adapt semantic scene completion methods to our setting by changing the semantic class output to binary occupancy. We refer to Appendix for more details on datasets, evaluation metrics and baselies.

\textbf{Implementation details.}
For our coarse stage, we use $\text{20cm}^3$ voxel size, $T_1=30$ transition steps with radius $r=1$.
In the upsampling model, we use $\text{10cm}^3$ size, $T_2=15$ transition steps with radius $r=2$.
We set BEV loss weight $\beta=0.1$ for all experiments, with planner parameters $h_r = h_{max}/3, w_r=w_{max}/3, z_r=4$ unless stated otherwise. 
We train and infer (unless specified) on scenes in a volume of $38.4 \times 38.4 \times 8$ meters with height ranging from $[-1, 7]$ meters in a ego vehicle frame, randomly selected from one of the poses the input scans. 
At $\text{20cm}^3$ voxel resolution, this corresponds to $h_{max}=192$, $w_{max}=192$ and $z_{max}=40$. All experiments were performed on a single 24GB RTX 3090 GPU.
We obtain and reuse the pre-trained latent autoencoder $g_\xi$ and 
$f_\omega$ for local-implicits used in cGCA~\cite{zhang2022probabilistic}, trained on indoor scenes from 3DFront~\cite{3dfront}, which generalizes well to our data.
We simulate synthetic LiDAR with simple ray-casting and add noise to LiDAR scan coordinates and relative poses, which improves sim-to-real generalization. We report results without LiDAR noise in the Appendix.

\subsection{Synthetic Scene Completion}\label{sec:exp_synthetic}

\def \qualwidth{0.47}
\begin{figure*}
    \begin{minipage}{0.5\textwidth}
        \rotatebox[origin=lB]{90}{\small{\;\;\;\;\;\;\;\;Input}}
        \includegraphics[width=\qualwidth\linewidth]{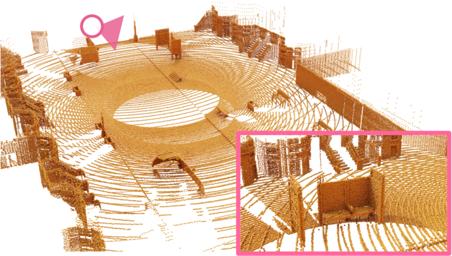}
        \hfill
        \includegraphics[width=\qualwidth\linewidth]{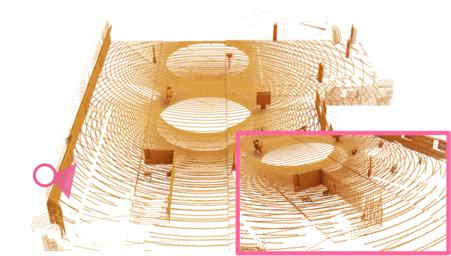}
        \hfill
        \\
        \rotatebox[origin=lB]{90}{\small{\;\;\;\;\;\;\;\;\;\;GT}}
        \includegraphics[width=\qualwidth\linewidth]{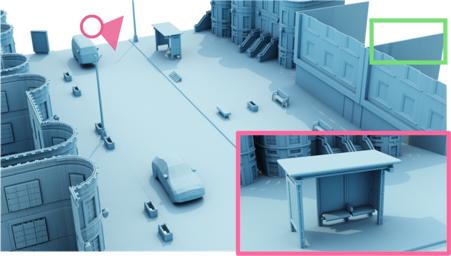}
        \hfill
        \includegraphics[width=\qualwidth\linewidth]{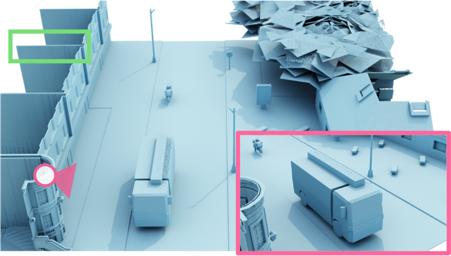}
        \hfill
        \\        
        \rotatebox[origin=lB]{90}{\small{\;\;\;\;\;\;\;SCPNet}}
        \includegraphics[width=\qualwidth\linewidth]{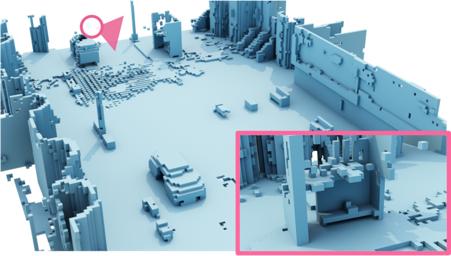}
        \hfill
        \includegraphics[width=\qualwidth\linewidth]{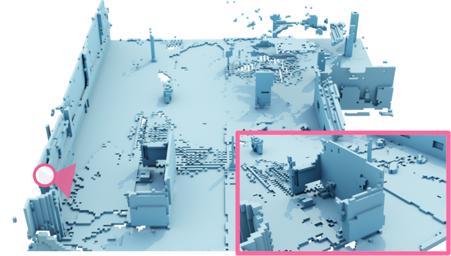}
        \hfill
        \\
        \rotatebox[origin=lB]{90}{\small{\;\;\;\;\;\;\;SG-NN}}
        \includegraphics[width=\qualwidth\linewidth]{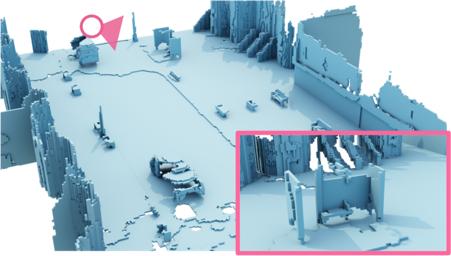}
        \hfill
        \includegraphics[width=\qualwidth\linewidth]{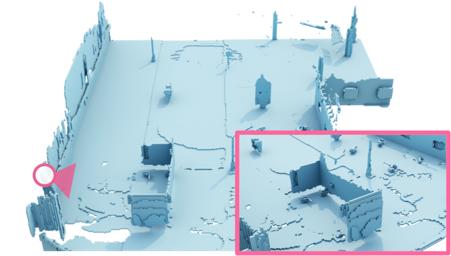}
        \hfill
        \\
        \rotatebox[origin=lB]{90}{\small{\;\;\;\;GCA }\scriptsize{($\text{20cm}$)}}
        \includegraphics[width=\qualwidth\linewidth]{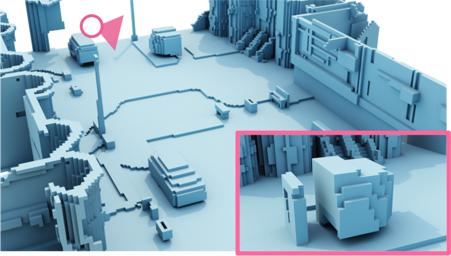}
        \hfill
        \includegraphics[width=\qualwidth\linewidth]{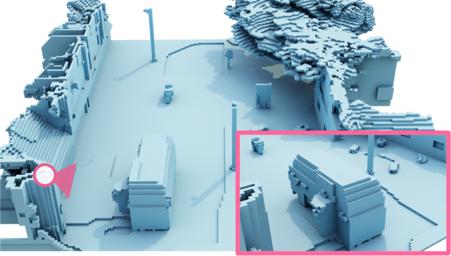}
        \hfill
        \\
        \rotatebox[origin=lB]{90}{\small{\;\;\;\;\;\;\;\;cGCA}}
        \includegraphics[width=\qualwidth\linewidth]{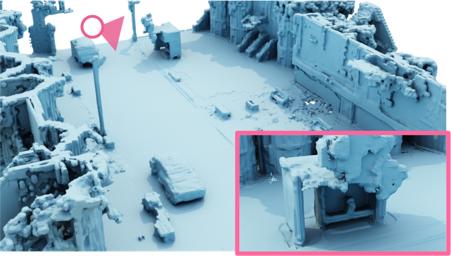}
        \hfill
        \includegraphics[width=\qualwidth\linewidth]{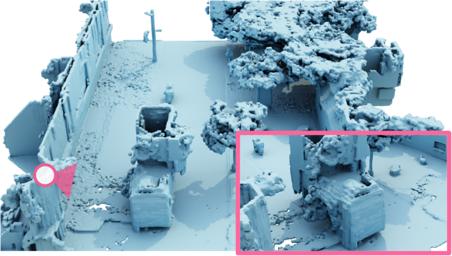}
        \hfill
        \\
        \rotatebox[origin=lB]{90}{\small{\;\;\;\;\;\;\;\;hGCA}}
        \includegraphics[width=\qualwidth\linewidth]{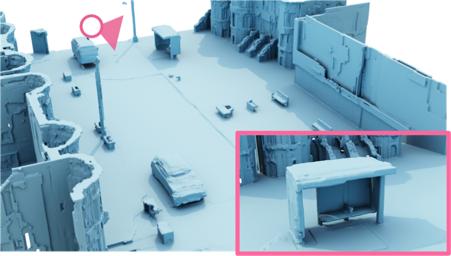}
        \hfill
        \includegraphics[width=\qualwidth\linewidth]{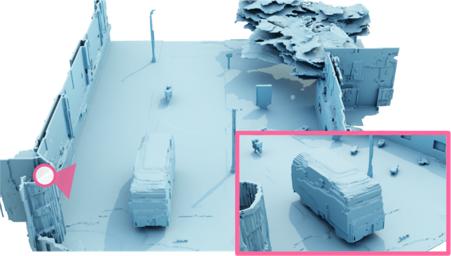}
        \hfill
    \end{minipage}
    \begin{minipage}{0.5\textwidth}
        \rotatebox[origin=lB]{90}{\small{\;\;\;\;\;\;\;\;Input}}
        \includegraphics[width=\qualwidth\linewidth]{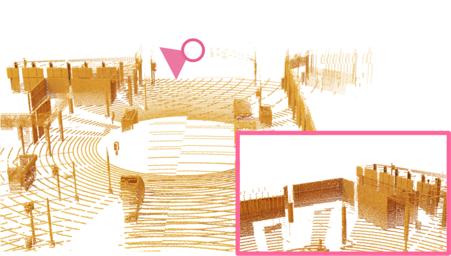}
        \hfill
        \includegraphics[width=\qualwidth\linewidth]{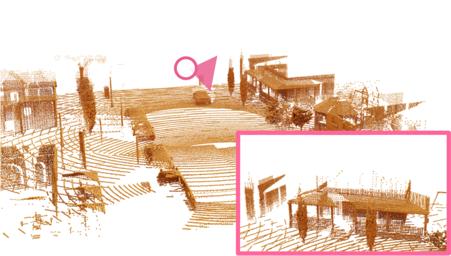}
        \hfill
        \\
        \rotatebox[origin=lB]{90}{\small{\;\;\;\;\;\;\;\;\;\;GT}}
        \includegraphics[width=\qualwidth\linewidth]{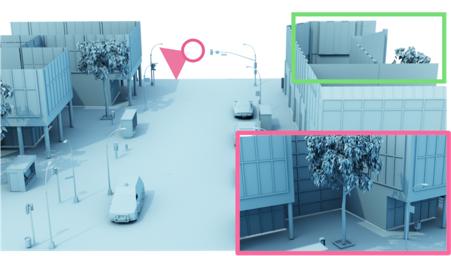}
        \hfill
        \includegraphics[width=\qualwidth\linewidth]{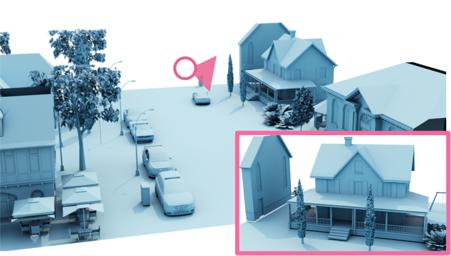}
        \hfill
        \\
        \rotatebox[origin=lB]{90}{\small{\;\;\;\;\;ConvOcc}}
        \includegraphics[width=\qualwidth\linewidth]{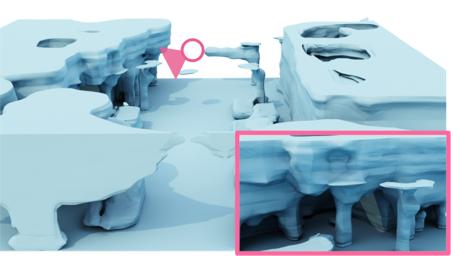}
        \hfill
        \includegraphics[width=\qualwidth\linewidth]{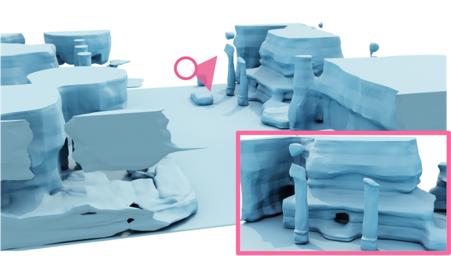}
        \hfill
        \\
        \rotatebox[origin=lB]{90}{\small{\;\;JS3CNet } \scriptsize{($\text{20cm}$)}}
        \includegraphics[width=\qualwidth\linewidth]{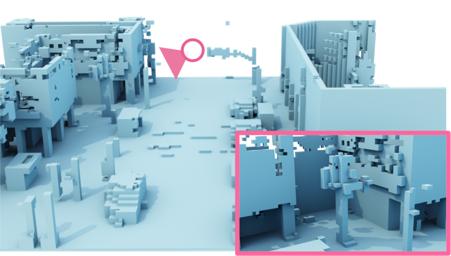}
        \hfill
        \includegraphics[width=\qualwidth\linewidth]{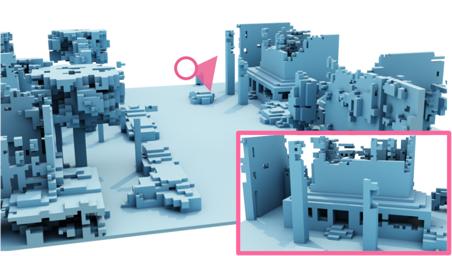}
        \hfill
        \\
        \rotatebox[origin=lB]{90}{\small{\;\;JS3CNet } \scriptsize{($\text{10cm}$)}}
        \includegraphics[width=\qualwidth\linewidth]{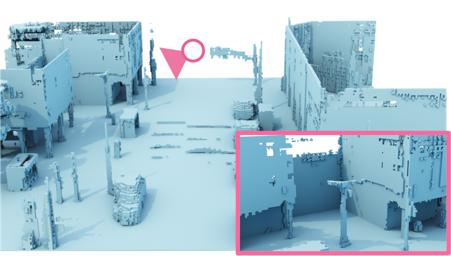}
        \hfill
        \includegraphics[width=\qualwidth\linewidth]{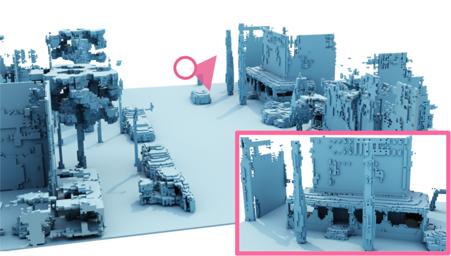}
        \hfill
        \\
        \rotatebox[origin=lB]{90}{\small{\;\;\;\;GCA }\scriptsize{($\text{10cm}$)}}
        \includegraphics[width=\qualwidth\linewidth]{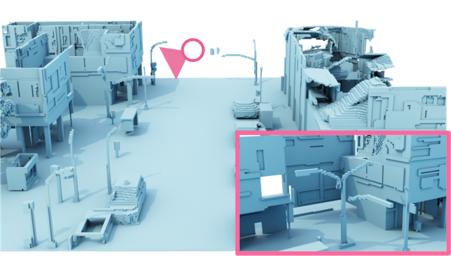}
        \hfill
        \includegraphics[width=\qualwidth\linewidth]{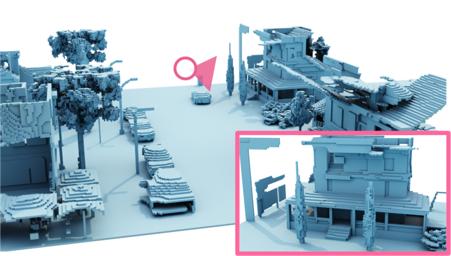}
        \hfill
        \\
        \rotatebox[origin=lB]{90}{\small{\;\;\;\;\;\;\;\;hGCA}}
        \includegraphics[width=\qualwidth\linewidth]{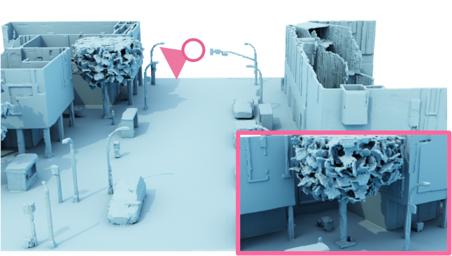}
        \hfill
        \includegraphics[width=\qualwidth\linewidth]{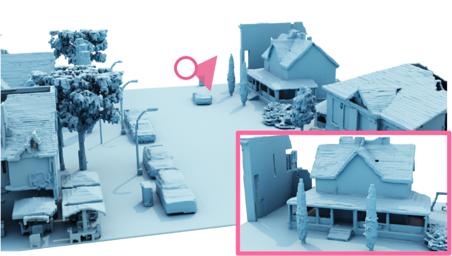}
        \hfill
    \end{minipage}
    \vspace{-1em}
    \caption{
    Visualizations on CARLA (first 2 columns) and Karton City (last 2 columns) from 5 scans.
    hGCA generates high-resoluton geometry beyond field of view (bus stops, trees, roofs) and occlusions (cars) compared to existing baselines. Deterministic baselines tend to conservatively complete high-confidence regions near the input.
    Green boxes demonstrate inconsistency of building interiors in GT data.
    }
    \vspace{-1em}
    \label{fig:qual_synthetic}
\end{figure*}

Scene extrapolation results on synthetic scenes are reported in Tab.~\ref{table:scan_5_main}. We accumulate 5 or 10 LiDAR scans from random poses as input. We train all models on combined CARLA and Karton City data for added diversity, but evaluate separately. Output representation for baselines are voxels or continuous surfaces when available.
hGCA outperforms all baselines by a margin in reconstruction metrics while generating diverse outputs. For hGCA, we report scores of $\text{10cm}^3$ voxel occupancy and implicit representation, both obtained after upsampling. We notice that IoUs are slightly higher with our $\text{10cm}^3$ voxels, resulting from our unsigned distance fields sometimes not generating clear zero-level iso-surfaces, creating thick meshes for thin structures after thresholding, similar to ~\cite{zhang2022probabilistic}. We find that the planner trades off diversity for quality and global consistency, discussed further in Sec.~\ref{sec:ablation_planner}.
We show qualitative results in Fig.~\ref{fig:qual_synthetic}, and many more in the Appendix. Deterministic completion models (ConvOcc~\cite{Peng2020ECCV}, SCPNet~\cite{scpnet}, JS3CNet~\cite{JS3CNet}, SG-NN~\cite{dai2020sgnn}) tend to conservatively generate geometry beyond the input, such as bus stops or cars in Fig.~\ref{fig:qual_synthetic}. These approaches lack multi-modality and we hypothesize that it limits generation to high-confidence geometry near sparse inputs by tending to model a mean distribution.
In contrast, hGCA generates well completed geometry with high fidelity.

\subsection{Generalization across Domains}\label{sec:exp_real}
\begin{figure*}[t!]
    \centering
    \includegraphics[width=\linewidth]{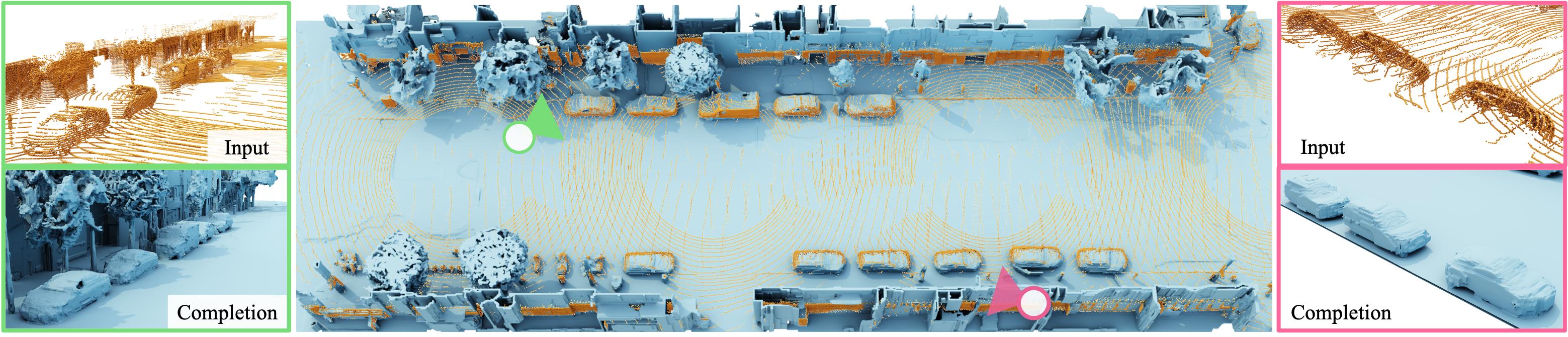}
    \vspace{-2em}
    \caption{
    Completion given accumulation of 5 LiDAR scans (yellow spheres) on real-world Waymo-open dataset. 
    hGCA can extrapolate beyond input field of view (walls) and occlusion (cars).
    Walls in pink boxes are cut off for ease of visualization.
    }
    \vspace{-2em}
    \label{fig:waymo_big}
\end{figure*}
\def \qualwidth{0.47}
\def \whrate{0.1}
\begin{figure}[]
    \vspace{-2em}
    \centering
    \begin{subfigure}{\linewidth}
        \rotatebox[origin=lB]{90}{\small{\;\;\;\;\;\;\;\;Input}}
        \includegraphics[width=\qualwidth\linewidth]{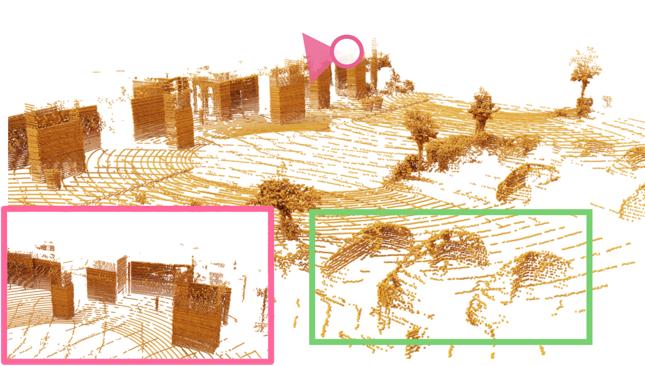}
        \hfill
        \includegraphics[width=\qualwidth\linewidth]{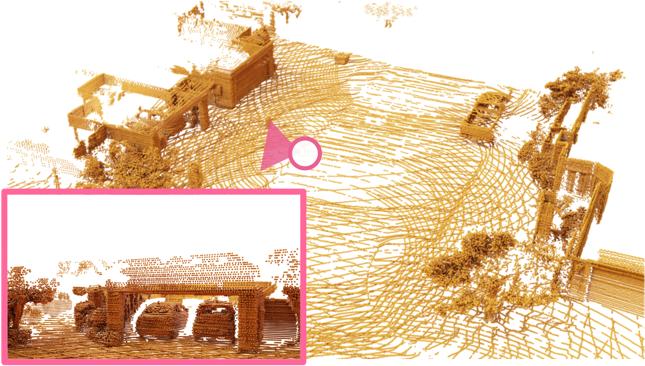}
        \\
        \rotatebox[origin=lB]{90}{\small{\;\;\;Acc. scans}}
        \includegraphics[width=\qualwidth\linewidth]{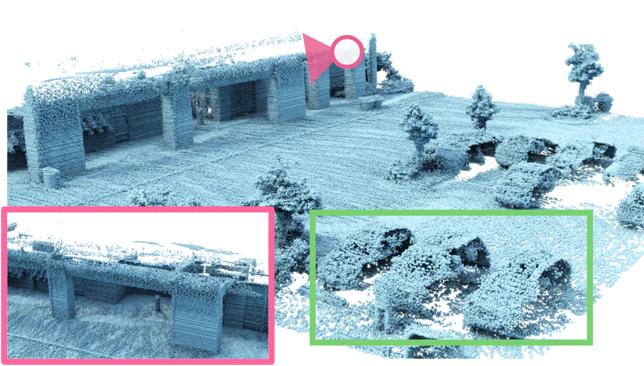}
        \hfill
        \includegraphics[width=\qualwidth\linewidth]{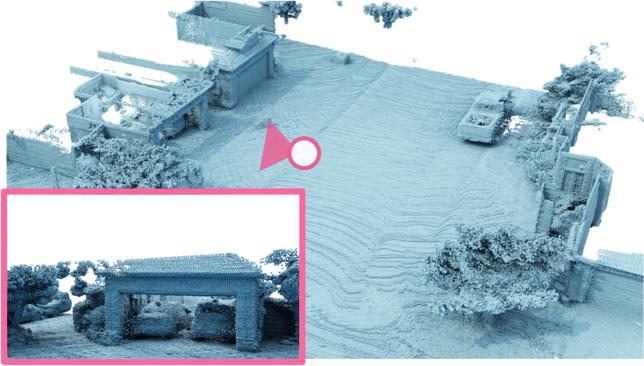}
        \\
        \rotatebox[origin=lB]{90}{\small{\;\;\;\;\;\;\;SCPNet}}
        \includegraphics[width=\qualwidth\linewidth]{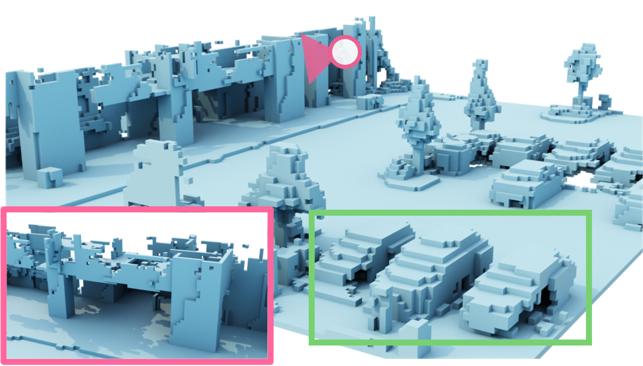}  
        \hfill
        \includegraphics[width=\qualwidth\linewidth]{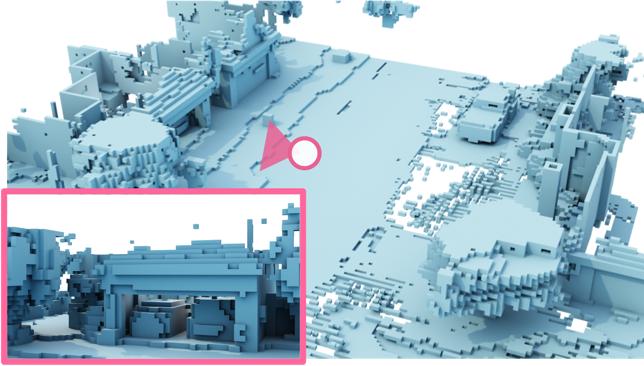}
        \\
        \rotatebox[origin=lB]{90}{\small{\;\;\;\;\;\;\;SG-NN}}
        \includegraphics[width=\qualwidth\linewidth]{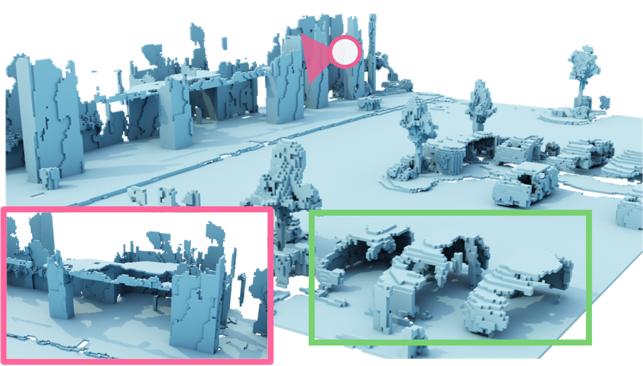}  
        \hfill
        \includegraphics[width=\qualwidth\linewidth]{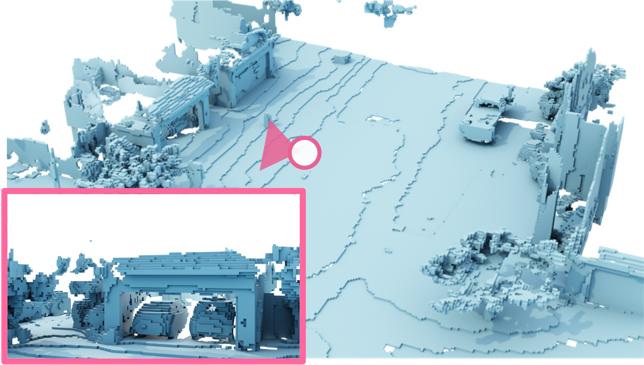}
        \\
        \rotatebox[origin=lB]{90}{\small{\;\;\;\;GCA }\scriptsize{(20cm)}}
        \includegraphics[width=\qualwidth\linewidth]{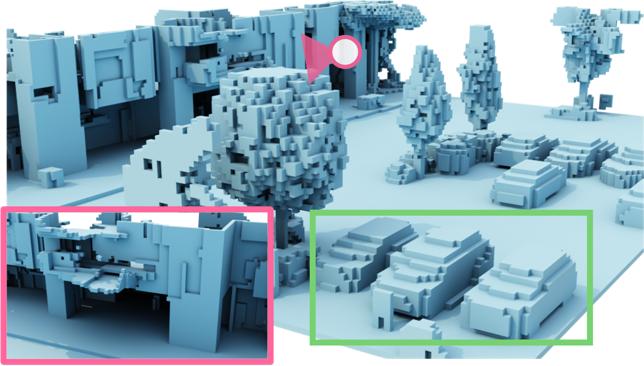}  
        \hfill
        \includegraphics[width=\qualwidth\linewidth]{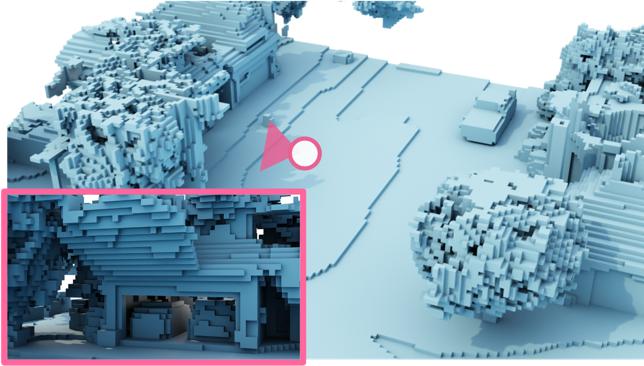}
        \\
        \rotatebox[origin=lB]{90}{\small{\;\;\;\;\;\;\;\;hGCA}}
        \includegraphics[width=\qualwidth\linewidth]{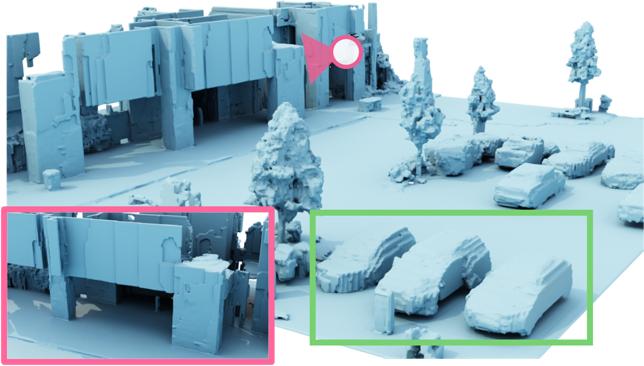}  
        \hfill
        \includegraphics[width=\qualwidth\linewidth]{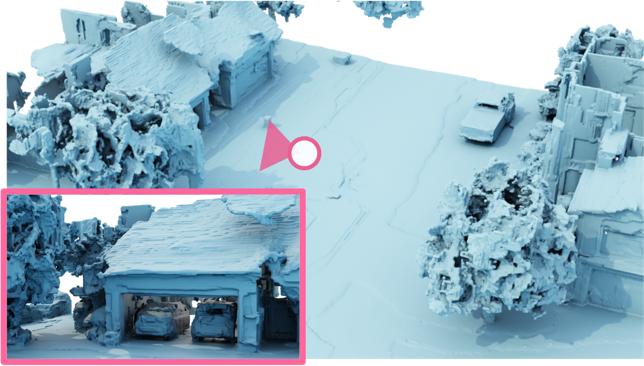}  

	\end{subfigure}
    \vspace{-2em}
    \caption{
        Visualizations on real-world Waymo-open dataset.
        hGCA exhibits great sim-to-real performance compared to existing method with high fidelity (pink box) and can generate more complete shapes than accumulated scans (green box).
    }
    \vspace{-2em}
    \label{fig:qual_real}
\end{figure}

\textbf{Generalization to real LiDAR scans.}
We find that hGCA generalizes well from sim-to-real, successfully completing cars and trees from Waymo-open LiDAR scans, using 5 accumulated scans as input, visualized in Fig.~\ref{fig:waymo_big}, Fig.~\ref{fig:qual_real}, and more in the Appendix.
Fig.~\ref{fig:qual_real} shows that deterministic baselines again exhibit conservative behavior, whereas naive-GCA suffers from inconsistency, in one case generating a tree from a house.
Both GCA and hGCA complete occluded cars in the parking lot or even inside of a garage (Fig.~\ref{fig:qual_real}), where the latter has never been seen in the synthetic training data. 
We hypothesize that this generalization stems from the locality of GCA and the two-stage approach where the coarser GCA is more robust to real-world noise.
Overall, hGCA can generate convincing completions, exhibiting geometric quality inferior, yet closer, to the synthetic data it was trained on compared to baselines, taking a step towards simulation-ready environment creation using AV LiDAR as a content scanner.
Evaluating a 3D generative model on real AV data is challenging. 
The best source of ground truth geometry available to us is using all accumulated scans as in semantic scene completion, also shown in Fig.~\ref{fig:qual_real}, which is highly incomplete and has limited height range. 
For example, in the left of Fig.~\ref{fig:qual_real}, hGCA generates more complete geometry, but is worser on LiDAR Resim or IoU scores compared to baselines (SCPNet: 3.94/63.93, SG-NN: 2.97/58.67,  hGCA: 3.58/63.46), using held-out real scans in the scene for LiDAR ReSim and IoU against accumulated scans. 
We discuss difficulties of evaluation on real-world datasets further in the Appendix.

\begin{figure}[t]
    \centering
    \includegraphics[width=\linewidth]{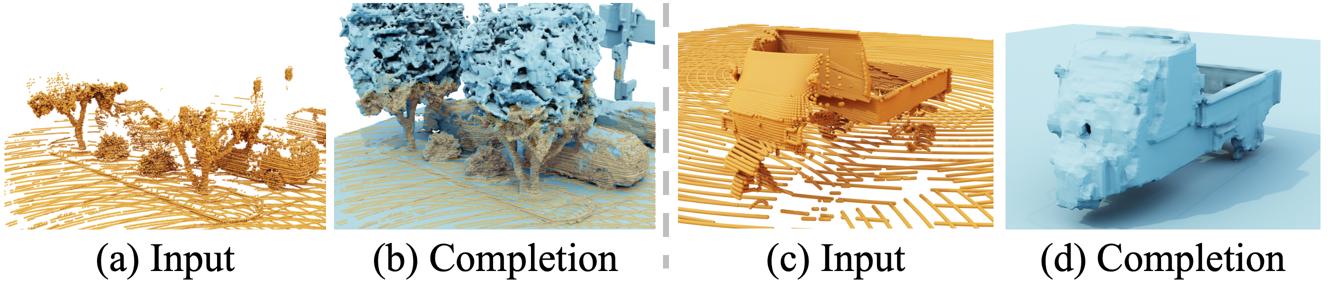}
    \vspace{-2em}
    \caption{
        (a), (b): Completion on LiDAR scan from Waymo-open.
        (c), (d): Completion on synthetic LiDAR of a three-wheeler asset from sketchfab~\ref{sketchfab_footnote}
        hGCA can realistically complete from tree trunks or three-wheeler cars unseen in training, taking geometric cues from the input (yellow spheres).
    }
    \vspace{-3em}
    \label{fig:asset_rep}
\end{figure}

\textbf{Out-of-distribution inputs.}
A fair concern with training on synthetic content is the limited diversity of assets the generative model is trained on, which could be reflected in the outputs. We show anecdotal evidence of novel asset completion in Fig.~\ref{fig:asset_rep}.
On the right, we show geometry completion from sythetic LiDAR of a three-wheeler vehicle asset taken from Objaverse-XL~\cite{deitke2023objaverse}\footnote{\label{sketchfab_footnote}\href{https://skfb.ly/osHxF}{Clickable link to asset on sketchfab.com}}.
We verify that no three-wheelers exist in our training data. hGCA generates a convincing three-wheeler, respecting the input, better than deterministic baselines, which we visualize in Appendix. We also show how hGCA can complete this asset with some lower density input scans in Appendix. The left of Fig.~\ref{fig:asset_rep}, shows LiDAR scans from Waymo-open of complex, unique tree trunks. hGCA, trained only on single trunk trees, is able to generate realistic trees that preserve the observed structure. These results require further rigorous validation, but show promise towards environment creation with diverse assets from geometric cues taken from real LiDAR scans.

\newcommand{\multic}[1]{\multicolumn{2}{c|}{#1}}

\begin{table}[t]
    \centering
    \vspace{-2em}
    \resizebox{\linewidth}{!}{
        \begin{tabular}{l|c|cc|c|c|cc|c|c}
            \toprule
            & & \multicolumn{4}{c|}{CARLA} &  \multicolumn{4}{c}{Karton City}  \\
            \midrule
            \multirow{2}{*}{\shortstack[l]{Voxel\\Size}} & \multirow{2}{*}{$z_r$} & \multicolumn{3}{c|}{LiDAR ReSim} & \multirow{2}{*}{IoU} &  \multicolumn{3}{c|}{LiDAR ReSim} & \multirow{2}{*}{IoU} \\
            & & min. $\downarrow$ & avg. $\downarrow$ & TMD $\uparrow$ & & min. $\downarrow$ & avg. $\downarrow$ & TMD $\uparrow$ & \\
            \midrule
            \multirow{4}{*}{$\text{20cm}^3$} & \xmark & 5.58 & 5.83 & \textbf{1.45} & 50.91 & 3.95 & 4.03 & \textbf{0.61} & 74.95 \\
            \cmidrule(rl){2-10}
            & 27 & 5.44 & 5.57 & 0.94 & 49.37 & \textbf{3.92} & 3.97 & 0.41 & \textbf{76.34}\\
            & 16 & 5.37 & 5.54 & 0.91 & 51.67 & 3.96 & 4.00 & 0.41 & 76.2\\
            & 4 & \textbf{5.28} & 5.40 & 0.77 & \textbf{53.81} & 3.97 & 4.03 & 0.44 & 75.86\\
            & 2 & 5.32 & 5.46 & 0.86 & 52.49 & 3.99 & 4.05 & 0.47 & 75.49\\
            
            \midrule
            \multirow{2}{*}{$\text{10cm}^3$} & \xmark & 5.66 & 6.17 & \textbf{2.25} & 44.26 & 3.93 & 4.10 & 1.16 & 68.23\\
            \cmidrule(rl){2-10}
            & 4 & \textbf{4.58} & 4.74 & 0.93 & \textbf{51.40} & \textbf{3.44} & 3.52 & 0.67 & \textbf{72.12} \\
            \bottomrule
        \end{tabular}%
    }
    \vspace{-1em}
        \caption{
            Ablation study on effects of Planner from 5 input scans.
            \xmark \; in $z_r$ refers to vanilla GCA without Planner module.
        }
        \vspace{-2em}
    \label{table:planner_z}
   
\end{table}

\subsection{Ablation Studies on Planner}\label{sec:ablation_planner}
The planner module aims to induce global consistency in hGCA. Table~\ref{table:planner_z} shows quantatively that it trades of diversity for completion performance compared to vanilla GCA
We test different resolutions of planner occupancy prediction in height, indicated by $z_r$. We found that $z_r = 4$ is a good balance between providing global context without hurting generation performance on CARLA, which has a more diverse validation set.
We can infer that predicting occupancy in finer vertical resolution (large $z_r$) may be beyond the capacity of our simple planner module and hinders joint optimization of the local GCA loss with the global planner loss.
We observed that the planner does not boost the performance of the upsampling module, indicating that local upsampling does not benefit from coarse global context.

\section{Conclusion}
We proposed hierarchical Generative Cellular Automata (hGCA), a spatially scalable generative model that generates 3D scenes beyond occlusions and input field of view from several LiDAR scans.
Our model generates scenes in a two-stage hierarchical coarse-to-fine manner, where the first stage generates coarse geometry by providing global consistency to GCA with a light-weight planner module. 
The second stage synthesizes finer details by applying cGCA conditioned on the coarse geometry.
On synthetic scenes, hGCA generates plausible scenes with higher fidelity and completness compared to prior state-of-the-art works.
hGCA demonstrates strong sim-to-real generalization, capable of extrapolating LiDAR scans on real-world Waymo dataset.
While hGCA takes a step towards content creation from LiDAR scans, several desiderata remain. Improving fidelity of geometry and generating textures, materials etc. is needed for usability of the completed geometry.
For example, in Fig.~\ref{fig:qual_real} right column, hGCA generates inconsistent roofs. 
The current generative process of hGCA is slow, disabling use of the model in real-time, which we leave to future work.


\textbf{Acknowledgements.} This work was supported by Institute of Information communications Technology Planning Evaluation (IITP) grant funded by the Korea government(MSIT) [NO.2021-0-01343, Artificial Intelligence Graduate School Program (Seoul National University)] and Creative-Pioneering Researchers Program through Seoul National University.

{
    \small
    \bibliographystyle{ieeenat_fullname}
    \bibliography{main}
}
\appendix
\section{Additional Analysis} \label{app:further_analysis}
In this section, we include additional results that highlight the practical aspects of the proposed method.

\subsection{Evaluation on Real-World Data}\label{app:real_data}

\subsubsection{Waymo-Open}

\begin{figure*}[]
    \centering
    \includegraphics[width=\linewidth]{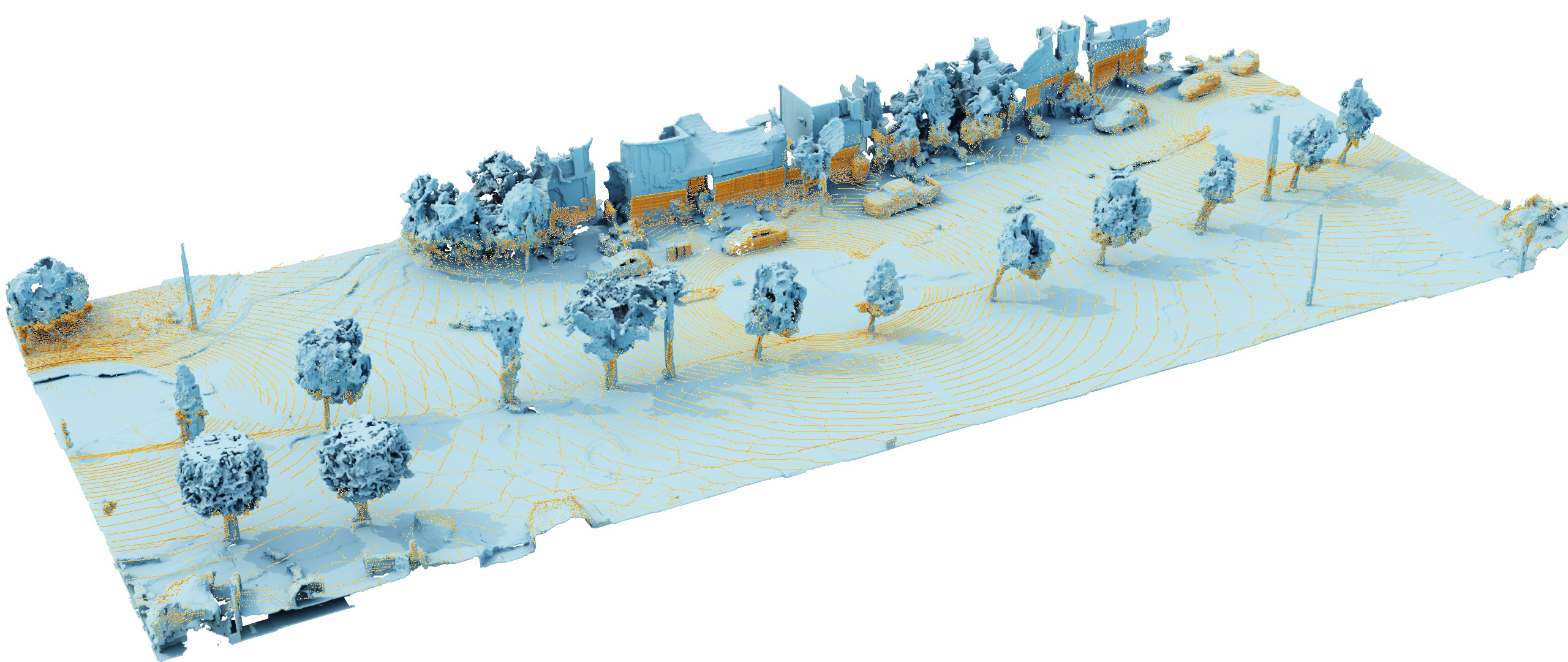}
    \includegraphics[width=\linewidth]{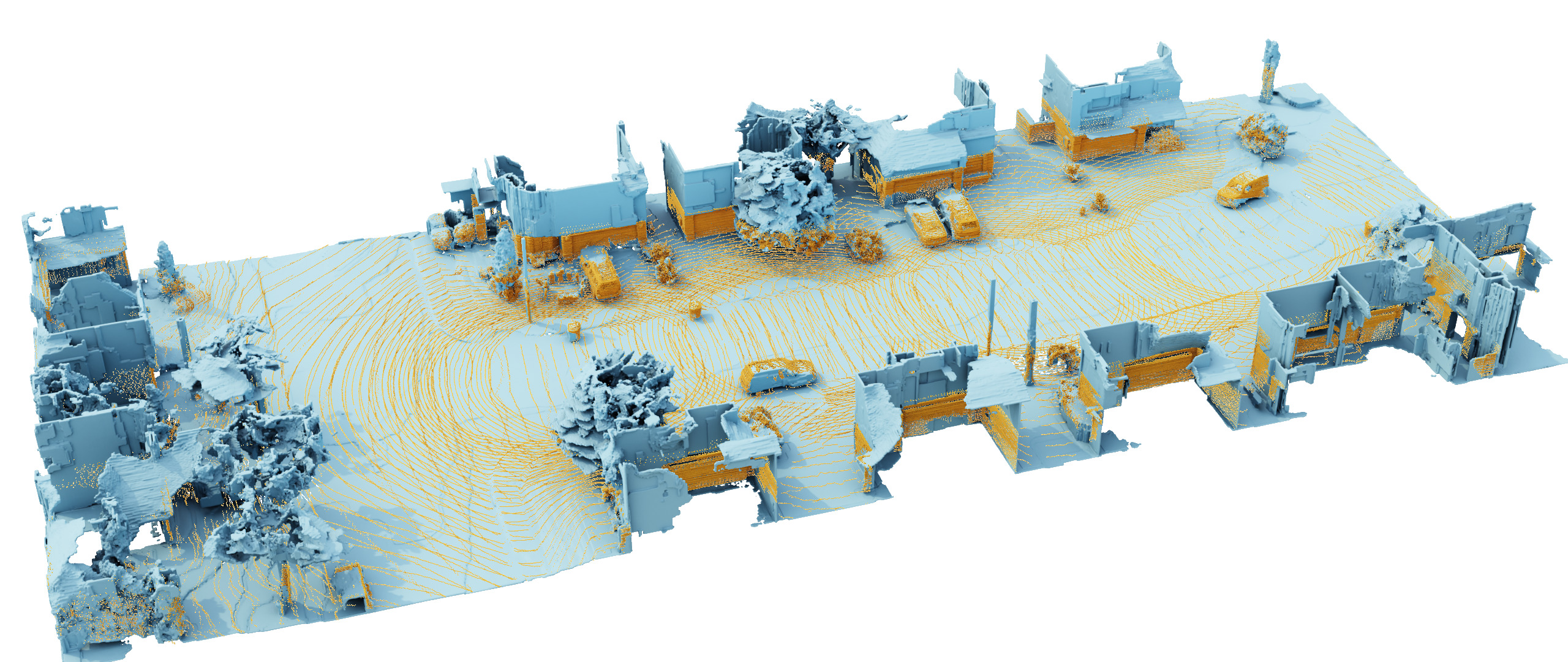}
    \includegraphics[width=\linewidth]{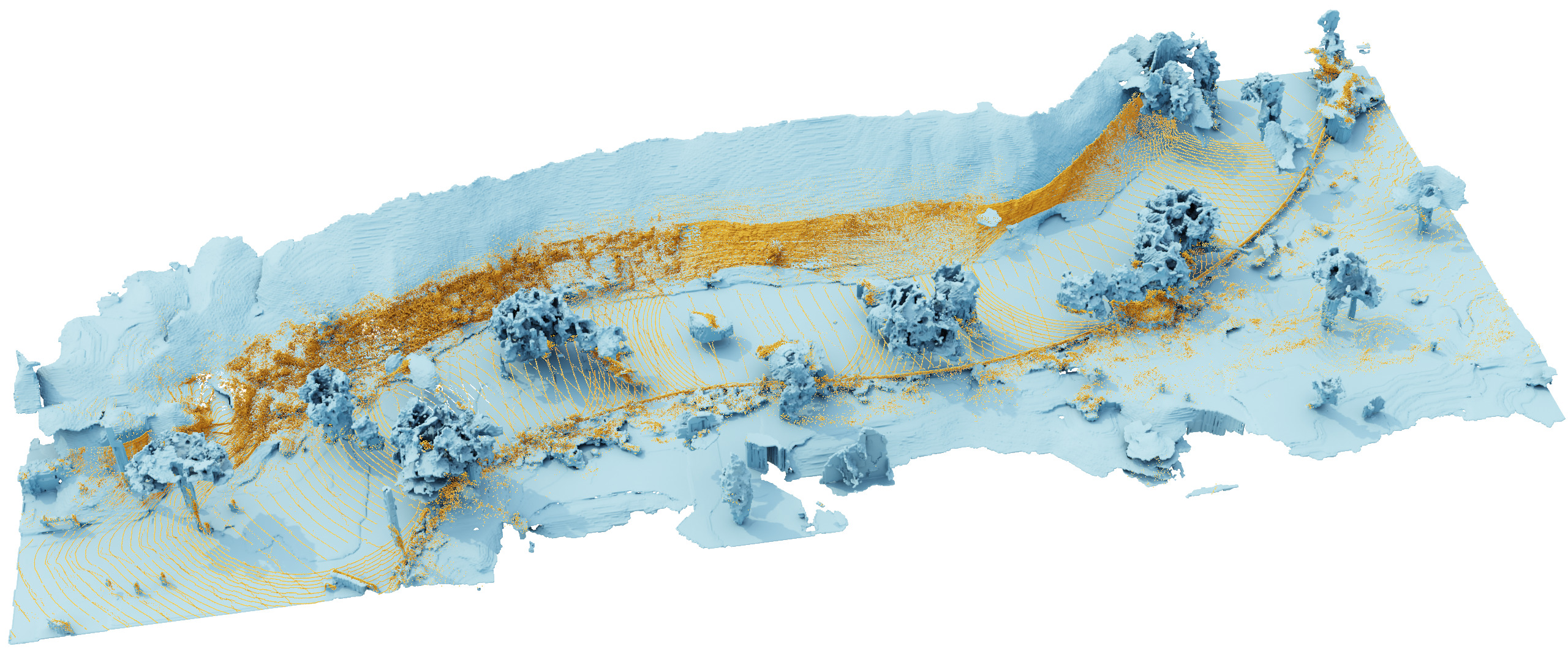}
    \vspace{-2em}
    \caption{
        Additional completion visualizations on real-world Waymo-Open dataset on 100m scenes.
        Yellow spheres indicate input.
        hGCA is spatially scalable, completing this whole scene (100 meters) at high resolution on a single 24GB GPU without additional tricks.
        hGCA can even extrapolate hills (bottom) from real-world scans.
    }
    \label{fig:waymo_big}
\end{figure*}
\def \qualwidth{0.32}

\begin{figure*}[]
    \centering
    \begin{subfigure}{\linewidth}
        \rotatebox[origin=lB]{90}{\large{\;\;\;\;\;\;\;\;Input}}
        \includegraphics[width=\qualwidth\linewidth]{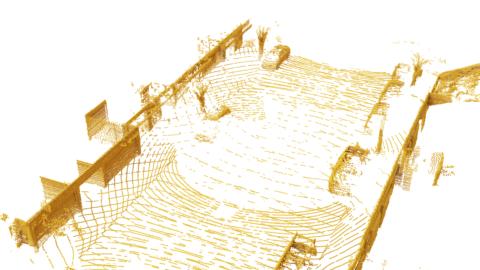}
        \hfill
        \includegraphics[width=\qualwidth\linewidth]{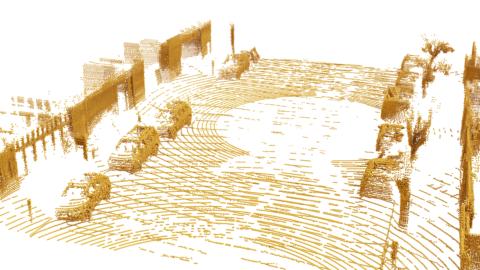}
        \hfill
        \includegraphics[width=\qualwidth\linewidth]{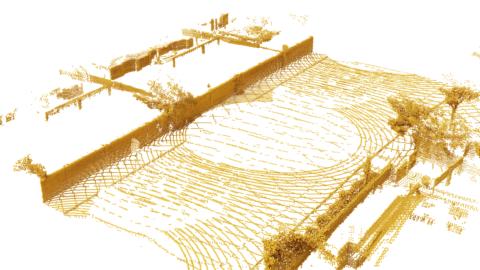}
        \hfill
        \\
        \rotatebox[origin=lB]{90}{\large{\;\;\;Acc. scans}}
        \includegraphics[width=\qualwidth\linewidth]{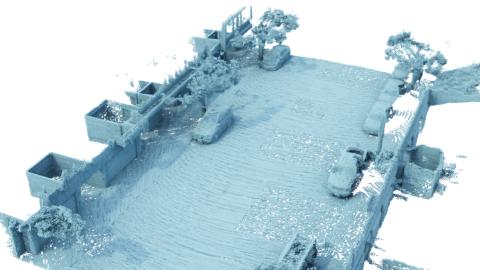}
        \hfill
        \includegraphics[width=\qualwidth\linewidth]{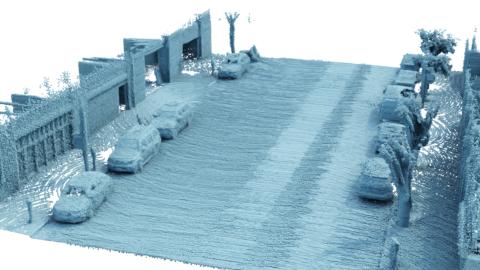}
        \hfill
        \includegraphics[width=\qualwidth\linewidth]{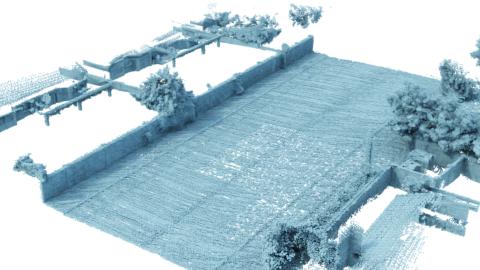}
        \hfill
        \\
        \rotatebox[origin=lB]{90}{\large{\;\;\;\;\;\;\;SCPNet}}
        \includegraphics[width=\qualwidth\linewidth]{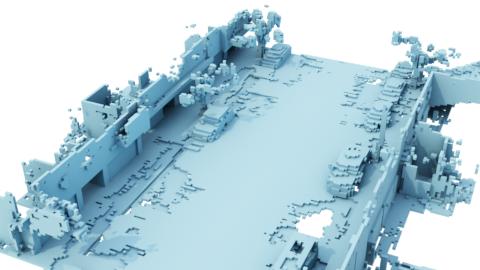}
        \hfill
        \includegraphics[width=\qualwidth\linewidth]{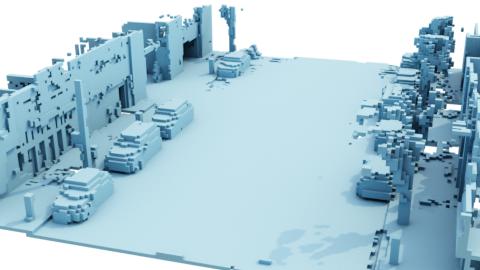}
        \hfill
        \includegraphics[width=\qualwidth\linewidth]{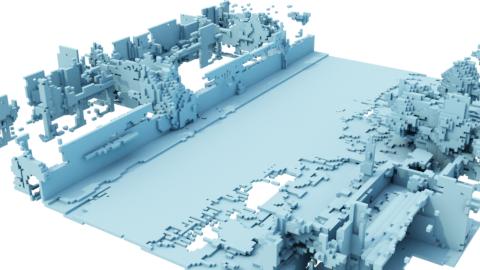}
        \hfill
        \\
        \rotatebox[origin=lB]{90}{\large{\;\;\;\;\;\;\;SG-NN}}
        \includegraphics[width=\qualwidth\linewidth]{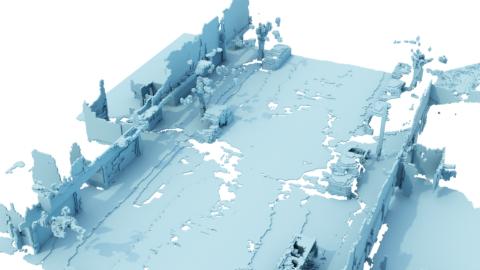}
        \hfill
        \includegraphics[width=\qualwidth\linewidth]{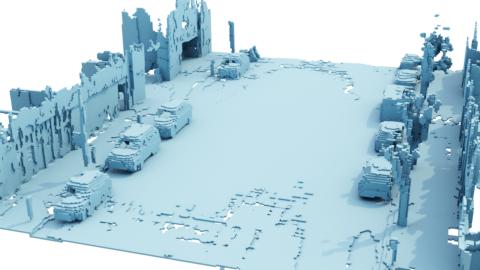}
        \hfill
        \includegraphics[width=\qualwidth\linewidth]{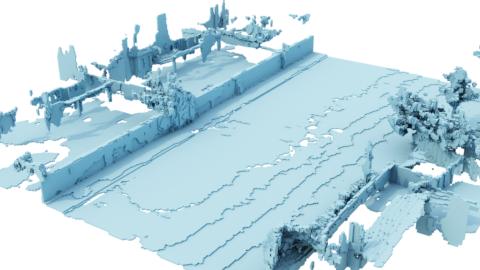}
        \hfill
        \\
        \rotatebox[origin=lB]{90}{\large{\;\;\;\;GCA }\small{(20cm)}}
        \includegraphics[width=\qualwidth\linewidth]{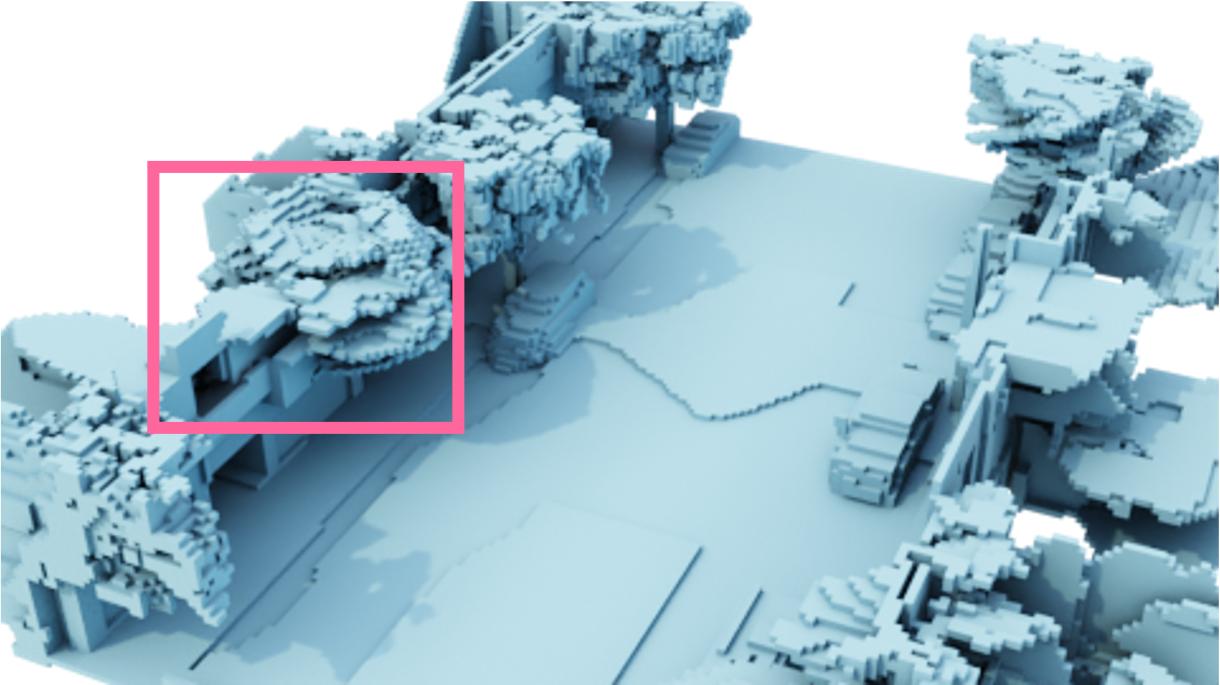}
        \hfill
        \includegraphics[width=\qualwidth\linewidth]{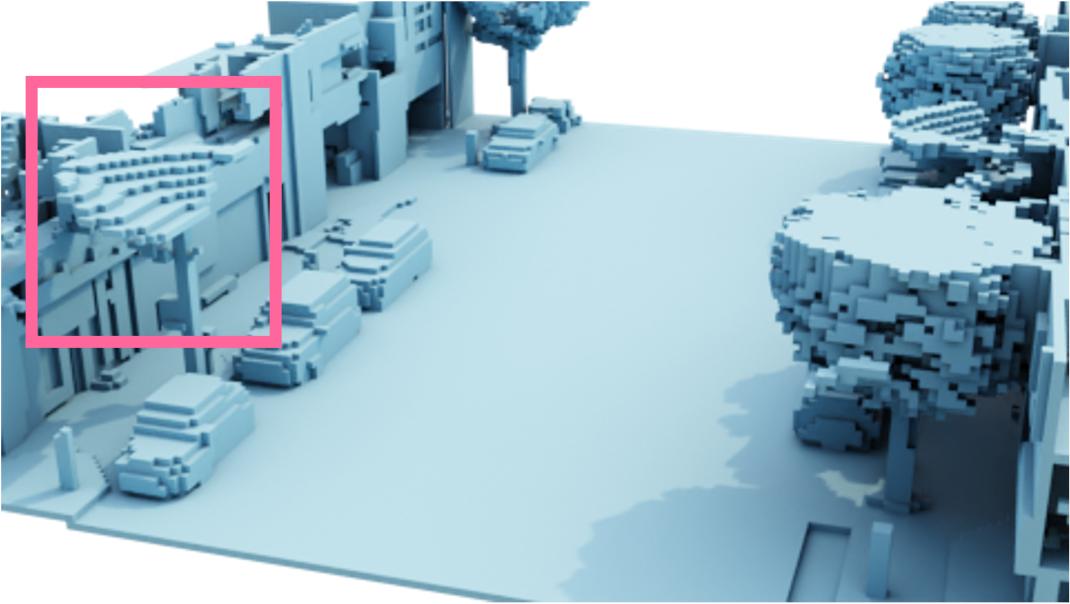}
        \hfill
        \includegraphics[width=\qualwidth\linewidth]{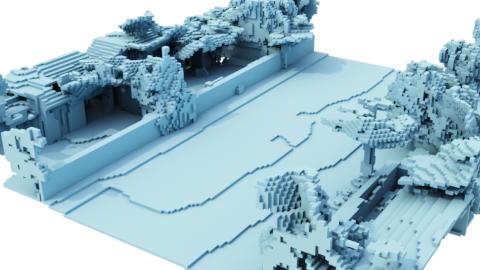}
        \hfill
        \\
        \rotatebox[origin=lB]{90}{\large{\;\;\;\;\;\;\;\;hGCA}}
        \includegraphics[width=\qualwidth\linewidth]{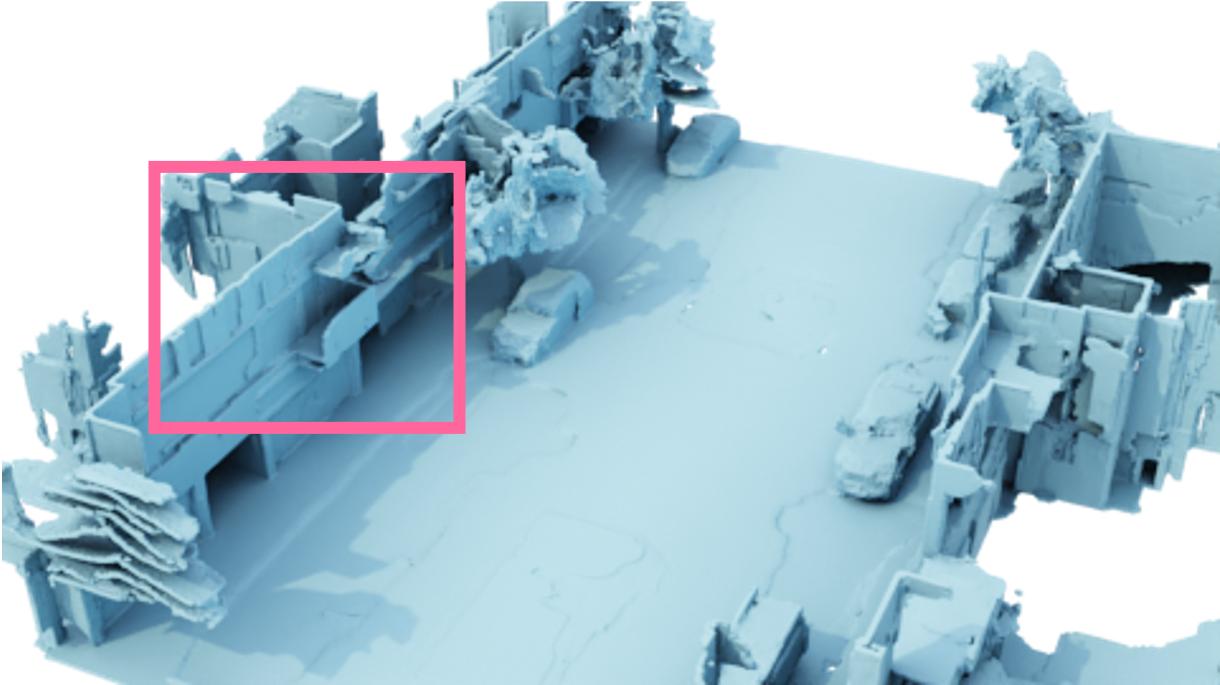}
        \hfill
        \includegraphics[width=\qualwidth\linewidth]{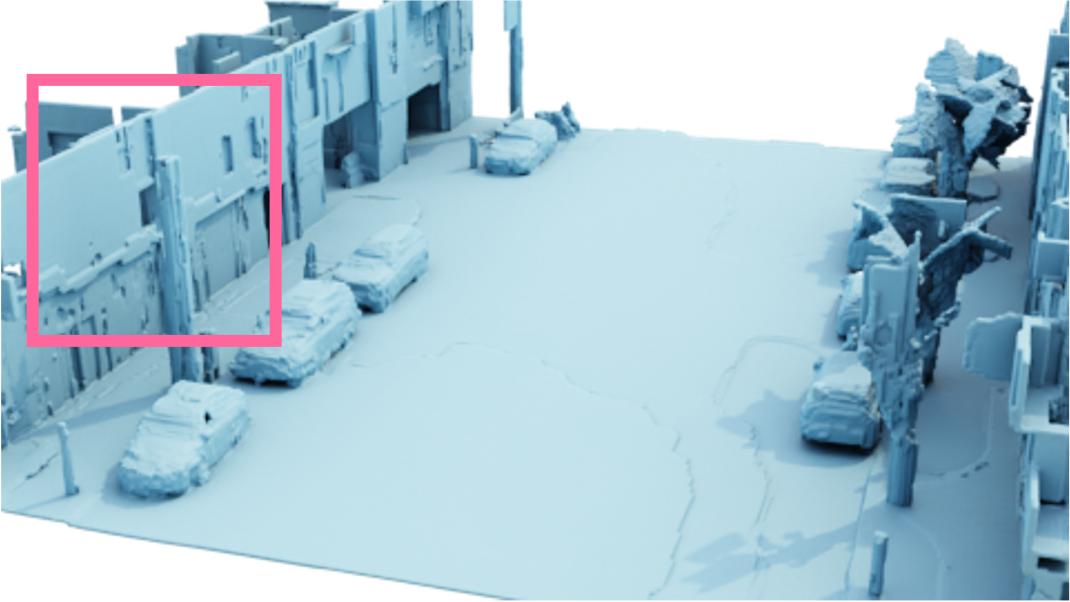}
        \hfill
        \includegraphics[width=\qualwidth\linewidth]{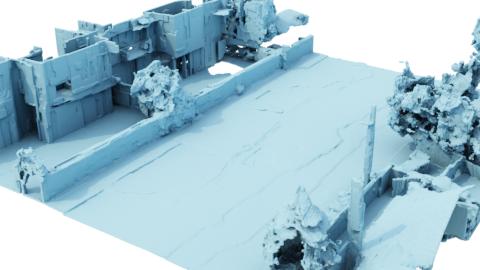}
        \hfill
    \end{subfigure}
    \caption{
        Additional visualizations on real-world Waymo-open dataset.
        hGCA exhibits great sim-to-real performance, while naive GCA suffers from inconsistency (pink).
        \vspace{5em}
    }
    \label{fig:qual_real_1}
\end{figure*}

\begin{figure*}[]
    \centering
    \begin{subfigure}{\linewidth}
        \rotatebox[origin=lB]{90}{\large{\;\;\;\;\;\;\;\;Input}}
        \includegraphics[width=\qualwidth\linewidth]{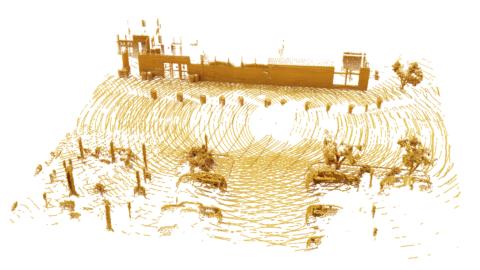}
        \hfill
        \includegraphics[width=\qualwidth\linewidth]{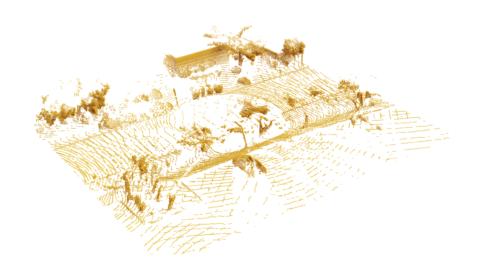}
        \hfill
        \includegraphics[width=\qualwidth\linewidth]{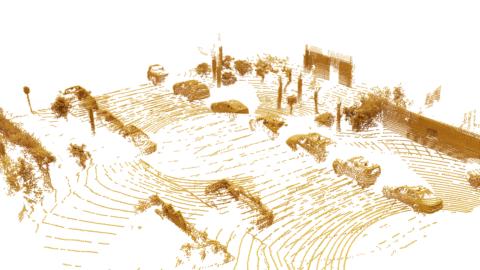}
        \hfill
        \\
        \rotatebox[origin=lB]{90}{\large{\;\;\;Acc. scans}}
        \includegraphics[width=\qualwidth\linewidth]{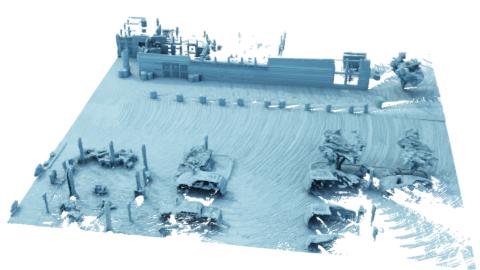}
        \hfill
        \includegraphics[width=\qualwidth\linewidth]{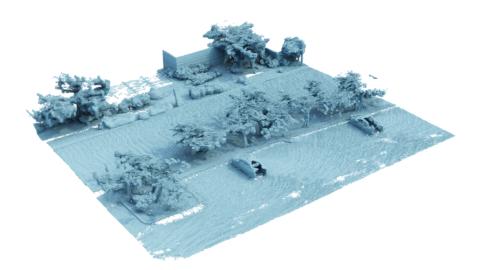}
        \hfill
        \includegraphics[width=\qualwidth\linewidth]{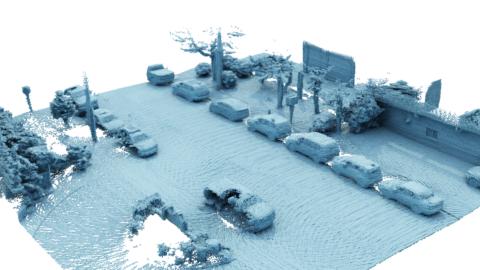}
        \hfill
        \\
        \rotatebox[origin=lB]{90}{\large{\;\;\;\;\;\;\;SCPNet}}
        \includegraphics[width=\qualwidth\linewidth]{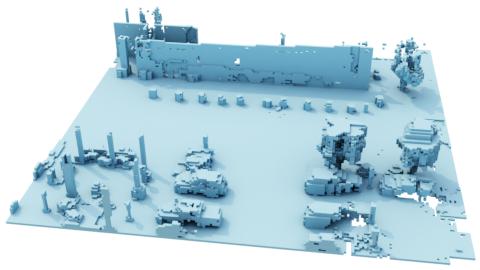}
        \hfill
        \includegraphics[width=\qualwidth\linewidth]{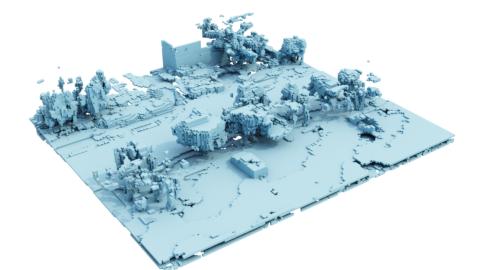}
        \hfill
        \includegraphics[width=\qualwidth\linewidth]{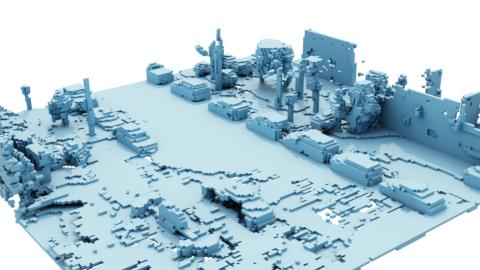}
        \hfill
        \\
        \rotatebox[origin=lB]{90}{\large{\;\;\;\;\;\;\;SG-NN}}
        \includegraphics[width=\qualwidth\linewidth]{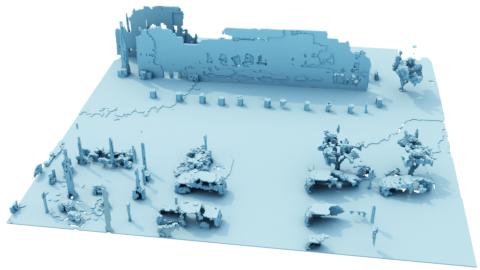}
        \hfill
        \includegraphics[width=\qualwidth\linewidth]{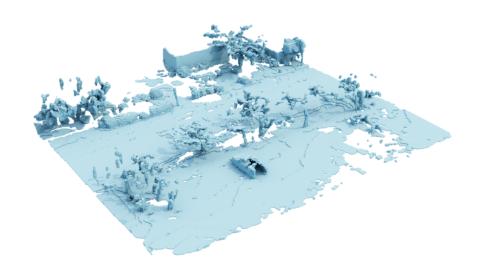}
        \hfill
        \includegraphics[width=\qualwidth\linewidth]{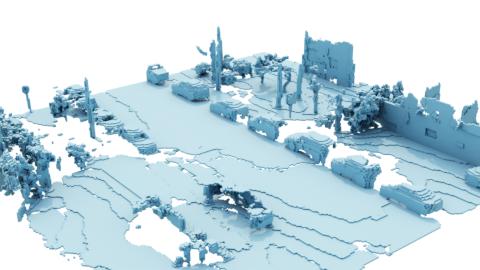}
        \hfill
        \\
        \rotatebox[origin=lB]{90}{\large{\;\;\;\;GCA }\small{(20cm)}}
        \includegraphics[width=\qualwidth\linewidth]{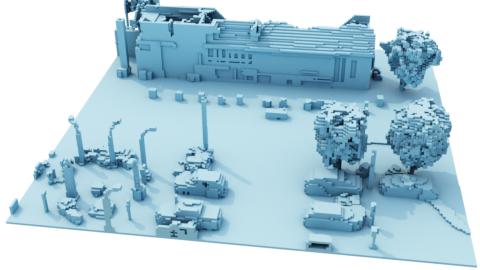}
        \hfill
        \includegraphics[width=\qualwidth\linewidth]{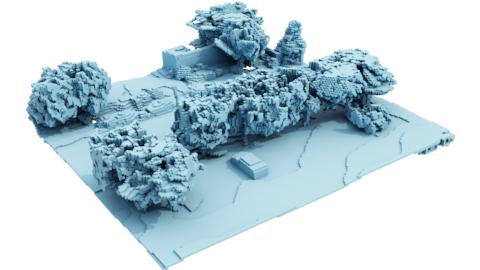}
        \hfill
        \includegraphics[width=\qualwidth\linewidth]{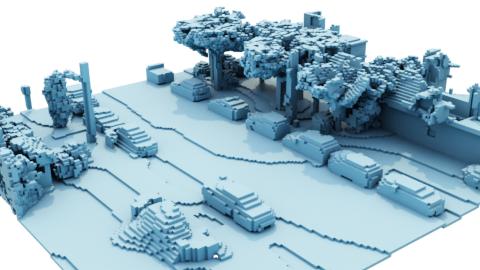}
        \hfill
        \\
        \rotatebox[origin=lB]{90}{\large{\;\;\;\;\;\;\;\;hGCA}}
        \includegraphics[width=\qualwidth\linewidth]{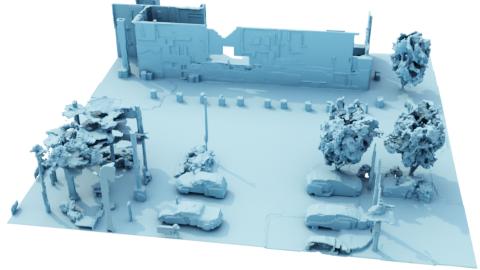}
        \hfill
        \includegraphics[width=\qualwidth\linewidth]{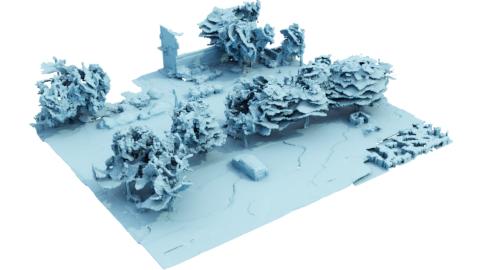}
        \hfill
        \includegraphics[width=\qualwidth\linewidth]{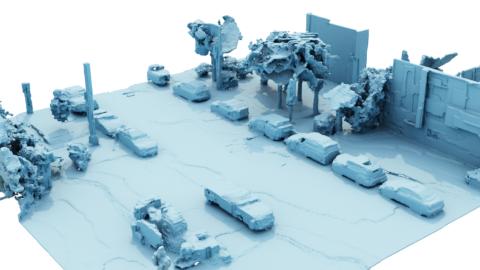}
        \hfill
    \end{subfigure}
    \caption{
        Additional visualizations on real-world Waymo-open dataset.
        \vspace{5em}
    }
    \label{fig:qual_real_2}
\end{figure*}

\begin{figure*}[]
    \centering
    \begin{subfigure}{\linewidth}
        \rotatebox[origin=lB]{90}{\large{\;\;\;\;\;\;\;\;Input}}
        \includegraphics[width=\qualwidth\linewidth]{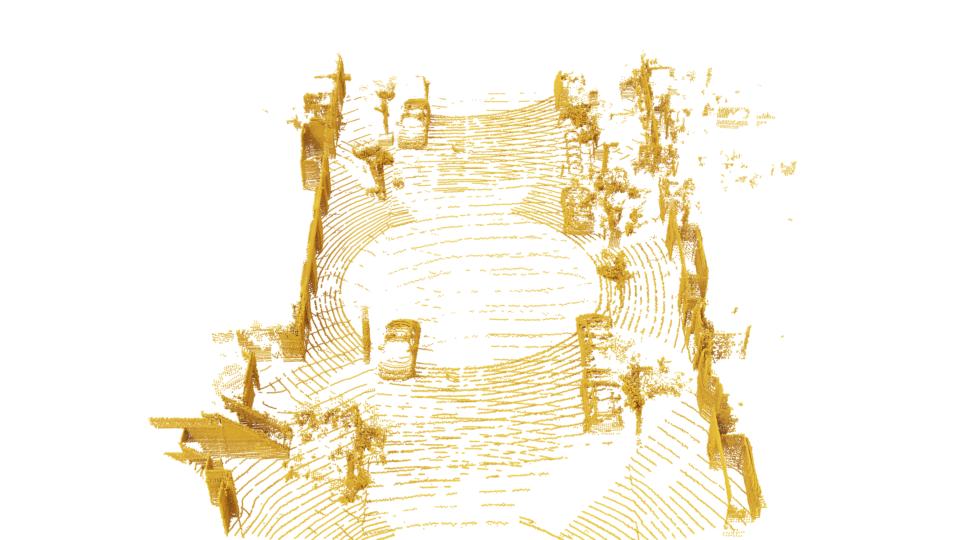}
        \hfill
        \includegraphics[width=\qualwidth\linewidth]{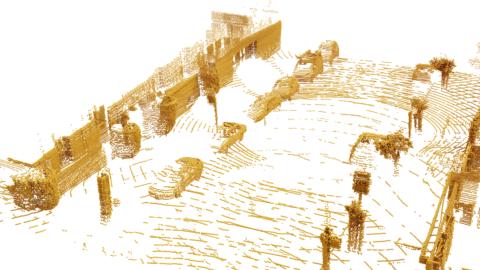}
        \hfill
        \includegraphics[width=\qualwidth\linewidth]{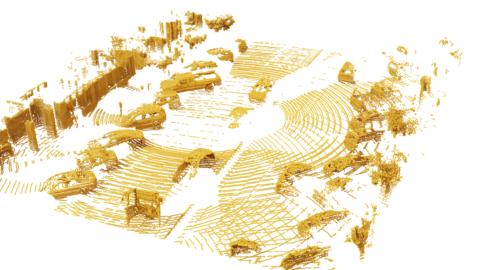}
        \hfill
        \\
        \rotatebox[origin=lB]{90}{\large{\;\;\;Acc. scans}}
        \includegraphics[width=\qualwidth\linewidth]{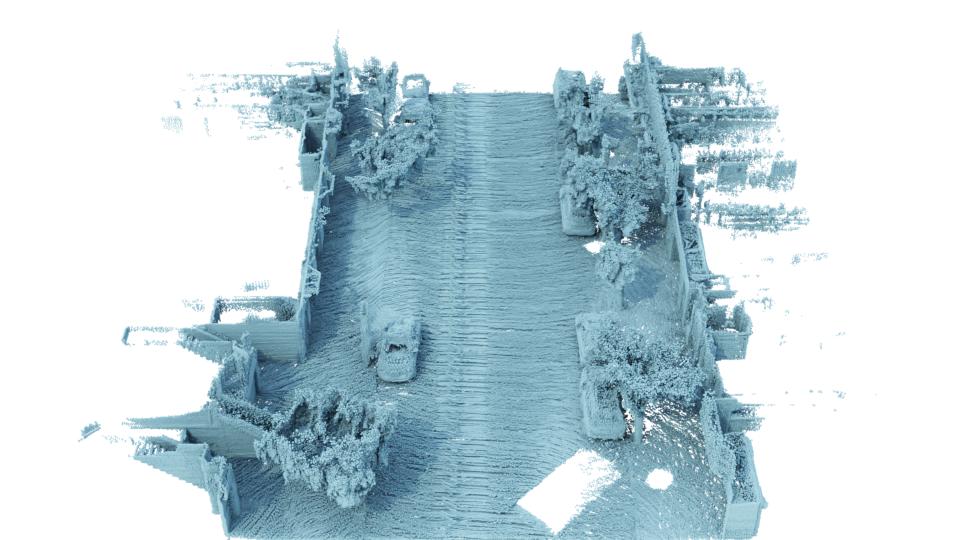}
        \hfill
        \includegraphics[width=\qualwidth\linewidth]{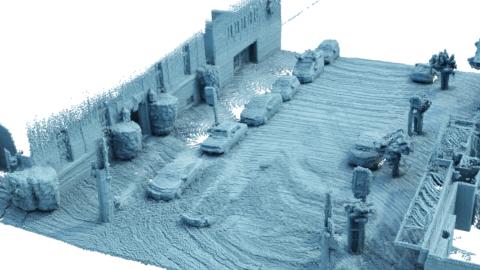}
        \hfill
        \includegraphics[width=\qualwidth\linewidth]{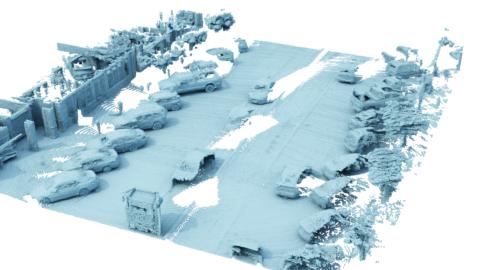}
        \hfill
        \\
        \rotatebox[origin=lB]{90}{\large{\;\;\;\;\;\;\;SCPNet}}
        \includegraphics[width=\qualwidth\linewidth]{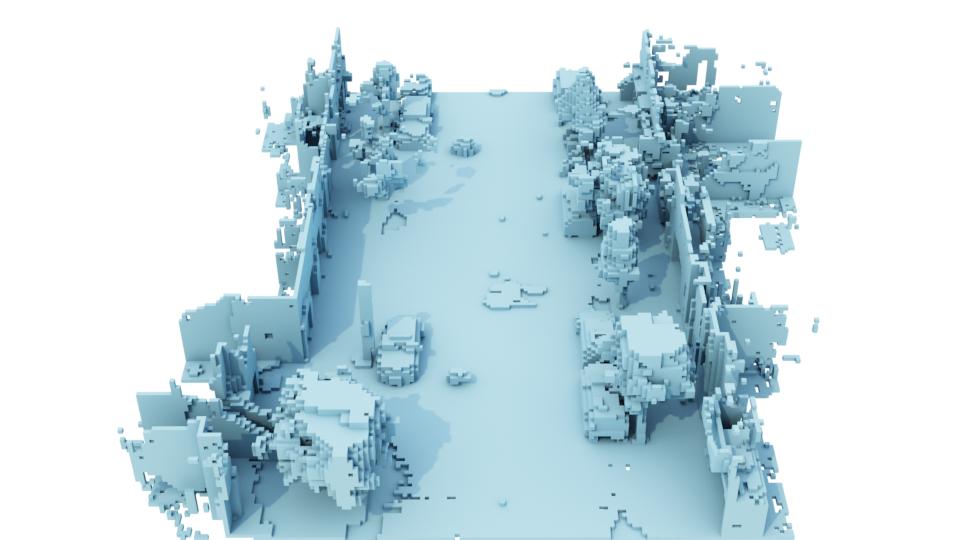}
        \hfill
        \includegraphics[width=\qualwidth\linewidth]{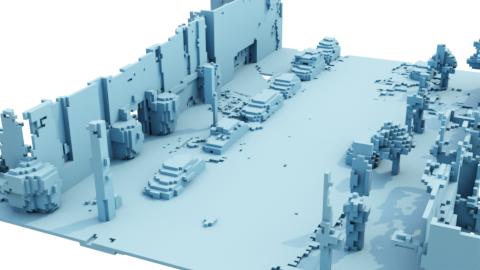}
        \hfill
        \includegraphics[width=\qualwidth\linewidth]{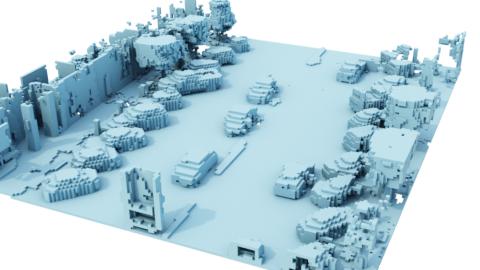}
        \hfill
        \\
        \rotatebox[origin=lB]{90}{\large{\;\;\;\;\;\;\;SG-NN}}
        \includegraphics[width=\qualwidth\linewidth]{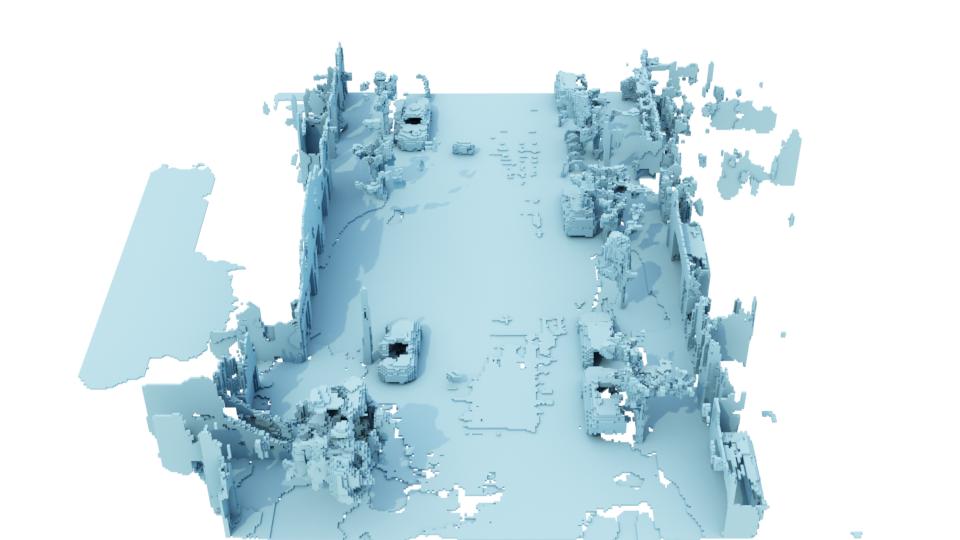}
        \hfill
        \includegraphics[width=\qualwidth\linewidth]{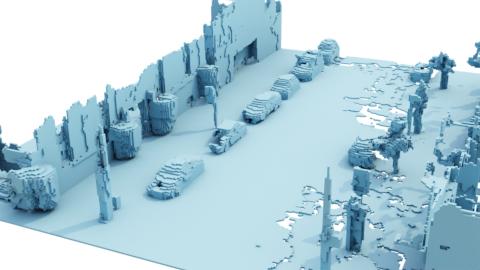}
        \hfill
        \includegraphics[width=\qualwidth\linewidth]{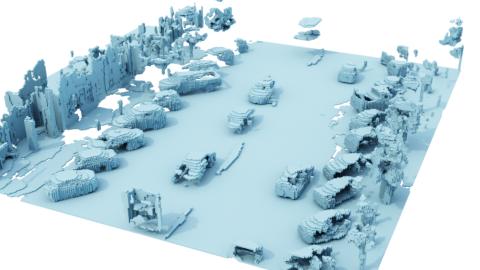}
        \hfill
        \\
        \rotatebox[origin=lB]{90}{\large{\;\;\;\;GCA }\small{(20cm)}}
        \includegraphics[width=\qualwidth\linewidth]{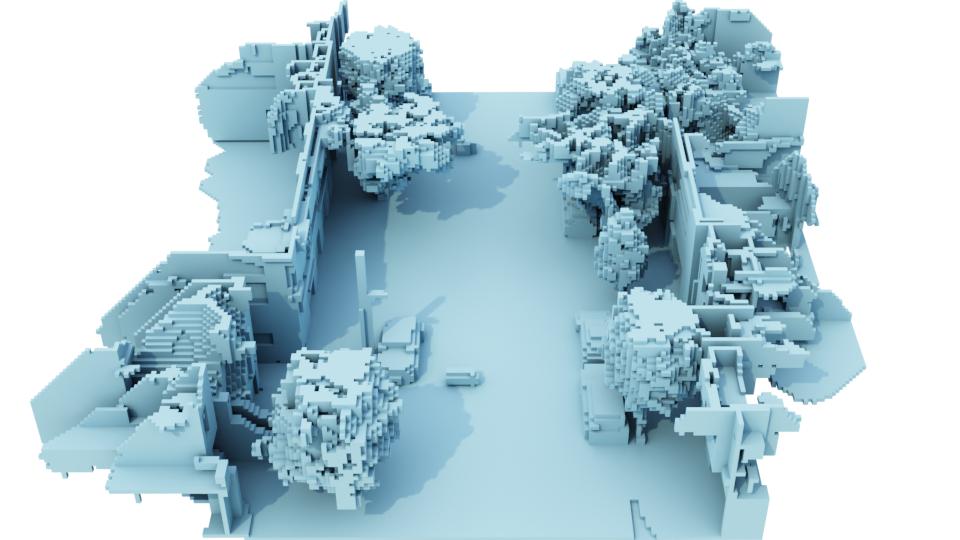}
        \hfill
        \includegraphics[width=\qualwidth\linewidth]{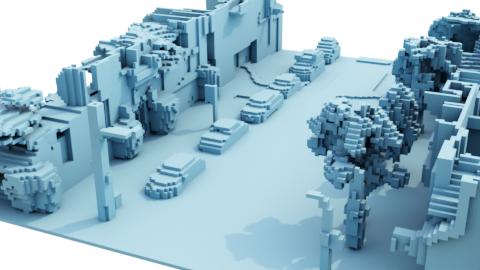}
        \hfill
        \includegraphics[width=\qualwidth\linewidth]{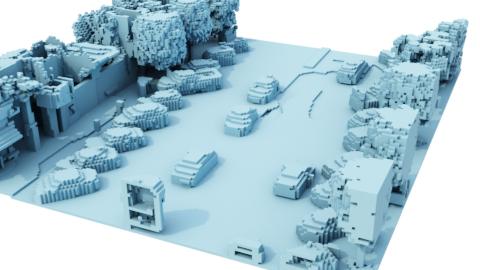}
        \hfill
        \\
        \rotatebox[origin=lB]{90}{\large{\;\;\;\;\;\;\;\;hGCA}}
        \includegraphics[width=\qualwidth\linewidth]{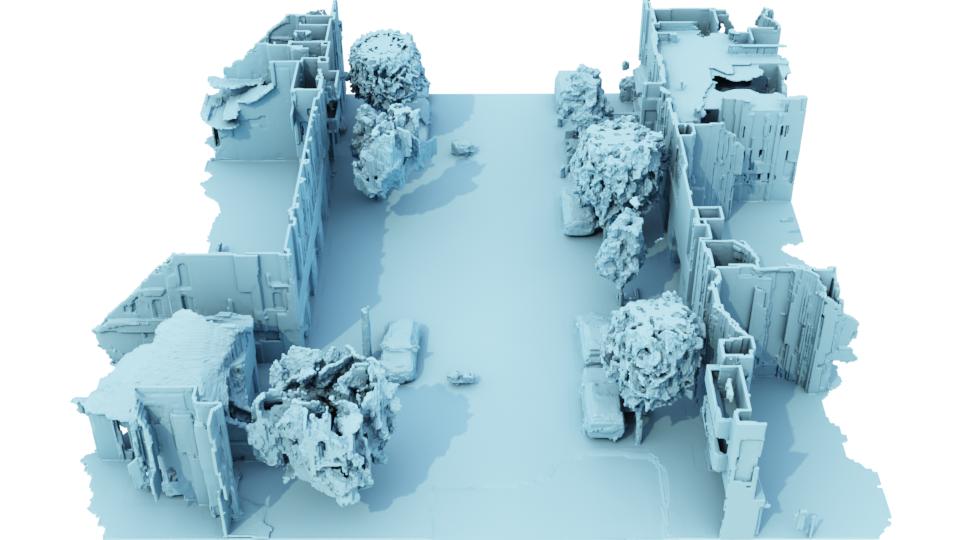}
        \hfill
        \includegraphics[width=\qualwidth\linewidth]{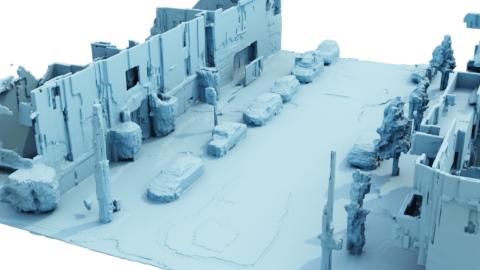}
        \hfill
        \includegraphics[width=\qualwidth\linewidth]{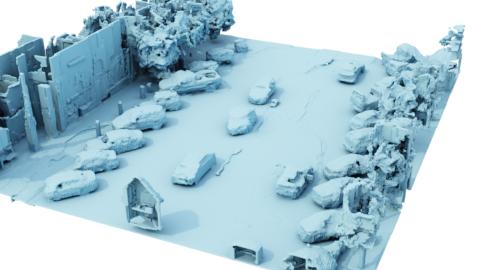}
        \hfill
    \end{subfigure}
    \caption{
        Additional visualizations on real-world Waymo-open dataset.
        \vspace{5em}
    }
    \label{fig:qual_real_3}
\end{figure*}

We first provide further evaluations on the sim-to-real performance of hGCA using Waymo-Open~\cite{Waymo-Open} dataset.
As stated in the main manuscript, abundant real-world AV data suffers from various noises and limited measurement ranges.
Most importantly, we do not have the ground truth shapes for our task of shape extrapolation, which challenges systematic analysis.
This section provides sub-optimal quantitative measures and comprehensive qualitative results to demonstrate that we can faithfully generate realistic scenes given partial and noisy real-world measurements.

As a means for quantitative evaluation, we use the accumulated scans as a pseudo ground truth as in semantic scene completion works~\cite{SemanticKITTI, JS3CNet, scpnet} and analyze its performance with LiDAR ReSim and IoU as demonstrated in our results of synthetic datasets.
We randomly chose 202 scenes and used five scans as input.
Then, the generated scene is compared against accumulated data using all scans.
While the accumulated scans are denser variations of the given measurement, they are noisy and highly incomplete measurements in a confined height range, as shown in Fig.~\ref{fig:waymo_big},~\ref{fig:qual_real_1},~\ref{fig:qual_real_2} and~\ref{fig:qual_real_3}.
We devise the evaluation metrics to adapt to the limitation of the reference data.
When computing the LiDAR ReSim score, we simulate LiDAR at the same height as the \emph{input} LiDAR scan instead of using a higher elevation LiDAR.
Likewise, we compute IoU only on regions visible from LiDAR scans as evaluated in semantic scene completion~\cite{SemanticKITTI, JS3CNet, scpnet}.
Therefore, neither of these metrics assesses the performance of extrapolation beyond the LiDAR height range and occlusion.


\begin{table}[t]
    \centering
    \resizebox{0.9\linewidth}{!}{
        \begin{tabular}{l|c|cc|c|c}
            \toprule
            \multirow{2}{*}{Method} & \multirow{2}{*}{\shortstack[l]{Represe\\-ntation}} & \multicolumn{3}{c|}{LiDAR ReSim} & \multirow{2}{*}{IoU} \\
            & & min. $\downarrow$ & avg. $\downarrow$ & TMD $\uparrow$ & \\
            \midrule
            SCPNet & 20cm & \dc{5.57} & - & 52.33 \\
            SG-NN & 10cm & \dc{5.81} & - & 49.83 \\
            GCA & 20cm & 5.73 & 6.04 & \textbf{1.07} & 51.91 \\
            GCA + Planner & 20cm & 5.48 & 5.57 & 0.70 & 52.26 \\
            \midrule
            \multirow{2}{*}{hGCA} & 10cm & 4.65 & 4.73 & 0.77 & 52.25 \\
            & implicit & \textbf{4.52} & 4.50 & 0.97 & \textbf{56.50} \\
            \bottomrule
        \end{tabular}%
        }
        \caption{
            Quantitative results on Waymo with 5 scans given as input.
            All results except IoU are multiplied by 10 in meter scale.
            LiDAR Resim evaluates the fidelity of completion and TMD measures the diversity of generation.
            Unlike synthetic results, LiDAR ReSim uses same elevation angle as the input and IoU is computed with accumulated scans.
    }
    \label{table:waymo}
   
\end{table}

Table~\ref{table:waymo} reports the performance on the shape completion within the measurement range, only trained with limited synthetic content.
Similar to synthetic results, hGCA outperforms all baselines in LiDAR ReSim and IoU by a margin.
The results indicate that our completion is closer to the dense measurements of real-world geometry than existing methods.
Interestingly, the IoU of the continuous completion outperforms the initial voxel occupancy (10cm) in the real-world analysis in Table~\ref{table:waymo} whereas the voxel occupancy achieves higher IoU values in our synthetic experiments (Table 1 of the main paper).
Recall that our high-resolution upsampling sometimes fails to create a narrow structure, as unsigned distance fields may obscure the exact zero-level location in high-frequency details.
We observe that the accumulated scans also experience similar ambiguity due to the inherent noise in the real-world measurement.
As the noisy scans serve as the ground truth, the possible misalignment due to the thick implicit generation may be evaluated as a faithful generation.
Such a phenomenon again shows difficulties in the quantitative evaluation of generative models on real-world datasets.

We further visualize various generation results observed from diverse viewpoints in Fig.~\ref{fig:waymo_big},~\ref{fig:qual_real_1},~\ref{fig:qual_real_2} and~\ref{fig:qual_real_3}.
Despite the limited measurement range and noisy input, our method can extrapolate the input measurement into large-scale real-world scenes in a scalable way.
Figure~\ref{fig:qual_real_1},~\ref{fig:qual_real_2} and~\ref{fig:qual_real_3} also show comparison against other baselines.
Deterministic baselines (SCPNet~\cite{scpnet} and SG-NN~\cite{dai2020sgnn}) exhibit conservative behavior, leaving holes in the ground and partially complete buildings or trees.
Na\"ive GCA, while it is a generative baseline, suffers from inconsistent generation, which results in overlapping structures (pink boxes in Fig.~\ref{fig:qual_real_1}).
On the other hand, hGCA can generate unseen geometry while maintaining global consistency in various real-world settings. 
For example, the input shown in the right column of Fig.~\ref{fig:qual_real_1} misses a significant portion of the buildings as a fence occludes them. Nonetheless, hGCA faithfully generates detailed facades of buildings.

\subsubsection{nuScenes}
\begin{figure*}[]
    \centering
    \includegraphics[width=\linewidth]{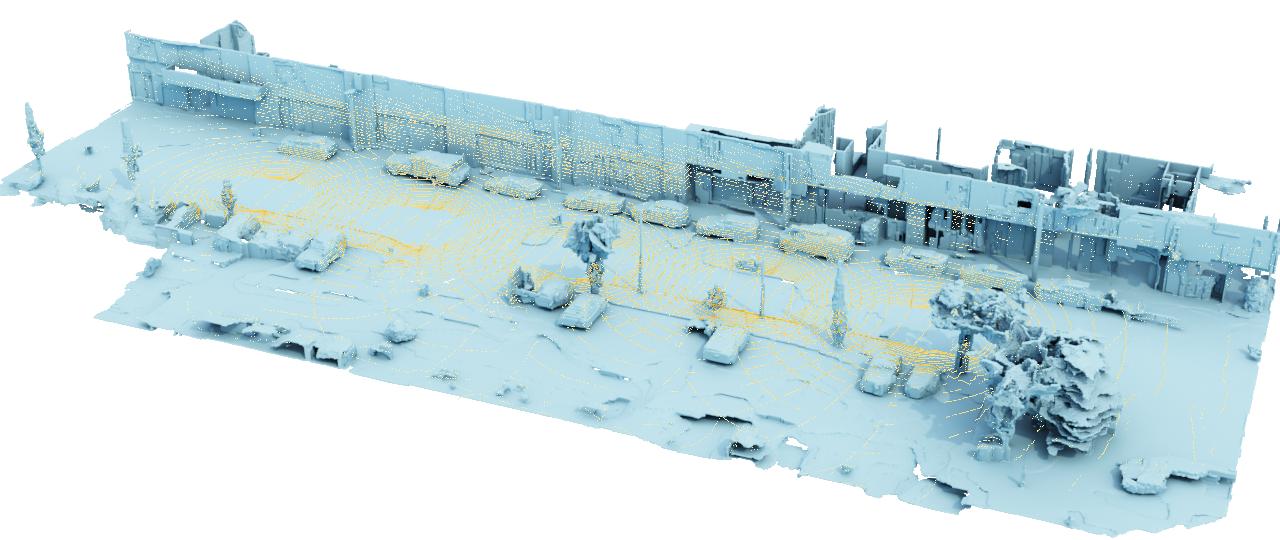}
    \includegraphics[width=\linewidth]{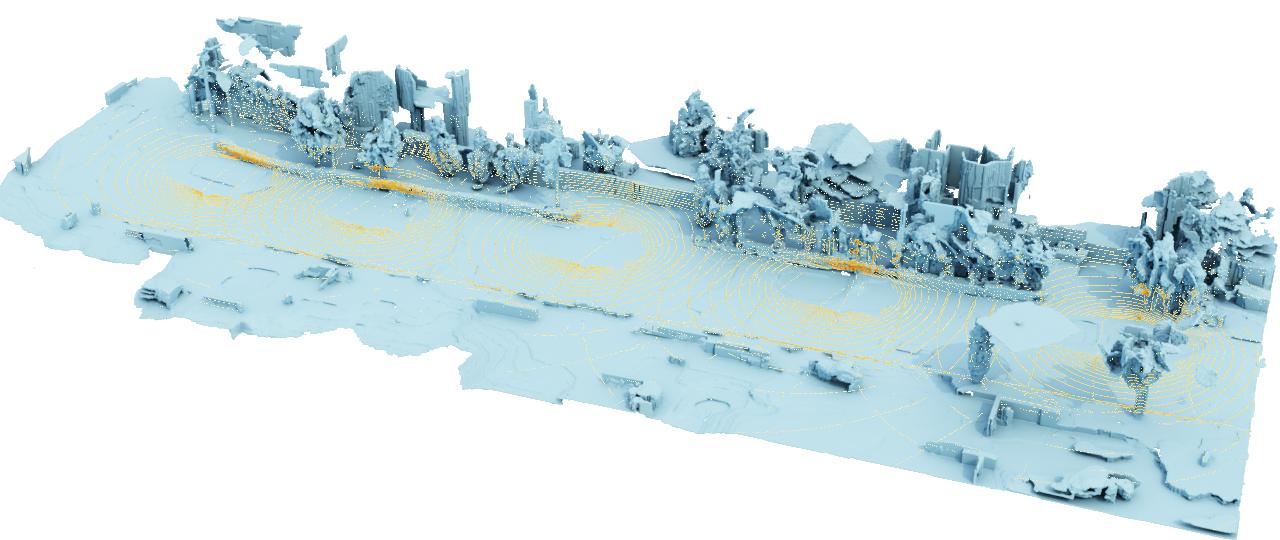}
    \vspace{-2em}
    \caption{
        Visualizations on real-world nuScenes dataset on 100m scenes.
        Yellow spheres indicate input.
        hGCA is spatially scalable, completing this whole scene (100 meters) at high resolution on a single 24GB GPU without additional tricks.
    }
    \label{fig:nuscenes}
\end{figure*}
To test generalization to a new sensor configuration, we show results on the nuScenes dataset in Fig.~\ref{fig:nuscenes}. 
While 64-beam LiDAR was used in Waymo and our synthetic training set, nuScenes scans were obtained using 32-beam LiDAR, which is a significant domain shift.
Nevertheless, hGCA generalizes to nuScenes out-of-the-box and is robust to the domain shift.
We observe some failures on nuScenes (bottom of Fig.~\ref{fig:nuscenes}), across the street, where the model hallucinates structures in extremely sparsely scanned regions.
We speculate that training with simulated 32-beam LiDAR scans as input will enhance the quality of completion by reducing the domain gap for sparse inputs.

\subsection{Generalization to Various Input Conditions}\label{app:generalization}
As partially demonstrated in sim-to-real results, our generative pipeline can robustly handle input variations not observed in the training data. 
This section provides additional quantitative evaluations of shapes generated from different input conditions.

\paragraph{Input Sparsity.}\label{app:sparsity}
\def \qualwidth{0.14}

\begin{figure*}

    \centering
    \begin{tabular}{ccccc}
       \rotatebox[origin=lB]{90}{\;\;\;Input} & \includegraphics[width=\qualwidth\linewidth]{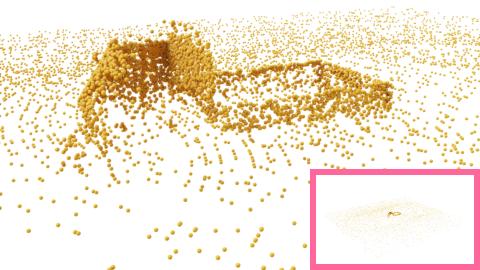} & \includegraphics[width=\qualwidth\linewidth]{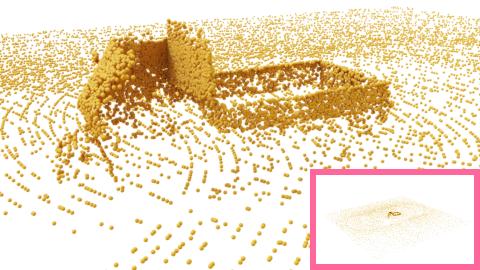} & \includegraphics[width=\qualwidth\linewidth]{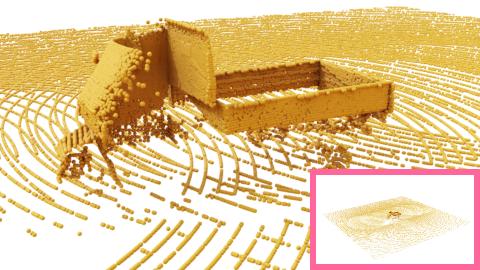} & \includegraphics[width=\qualwidth\linewidth]{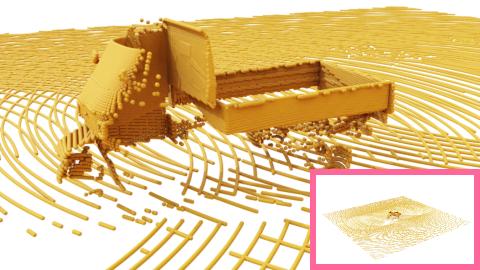} \\
       \rotatebox[origin=lB]{90}{\;\;SCPNet}& \includegraphics[width=\qualwidth\linewidth]{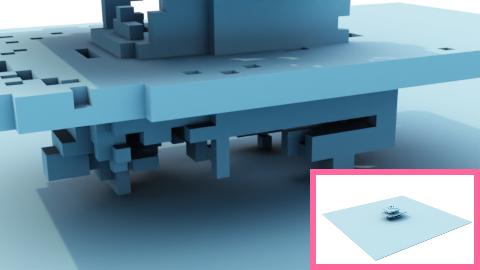} & \includegraphics[width=\qualwidth\linewidth]{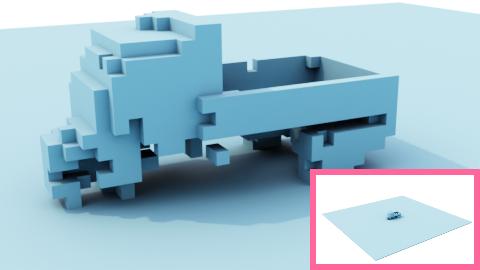} & \includegraphics[width=\qualwidth\linewidth]{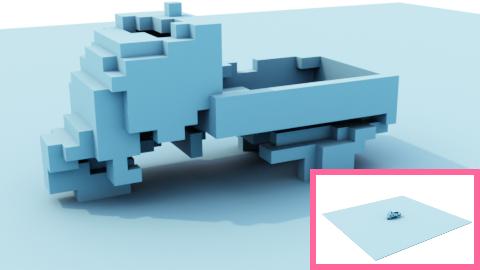} & \includegraphics[width=\qualwidth\linewidth]{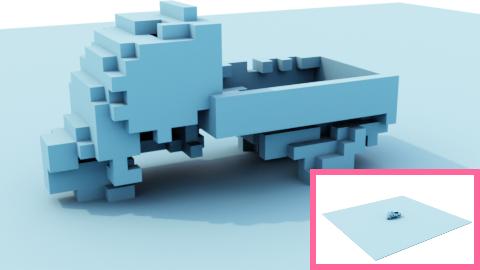} \\
       \rotatebox[origin=lB]{90}{\;JS3CNet} & \includegraphics[width=\qualwidth\linewidth]{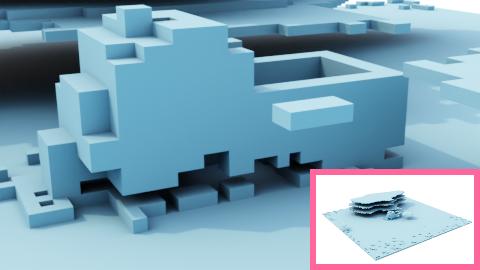} & \includegraphics[width=\qualwidth\linewidth]{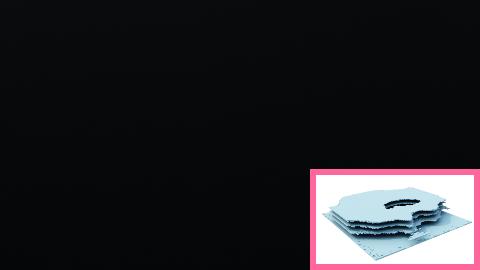} & \includegraphics[width=\qualwidth\linewidth]{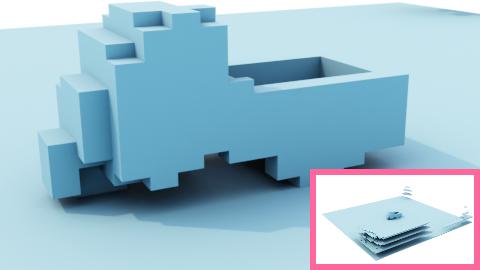} & \includegraphics[width=\qualwidth\linewidth]{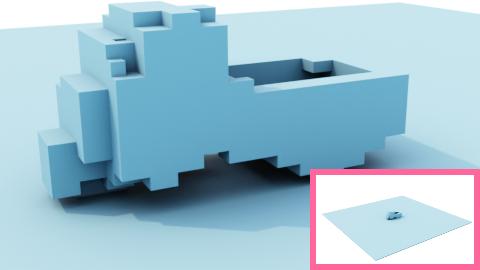} \\
       \rotatebox[origin=lB]{90}{\;\;SG-NN} & \includegraphics[width=\qualwidth\linewidth]{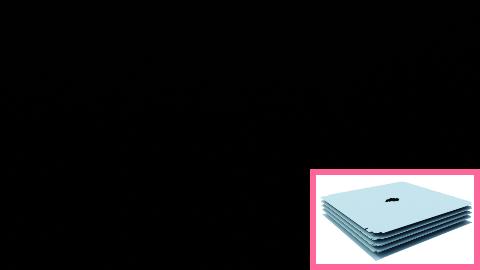} & \includegraphics[width=\qualwidth\linewidth]{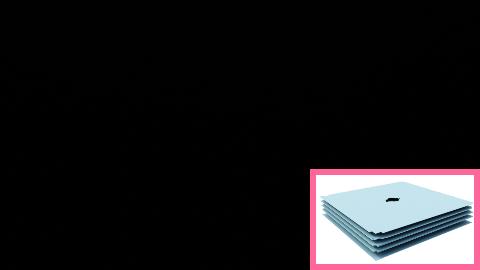} & \includegraphics[width=\qualwidth\linewidth]{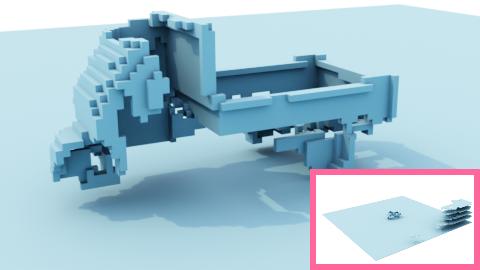} & \includegraphics[width=\qualwidth\linewidth]{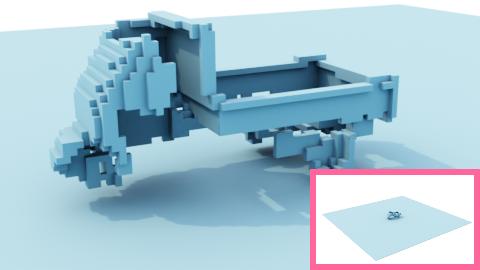} \\
       \rotatebox[origin=lB]{90}{GCA \scriptsize{($\text{20cm}$)}} & \includegraphics[width=\qualwidth\linewidth]{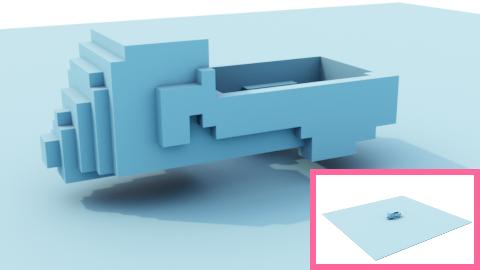} & \includegraphics[width=\qualwidth\linewidth]{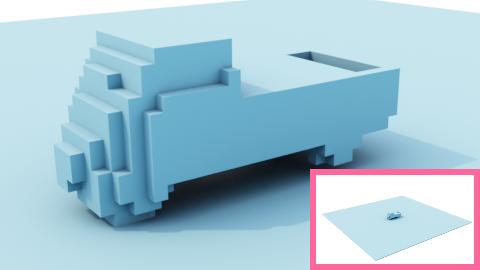} & \includegraphics[width=\qualwidth\linewidth]{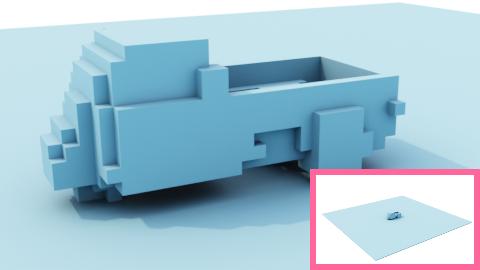} & \includegraphics[width=\qualwidth\linewidth]{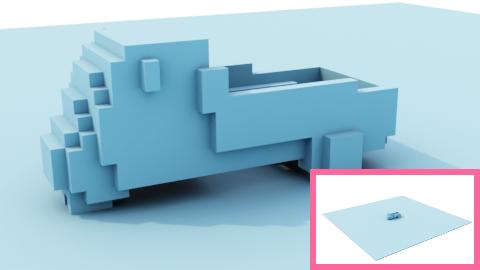} \\
       \rotatebox[origin=lB]{90}{\;\;\;hGCA}& \includegraphics[width=\qualwidth\linewidth]{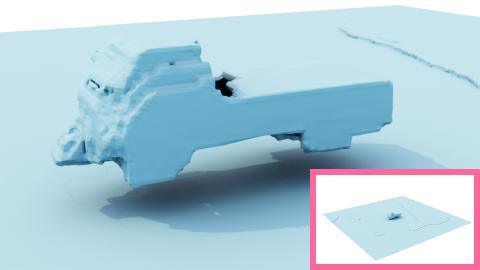} & \includegraphics[width=\qualwidth\linewidth]{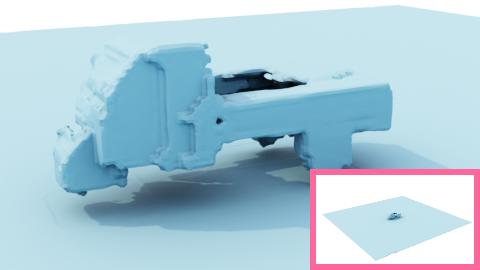} & \includegraphics[width=\qualwidth\linewidth]{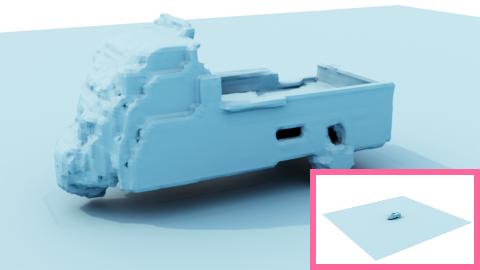} & \includegraphics[width=\qualwidth\linewidth]{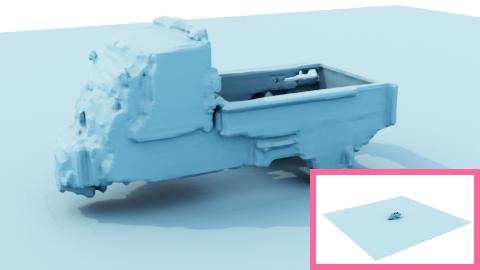} \\
       & 5\% Input & 10\% Input & 50\% Input & 100\% Input \\
    \end{tabular}
    \caption{
        Ablation study on a novel three-wheeler completion by varying density of 5 scans.
        Inset shows wide range view of the completion.
        Locality of GCA's enable generalization to sparse input producing stable completions, while method that only utilize global features fail.
    }
    \label{fig:density_scene}
\end{figure*}

\begin{figure*}
    \centering
    \begin{tabular}{ccccc}
       \rotatebox[origin=lB]{90}{\;\;\;Input} & \includegraphics[width=\qualwidth\linewidth]{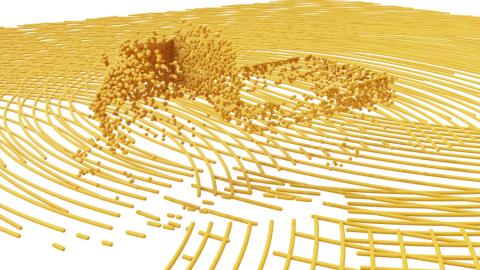} & \includegraphics[width=\qualwidth\linewidth]{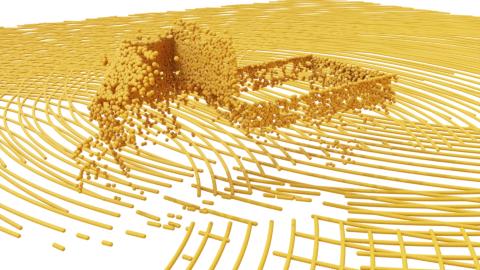} & \includegraphics[width=\qualwidth\linewidth]{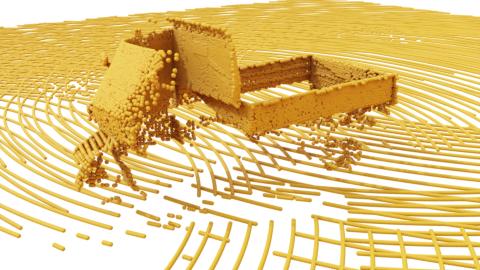} & \includegraphics[width=\qualwidth\linewidth]{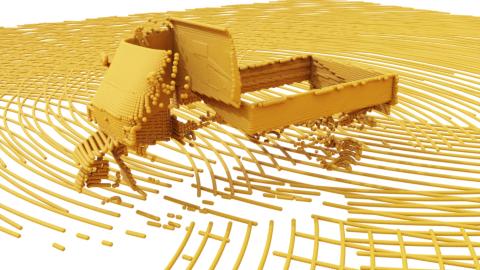} \\
       \rotatebox[origin=lB]{90}{\;\;SCPNet}& \includegraphics[width=\qualwidth\linewidth]{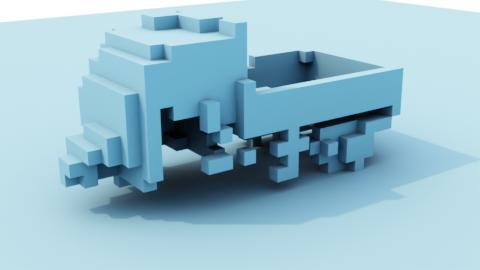} & \includegraphics[width=\qualwidth\linewidth]{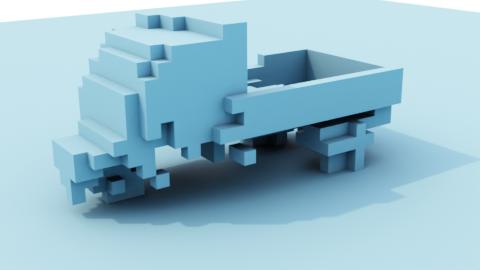} & \includegraphics[width=\qualwidth\linewidth]{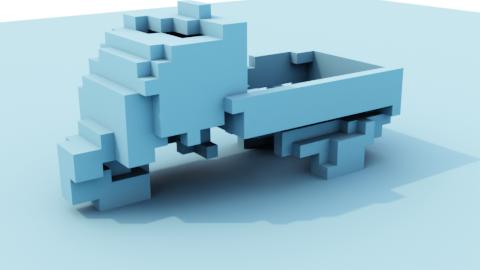} & \includegraphics[width=\qualwidth\linewidth]{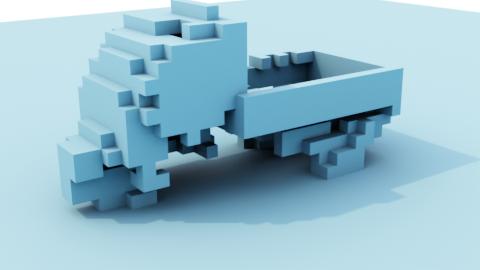} \\
       \rotatebox[origin=lB]{90}{\;JS3CNet} & \includegraphics[width=\qualwidth\linewidth]{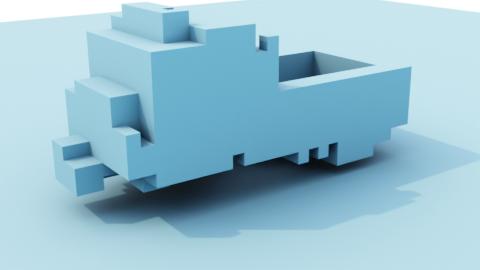} & \includegraphics[width=\qualwidth\linewidth]{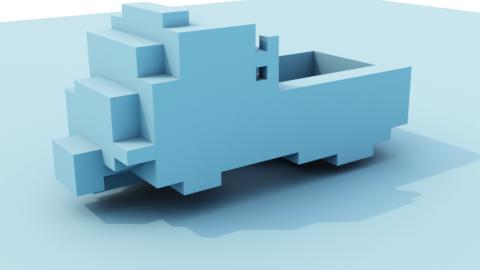} & \includegraphics[width=\qualwidth\linewidth]{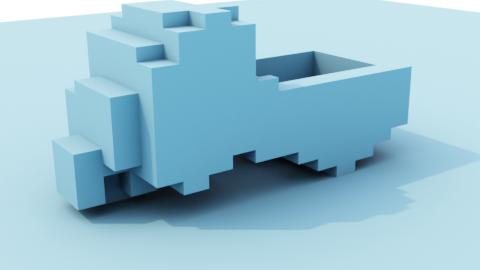} & \includegraphics[width=\qualwidth\linewidth]{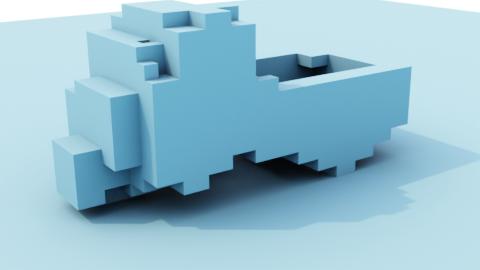} \\
       \rotatebox[origin=lB]{90}{\;\;SG-NN} & \includegraphics[width=\qualwidth\linewidth]{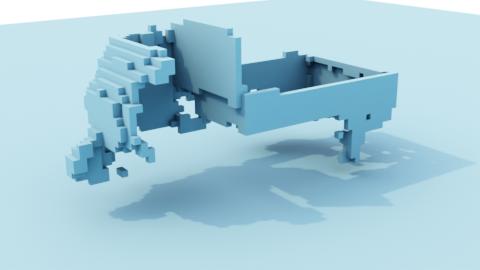} & \includegraphics[width=\qualwidth\linewidth]{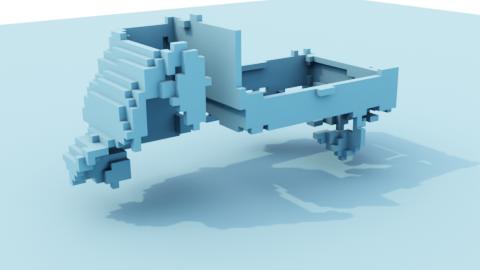} & \includegraphics[width=\qualwidth\linewidth]{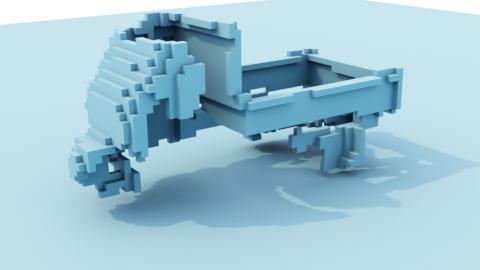} & \includegraphics[width=\qualwidth\linewidth]{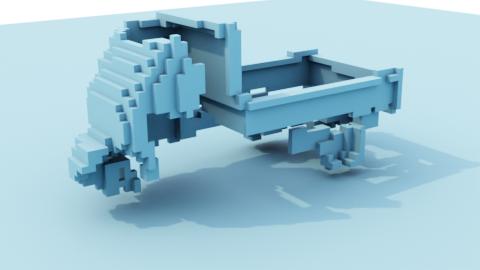} \\
       \rotatebox[origin=lB]{90}{GCA \scriptsize{($\text{20cm}$)}} & \includegraphics[width=\qualwidth\linewidth]{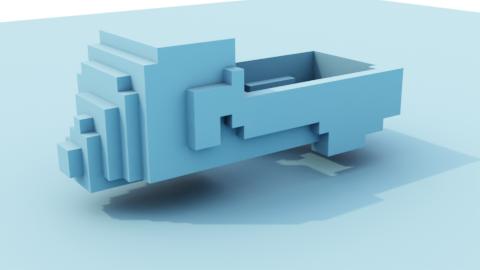} & \includegraphics[width=\qualwidth\linewidth]{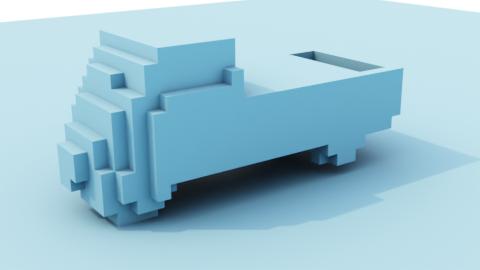} & \includegraphics[width=\qualwidth\linewidth]{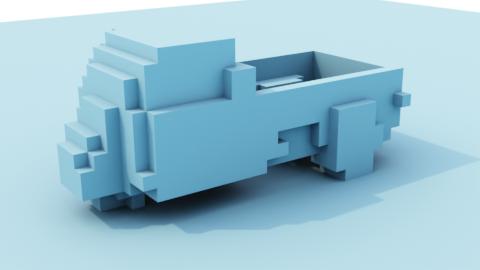} & \includegraphics[width=\qualwidth\linewidth]{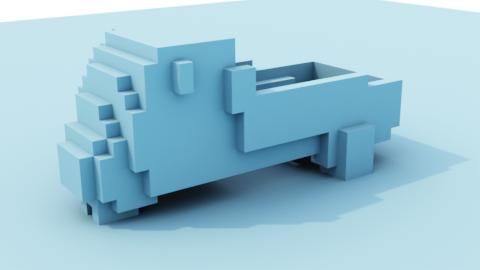} \\
       \rotatebox[origin=lB]{90}{\;\;\;hGCA}& \includegraphics[width=\qualwidth\linewidth]{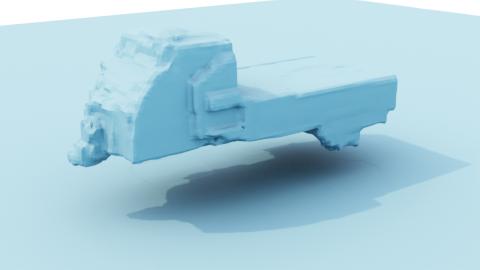} & \includegraphics[width=\qualwidth\linewidth]{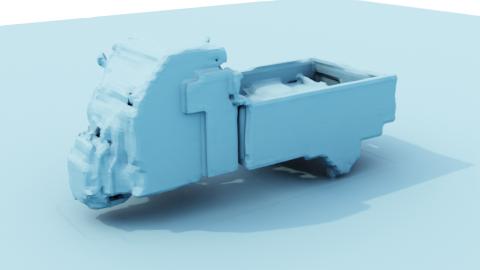} & \includegraphics[width=\qualwidth\linewidth]{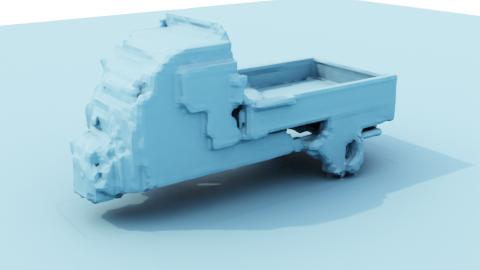} & \includegraphics[width=\qualwidth\linewidth]{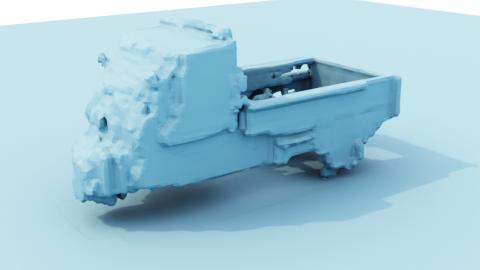} \\
       & 5\% Input & 10\% Input & 50\% Input & 100\% Input \\
    \end{tabular}
    \caption{
        Ablation study on a novel three-wheeler completion by varying density of 5 scans only for the car asset.
    }
    \label{fig:density_car}
\end{figure*}

\begin{table}[]
    \centering
    \resizebox{\linewidth}{!}{
        \begin{tabular}{l|cccc|cccc}
            \toprule
            & \multicolumn{4}{c|}{Sparse Scene} &  \multicolumn{4}{c}{Sparse Car}  \\
            \midrule
            Method & 5\% & 10\% & 50\% & 100\% & 5\% & 10\% & 50\% & 100\%  \\
            \midrule
            SCPNet ($\text{20cm}^3$)  & 10.44 & 2.33 & 2.28 & 2.25 & 2.38 & 2.30 & 2.37 & 2.25 \\ 
            JS3CNet ($\text{20cm}^3$) & 96.59 & 112.65 & 170.16 & 2.40 & 2.59 & 2.52 & 2.43 & 2.4 \\
            SG-NN ($\text{10cm}^3$) & 133.61 & 133.28 & 184.4 & 1.96 & 2.45 & 2.19 & 2.03 & 1.96 \\
            GCA ($\text{20cm}^3$) & 2.47 & 2.27 & 2.44 & 2.27 & 2.38 & 2.51 & 2.38 & 2.27 \\ 
            \midrule
            hGCA ($\text{10cm}^3$) & \textbf{2.41} & \textbf{2.11} & \textbf{1.77} & \textbf{1.71} & \textbf{2.30} & \textbf{2.08} & \textbf{1.82} & \textbf{1.69} \\
            \bottomrule
        \end{tabular}%
        }
    \caption{
        Chamfer distance between the ground truth geometry and completions by varying sparsity.
        Sparse scene and sparse car indicates a scenario where we sparsify the regions of entire scene (including ground) and only the car, respectively.
        Chamfer distance above ground are reported and we report average distance of $k=3$ generations for generative models (GCA, hGCA).
        hGCA generalizes well to sparse, novel data.
    }
    \label{table:asset_rep}
\end{table}

We first demonstrate the performance of hGCA given inputs of varying densities of the entire scene.
We place a three-wheeler\footnote{\label{sketchfab_footnote}\href{https://skfb.ly/osHxF}{Clickable link to asset on sketchfab.com}}, verified to be unseen from training data, in the center on a flat ground and simulate LiDAR scans captured from a simple trajectory.
Then, we vary the input density by randomly sampling 5\%, 10\%, and 50\% of the accumulated scans of two settings: the whole scene and only the car.

We report Chamfer distance between the completion and the ground truth in Table~\ref{table:asset_rep} and visualize the completions on both settings in Fig.~\ref{fig:density_scene} and~\ref{fig:density_car}.
hGCA generates the most accurate geometry compared to all baselines in every scenario according to Table~\ref{table:asset_rep}.
The completions of sparse scans for both the scene and the car show that vanilla GCA and hGCA still generate reasonable completions for a novel object.
Because other baselines (SCPNet~\cite{scpnet}, JS3CNet~\cite{JS3CNet}, SG-NN~\cite{dai2020sgnn}) utilize global features, they create random artifacts when the input is a severely sparse scene (5\% and 10\% in Fig.~\ref{fig:density_scene}).
Such challenging scenarios are effectively handled with local generations of GCAs, demonstrating superior performance on generalization.
As it is impossible to control the input quality or provide accurate object-wise segmentation in actual scans, the robustness of GCA may pave the way toward practical large-scale scene generation.

\paragraph{Varying Number of Input Scans}\label{app:scan_1_all}
In addition to random samples, we test a more realistic variation of densities, collecting simulated scans exhibiting occlusions.
We train models with five and ten scans on synthetic scenes (Karton City and CARLA~\cite{carla}) and evaluate results on inputs with different numbers of scans.
hGCA can also stably create scenes given various numbers of input scans.


\begin{table*}[]
    \centering
    \resizebox{0.8\linewidth}{!}{
        \begin{tabular}{l|c|cc|c|c|cc|c|c|cc|c}
            \toprule
            & & \multicolumn{4}{c|}{CARLA} &  \multicolumn{7}{c}{Karton City}  \\
            \midrule
            \multirow{2}{*}{Method} & \multirow{2}{*}{Representation} & \multicolumn{3}{c|}{High LiDAR ReSim} & \multirow{2}{*}{IoU} &  \multicolumn{3}{c|}{High LiDAR ReSim} & \multirow{2}{*}{IoU} & \multicolumn{3}{c}{Street CD}   \\
            & & min. $\downarrow$ & avg. $\downarrow$ & TMD $\uparrow$ & & min. $\downarrow$ & avg. $\downarrow$ & TMD $\uparrow$ & & min. $\downarrow$ & avg. $\downarrow$ & TMD $\uparrow$ \\
            \midrule
            ConvOcc & implicit & \dc{21.73} & - & 12.05 & \dc{16.6} & - & 22.05 & \dc{25.20} & - \\
            SCPNet & 20cm & \dc{9.46} & - & 38.94 & \dc{8.23} & - & 55.01 & \dc{7.37} & - \\
            \multirow{2}{*}{JS3CNet} & 20cm & \dc{8.70} & - & \textbf{43.04} & \dc{7.20} & - & 57.05 & \dc{7.65} & - \\
            & 10cm &  \dc{8.49} & - & 32.78 & \dc{6.33} & - & 62.21 & \dc{\textbf{5.48}} & - \\ 
            SG-NN & 10cm & \dc{9.53} & - & 37.55 & \dc{7.11} & - & 60.62 & \dc{6.44} & - \\
            \multirow{2}{*}{GCA} & 20cm & 8.37 & 8.82 & 1.89 & 39.07 & 5.60 & 5.86 & 1.40 & \textbf{63.69} & 6.17 & 7.47 & 3.21\\
            & 10cm & 9.16 & 10.23 & 4.14 & 34.73 & 7.37 & 7.77 & 2.14 & 57.33 & 8.00 & 9.18 & 3.89\\
            cGCA & implicit & 9.27 & 10.10 & \textbf{4.45} & 31.56 & 6.63 & 7.06 & \textbf{3.01} & 47.33 & 10.09 & 11.83 & \textbf{7.99} \\

            \midrule
            \multirow{2}{*}{hGCA} & 10cm & 7.97 & 8.25 & 1.19 & 40.05 & \textbf{5.16} & 5.29 & 0.91 & 63.29 & 6.27 & 6.93 & 1.43\\
            & implicit & \textbf{7.94} & 8.22 & 1.39 & 40.47 & 5.18 & 5.30 & 0.96 & 59.97 & 5.99 & 6.67 & 1.27 \\

            \midrule
            \dc{input} & \dc{12.72} & - & 19.26 & \dc{12.87} & - & 21.77 & \dc{10.14} & - \\ 
            \bottomrule
        \end{tabular}%
        }
        \caption{
            Quantitative results on CARLA and Karton City with a single scan given as input.
            All results except IoU are multiplied by 10 in meter scale.
            LiDAR Resim and Street CD evaluates the fidelity of completion and TMD measures the diversity of generation.
            High LiDAR Resim uses high elevation LiDAR to evaluate the extrapolation.
            IoU is computed with ground truth geometry.
    }
    \label{table:scan_1}
   
\end{table*}

We report quantitative results in Table~\ref{table:scan_1} and visualize random samples in Fig.~\ref{fig:synthetic_1} for the extreme case, where only a single scan is provided.
Quantitatively, hGCA outperforms other baselines by a large margin in LiDAR Resim and shows competitive (second best) performance on IoU and Street CD, where the best method differs depending on the dataset.
While hGCA demonstrates superior generalization performance to a single scan compared to other baselines, it suffers from degradation with the lack of evidence in the input.
For example, in the second column of Fig.~\ref{fig:synthetic_1}, the input scan provided in a limited height range results in the ambiguity between the facade of the building and the fence.


\begin{table*}[]
    \centering
    \resizebox{0.8\linewidth}{!}{
        \begin{tabular}{l|c|cc|c|c|cc|c|c|cc|c}
            \toprule
            & & \multicolumn{4}{c|}{CARLA} &  \multicolumn{7}{c}{Karton City}  \\
            \midrule
            \multirow{2}{*}{Method} & \multirow{2}{*}{Representation} & \multicolumn{3}{c|}{High LiDAR ReSim} & \multirow{2}{*}{IoU} &  \multicolumn{3}{c|}{High LiDAR ReSim} & \multirow{2}{*}{IoU} & \multicolumn{3}{c}{Street CD}   \\
            & & min. $\downarrow$ & avg. $\downarrow$ & TMD $\uparrow$ & & min. $\downarrow$ & avg. $\downarrow$ & TMD $\uparrow$ & & min. $\downarrow$ & avg. $\downarrow$ & TMD $\uparrow$ \\
            \midrule
            ConvOcc & implicit & \dc{13.4} & - & 17.81 & \dc{8.35} & - & 27.36 & \dc{13.4} & - \\
            SCPNet & 20cm & \dc{6.03} & - & 52.10 & \dc{4.18} & - & 75.61 & \dc{3.09} & - \\
            \multirow{2}{*}{JS3CNet} & 20cm & \dc{6.41} & - & 54.32 & \dc{5.37} & - & 65.32 & \dc{3.59} & - \\
            & 10cm & \dc{5.49} & - & 48.15 & \dc{3.45} & - & 75.44 & \dc{1.93} & - \\ 
            SG-NN & 10cm & \dc{\textbf{4.20}} & - & \textbf{59.17} & \dc{3.18} & - & 76.76 & \dc{1.84} & - \\
            \multirow{2}{*}{GCA} & 20cm & 5.31 & 5.54 & 1.45 & 56.18 & 3.79 & 3.83 & 0.40 & 79.76 & 2.64 & 2.91 & 0.76\\
            & 10cm & 5.35 & 5.85 & 1.94 & 50.97 & 3.17 & 3.24 & 0.64 & 77.47 & 2.02 & 2.28 & 0.84\\
            cGCA & implicit & 6.43 & 6.82 & \textbf{2.34} & 40.72 & 3.92 & 3.97 & \textbf{0.67} & 66.86 & 3.00 & 3.28 & \textbf{1.36} \\

            \midrule
            \multirow{2}{*}{hGCA} & 10cm & 4.54 & 4.64 & 0.84 & 58.78 & 2.95 & 2.98 & 0.35 & \textbf{81.87} & 1.73 & 1.84 & 0.47\\
            & implicit & 4.36 & 4.46 & 0.89 & 56.85 & \textbf{2.89} & 2.92 & 0.39 & 75.48 & \textbf{1.54} & 1.66 & 0.38 \\

            \midrule
            \dc{input} & \dc{5.68} & - & 45.77 & \dc{5.09} & - & 62.36 & \dc{5.58} & - \\ 
            \bottomrule
        \end{tabular}%
        }
        \caption{
            Quantitative results on CARLA and Karton City with a many scans (CARLA: 80 scans, Karton City: average of 132 scans) given as input.
            All results except IoU are multiplied by 10 in meter scale.
            LiDAR Resim and Street CD evaluates the fidelity of completion and TMD measures the diversity of generation.
            High LiDAR Resim uses high elevation LiDAR to evaluate the extrapolation.
            IoU is computed with ground truth geometry.
    }
    \label{table:scan_all}
   
\end{table*}

We also test our methods by completing dense accumulated scans.
For CARLA, we accumulated 80 nearby scans from a random pose; for Karton City, we gathered all the scans, which have an average of 132 scans, as input.
We report quantitative results in Table~\ref{table:scan_all} and visualize completions randomly in Fig.~\ref{fig:synthetic_all}.
hGCA outperforms previous methods on all reconstruction methods in Karton City and performs competitively (second best) in CARLA.
While SG-NN reports best results on CARLA, we observe that SG-NN struggles to create scenes beyond the sensor range, such as faces of buildings (first column) or trees or cars (second column) in Fig.~\ref{fig:synthetic_all}.
In contrast, hGCA successfully generates reasonable geometry in our qualitative examples.


\subsection{Planner Feature Visualization}

\begin{figure*}
    \centering
    \begin{tabular}{cccc}
        \includegraphics[width=0.22\textwidth]{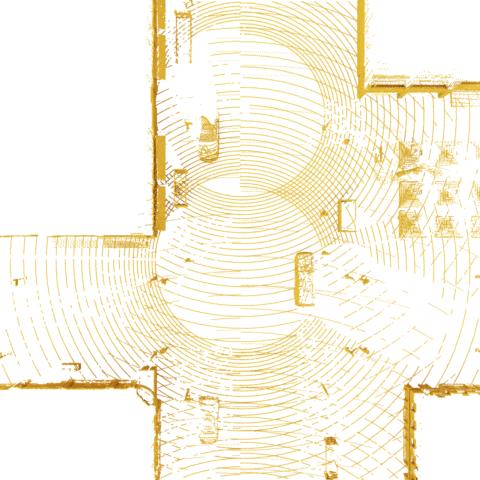} & 
         \includegraphics[width=0.22\textwidth]{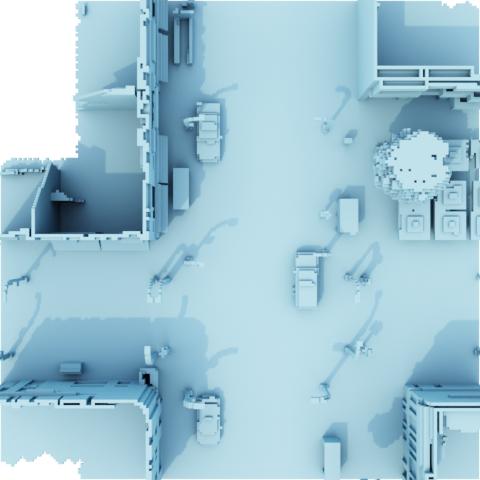} & \includegraphics[width=0.22\textwidth]{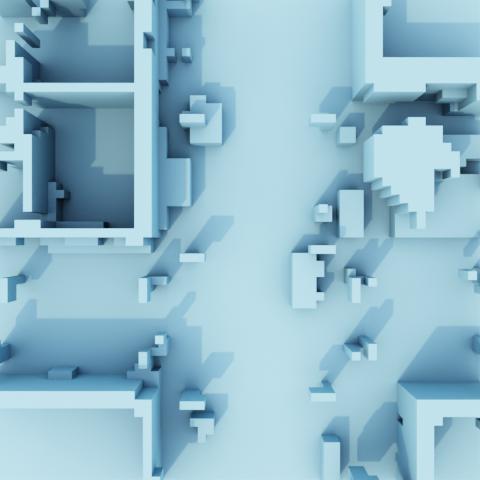} & \includegraphics[width=0.22\textwidth]{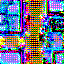}  \\
        \includegraphics[width=0.22\textwidth]{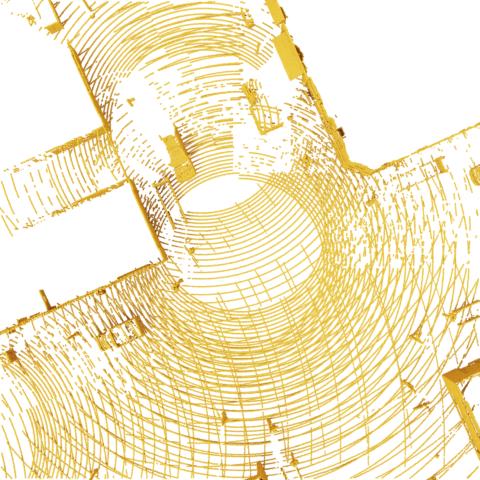} & \includegraphics[width=0.22\textwidth]{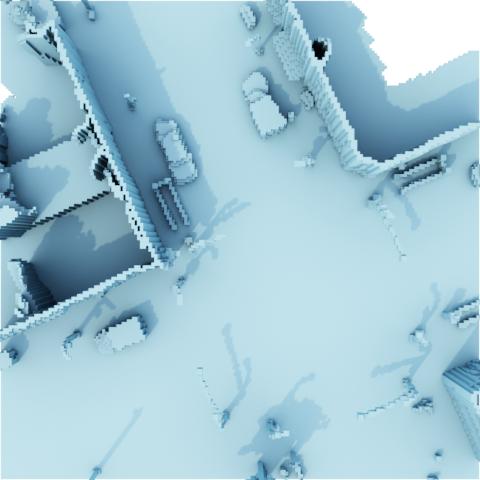} & \includegraphics[width=0.22\textwidth]{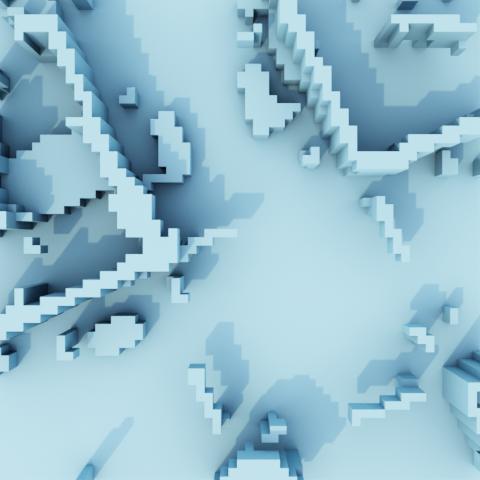} & \includegraphics[width=0.22\textwidth]{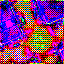}  \\
        \footnotesize{(a) Input} & \footnotesize{(b) Completion} & \footnotesize{(c) Rough dense occupancy} & \footnotesize{(d) PCA visualizations of BEV feature} \\
    \end{tabular}
    \caption{ 
        Planner with $z_r=4$ visualization. From left to right: 5 scan input from Karton City, completion (20$\text{cm}^3$ resolution), rough dense occupancy $O_r$ from planner, BEV feature $f_{BEV}$ visualization using PCA.
    }
    \label{fig:bev_feat}
\end{figure*}

For further understanding of the planner, we provide visualizations of outputs and features of planner.
Fig.~\ref{fig:bev_feat} shows the input, completion of GCA equipped with planner, rough dense occupancy $O_r$ of planner, and PCA visualizations of BEV feature.
For the PCA visualization, we project the 2D BEV features of the planner to RGB using the first 3 principal axes of PCA. 
We observe that the final completion follows the rough dense occupancy prediction of the planner.
This demonstrates that the planner acts as a memory that persists through the Markov process of GCA and \textit{plans ahead} persisting BEV feature solely with initial state $s^0$.
Also, we observe that BEV feature learns some global context that distinguishes some semantic classes trained without any semantic supervision.
We presume that checkerboard artifacts arise due to deconvolution layers of unet~\cite{odena2016deconvolution}.

\subsection{Effects of LiDAR noise}\label{app:clean_scan}
\begin{figure*}[]
    \centering
    \begin{subfigure}[b]{0.3\textwidth}
         \centering
         \includegraphics[width=\textwidth]{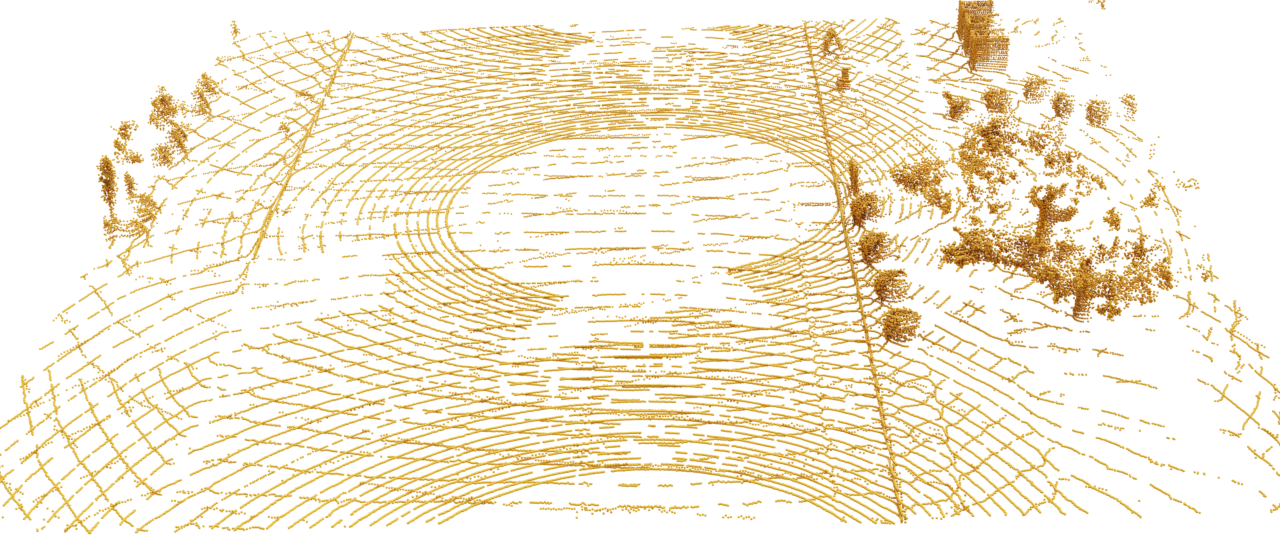}
         \caption{Input}
     \end{subfigure}
     \hfill
    \begin{subfigure}[b]{0.3\textwidth}
         \centering
         \includegraphics[width=\textwidth]{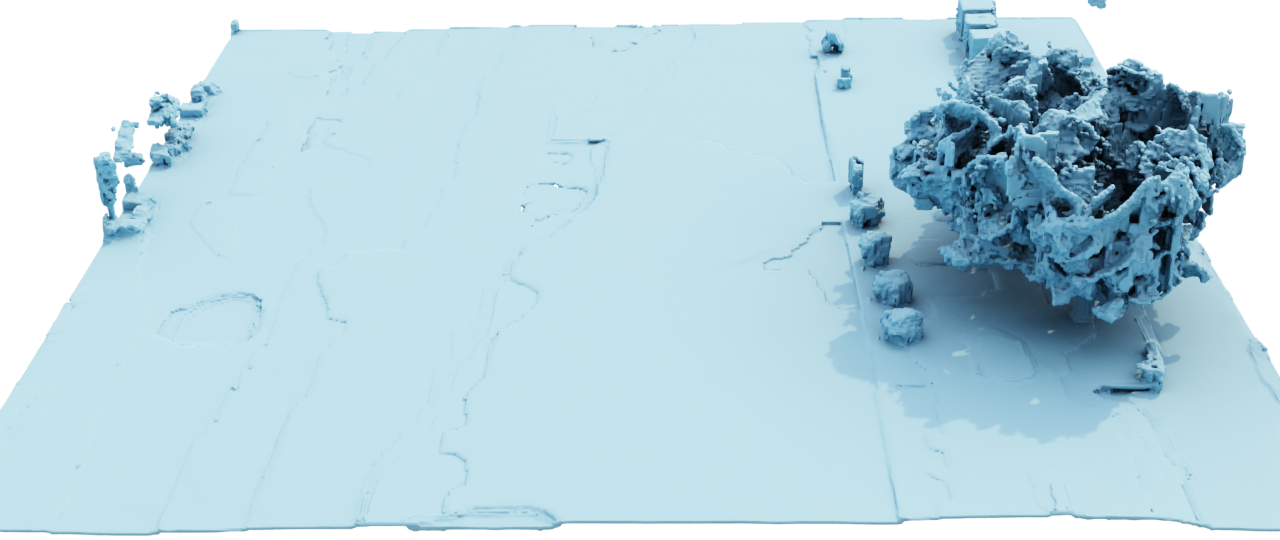}
         \caption{Trained with noise}
     \end{subfigure}
     \hfill
     \begin{subfigure}[b]{0.3\textwidth}
         \centering
         \includegraphics[width=\textwidth]{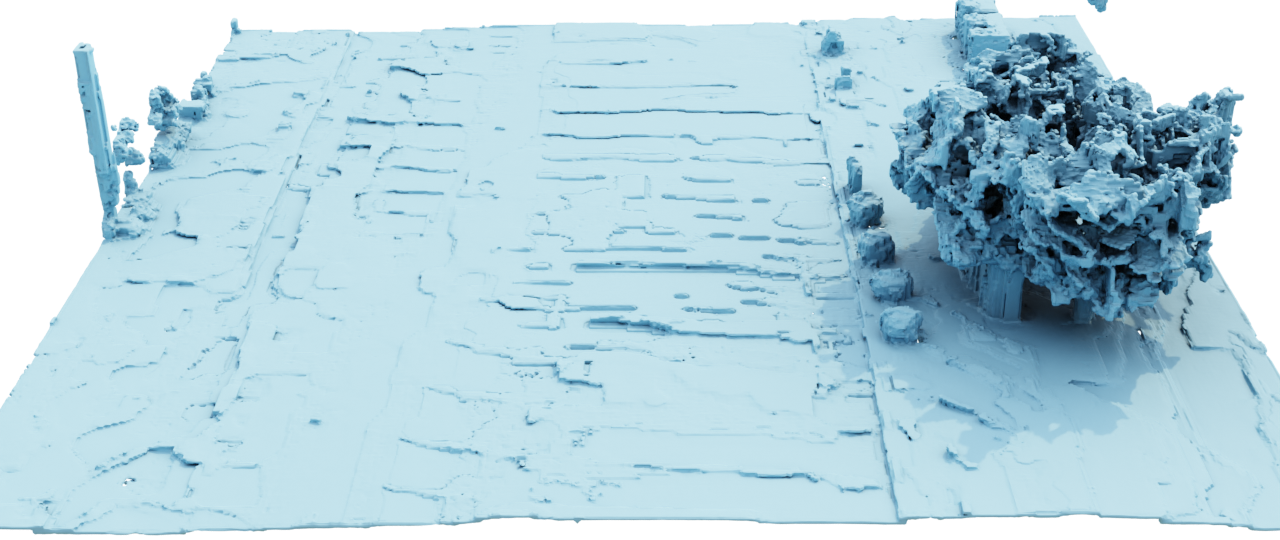}
         \caption{Trained without noise}
     \end{subfigure}
     \hfill
    \caption{
        Completion visualization on Waymo-Open dataset with and without noise.
        From left to right: input, completion trained with noise, completion trained without noise.
        Adding noise during training with synthetic data produces cleaner completions on real-world data.
    }
    \label{fig:waymo_noise}
\end{figure*}

\begin{table*}[t]
    \centering
    \resizebox{0.8\linewidth}{!}{
        \begin{tabular}{l|c|cc|c|c|cc|c|c|cc|c}
            \toprule
            & & \multicolumn{4}{c|}{CARLA} &  \multicolumn{7}{c|}{Karton City}  \\
            \midrule
            \multirow{2}{*}{Method} & \multirow{2}{*}{Representation} & \multicolumn{3}{c|}{High LiDAR ReSim} & \multirow{2}{*}{IoU} &  \multicolumn{3}{c|}{High LiDAR ReSim} & \multirow{2}{*}{IoU} & \multicolumn{3}{c}{Street CD}   \\
            & & min. $\downarrow$ & avg. $\downarrow$ & TMD $\uparrow$ & & min. $\downarrow$ & avg. $\downarrow$ & TMD $\uparrow$ & & min. $\downarrow$ & avg. $\downarrow$ & TMD $\uparrow$ \\
            \midrule
            ConvOcc & implicit & \dc{14.45} & - & 14.66 & \dc{9.97} & - & 25.32 & \dc{16.91} & - \\
            SCPNet & 20cm & \dc{6.28} & - & 54.20 & \dc{4.75} & - & 70.83 & \dc{3.29} & - \\
            \multirow{2}{*}{JS3CNet} & 20cm & \dc{6.30} & - & 52.76 & \dc{5.16} & - & 64.93 & \dc{3.30} & - \\
            & 10cm & \dc{4.87} & - & 54.80 & \dc{3.88} & - & 71.16 & \dc{2.48} & - \\ 
            SG-NN & 10cm & \dc{4.76} & - & 54.99 & \dc{3.84} & - & 72.82 & \dc{2.38} & - \\
            \multirow{2}{*}{GCA} & 20cm & 5.63 & 5.87 & 1.24 & 55.63 & 3.95 & 4.02 & 0.44 & 77.44 & 2.79 & 3.09 & 0.74\\
            & 10cm & 6.20 & 6.71 & \textbf{2.77} & 45.00 & 3.53 & 3.69 & 1.03 & 71.61 & 2.41 & 3.27 & 2.02\\
            cGCA & implicit & 6.50 & 6.96 & 2.62 & 33.21 & 4.28 & 4.46 & \textbf{1.33} & 60.42 & 2.7 & 3.86 & \textbf{2.35} \\

            \midrule
            \multirow{2}{*}{hGCA} & 10cm & \textbf{4.62} & 4.75 & 0.70 & \textbf{56.41} & \textbf{3.17} & 3.21 & 0.45 & \textbf{77.85} & 1.90 & 2.02 & 0.53\\
            & implicit & 4.71 & 4.83 & 0.79 & 55.20 & 3.18 & 3.22 & 0.51 & 72.70 & \textbf{1.65} & 1.77 & 0.41 \\

            \midrule
            \dc{input} & \dc{6.14} & - & 36.63 & \dc{6.76} & - & 39.51 & \dc{5.41} & - \\ 
            \bottomrule
        \end{tabular}%
        }
        \caption{
            Quantitative results on CARLA and Karton City with 5 scans given as input, both trained and evaluated without noise.
            All results except IoU are multiplied by 10 in meter scale.
            LiDAR Resim and Street CD evaluates the fidelity of completion and TMD measures the diversity of generation.
            High LiDAR Resim uses high elevation LiDAR to evaluate the extrapolation.
            IoU is computed with ground truth geometry.
    }
    \label{table:scan_5_clean}
   
\end{table*}


\begin{table*}[t]
    \centering
    \resizebox{0.8\linewidth}{!}{
        \begin{tabular}{l|c|cc|c|c|cc|c|c|cc|c}
            \toprule
            & & \multicolumn{4}{c|}{CARLA} &  \multicolumn{7}{c|}{Karton City}  \\
            \midrule
            \multirow{2}{*}{Method} & \multirow{2}{*}{Representation} & \multicolumn{3}{c|}{High LiDAR ReSim} & \multirow{2}{*}{IoU} &  \multicolumn{3}{c|}{High LiDAR ReSim} & \multirow{2}{*}{IoU} & \multicolumn{3}{c}{Street CD}   \\
            & & min. $\downarrow$ & avg. $\downarrow$ & TMD $\uparrow$ & & min. $\downarrow$ & avg. $\downarrow$ & TMD $\uparrow$ & & min. $\downarrow$ & avg. $\downarrow$ & TMD $\uparrow$ \\
            \midrule
            ConvOcc & implicit & \dc{13.67} & - & 15.11 & \dc{9.00} & - & 26.12 & \dc{15.24} & - \\
            SCPNet & 20cm & \dc{5.90} & - & 56.88 & \dc{4.29} & - & 74.84 & \dc{2.83} & - \\
            \multirow{2}{*}{JS3CNet} & 20cm & \dc{6.30} & - & 52.76 & \dc{4.99} & - & 66.63 & \dc{3.02} & - \\
            & 10cm & \dc{4.41} & - & 58.38 & \dc{3.40} & - & 74.32 & \dc{1.99} & - \\
            SG-NN & 10cm & \dc{4.31} & - & 58.05 & \dc{3.20} & - & 76.52 & \dc{1.84} & - \\
            \multirow{2}{*}{GCA} & 20cm & 5.46 & 5.65 & 1.09 & 58.12 & 3.81 & 3.86 & 0.31 & 81.02 & 2.62 & 2.88 & 0.62\\
            & 10cm & 5.68 & 6.21 & 2.45 & 48.27 & 3.08 & 3.19 & 0.69 & 76.54 & 1.94 & 2.49 & 1.38\\
            cGCA & implicit & 6.26 & 6.75 & 2.45 & 35.16 & 4.00 & 4.13 & \textbf{1.00} & 63.79 & 1.87 & 2.62 & \textbf{1.46} \\

            \midrule
            \multirow{2}{*}{hGCA} & 10cm & \textbf{4.28} & 4.39 & 0.67 & \textbf{59.00} & \textbf{2.96} & 2.99 & 0.34 & \textbf{81.34} & 1.62 & 1.69 & 0.44\\
            & implicit & 4.36 & 4.47 & 0.75 & 57.34 & \textbf{2.96} & 2.99 & 0.40 & 75.34 & \textbf{1.38} & 1.46 & 0.33 \\

            \midrule
            \dc{input} & \dc{5.19} & - & 44.02 & \dc{5.33} & - & 49.89 & \dc{4.56} & - \\ 
            \bottomrule
        \end{tabular}%
        }
        \caption{
            Quantitative results on CARLA and Karton City with 10 scans given as input, both trained and evaluated without noise.
            All results except IoU are multiplied by 10 in meter scale.
            LiDAR Resim and Street CD evaluates the fidelity of completion and TMD measures the diversity of generation.
            High LiDAR Resim uses high elevation LiDAR to evaluate the extrapolation.
            IoU is computed with ground truth geometry.
    }
    \label{table:scan_10_clean}
   
\end{table*}

We investigate the effects of input LiDAR noise on training data.
During training on synthetic data, we add Gaussian noise of standard deviation 0.01 in meter scale to the coordinates of each points and add noise to the pitch angle of the pose with standard deviation 0.02 in degree scale to simulate the LiDAR noise produced in data acquisition in real-world.
Fig.~\ref{fig:waymo_noise} visualizes the completion of hGCA with and without adding noise during training.
We observe that adding noise is crucial in terms of fine sim-to-real generalization, especially on the ground.
While we simulated the LiDAR noise with simple ray-casting and Gaussian noises, one could further reduce sim-to-real generalization by employing more sophisticated noise, such as~\cite{manivasagam2020lidarsim}.

For completeness, we report quantitative results compared to existing methods without adding noise during both training and validation in Table~\ref{table:scan_5_clean} and~\ref{table:scan_10_clean}.
Similar to results with noise, hGCA outperforms baselines on reconstruction metrics, demonstrating the superior extrapolation performance of hGCA.

\subsection{Space and Time Complexity}

\begin{table}[]
    \centering
    \resizebox{\linewidth}{!}{
        \begin{tabular}{l|ccccc}
            \toprule
            method & 40m & 60m & 80m & 100m & 120m\\
            \midrule
            Coarse completion & 2.8 & 3.1 & 3.6 & 4.4 & 4.8\\
            Upsampling &  5.8 & 7.0 & 10.2 & 12.8 & 15.7\\
            \bottomrule
        \end{tabular}%
        }
    \caption{
        Maximum GPU usage for 40 $\times$ $w$ meter completion on one Waymo scene (top visualization in Fig~\ref{fig:waymo_big}).
        $w$ denotes the width of the completion in the first row of the table and the unit of GPU memory is GB.
    }
    \label{table:memory}
\end{table}

In this section, we further analyze space and time complexity of our model.
We first investigate the GPU memory usage for hGCA.
In Table.~\ref{table:memory}, we report the GPU memory requirements by varying the completion size in one Waymo scene, visualized in top of Fig.~\ref{fig:waymo_big}.
Given an input of size 40 $\times w$ meters, we perform 3 completions and report the maximum GPU memory usage for the coarse completion and the upsampling module.
We find that hGCA can scalably generate fine geometry up to 40 $\times$ 120 meters without any tricks on a single 24GB GPU, demonstrating the superior scalability of hGCA by only employing efficient sparse convolutions and planner.
We also observe that only 4.8GB is required for the coarse completion on 120 meter scene, demonstrating the efficacy of the planner module.
While our ablation study was conducted on only a single scene, we find that GPU memory usage can vary  heavily depending on a scene. 

For time complexity, we find that coarse completion of our method takes 3 seconds and upsampling takes about 10 seconds (including IO) to create the final mesh in CARLA with 3090 GPU.
Indeed our method is slow, whereas other single inference methods (SCPNet~\cite{scpnet}) typically take around 0.1 seconds on A100 GPU to create the completion in 20cm voxel resolution. 
We expect faster inference using half-precision or more recently developed sparse convolution libraries, such as torchsparse~\cite{tang2022torchsparse}, but we leave it to future work.

\section{Dataset}\label{app:dataset}

\subsection{Karton City}
\begin{figure}
    \centering
    \includegraphics[width=0.5\textwidth]{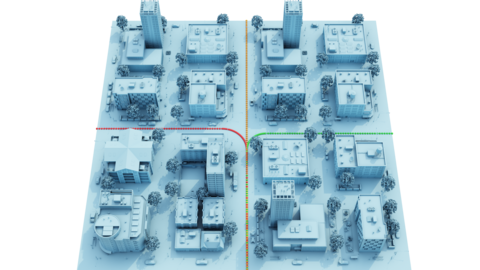}
    \includegraphics[width=0.5\textwidth]{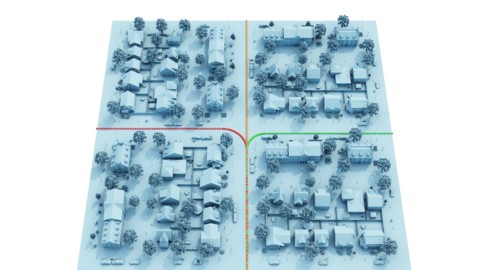}
    \caption{
        Karton City visualization. City (top) and suburb (bottom) scenes are visualized with three simple trajectories (red, yellow, green) for simulating a drive.
    }
    \label{fig:karton_city_vis}
\end{figure}
Karton City is a synthetic city comprised of 20 blocks, obtained from the Turbosquid marketplace\footnote{\label{turbosquidfootnote}\href{https://www.turbosquid.com/3d-models/3d-karton-city-2-model-1196110}{Clickable link to asset on turbosquid.com}} for 3D asset. 
We split 20 blocks into 12/3/5 train/val/test splits and re-combine 4 blocks in each split randomly to generate 300/30/60 unique train/val/test scenes of size 140 $\times$ 140 meters. 
Of the 12/3/5 split, 7/2/3 are city blocks and 5/1/2 are suburban blocks.
For realistic environment of scenes we re-combine city and suburban blocks separately, and generate 200/20/40, 100/10/20 scenes for train/val/test splits of city and suburban scenes, respectively, which we visualize in Fig.~\ref{fig:karton_city_vis}.
We place ShapeNet~\cite{shapenet2015} cars on each side of the main street to simulate parked cars.
The number of cars on each side of the street follow a Poisson distribution with lambda 2, with maximum 7 cars. 
We uniformly distribute the location of the cars and move them if collisions occur.  
The cars placed on the street follow the dataset split from ~\cite{yang2019pointflow} and remove meshes that contain unused vertices which makes it hard place the cars in the desired location, thus having total of  train/val/test split of 2383/339/670 cars.   
For each scene, we run three simple trajectories visualized in Fig.~\ref{fig:karton_city_vis} and randomly select 4615/30/180 center poses used for train/val/test split, respectively.
For each center pose, we create accumulated 5/10 scans for input by accumulating the scans from the center pose and 4/9 other poses, resepectively.
For obtaining ground truth implicit function, we sample 4,000,000 points from ground truth mesh with random noise variance 0.03 and 0.1 in meter scale, total of 8,000,000 point-distance pairs. 

\subsection{CARLA}\label{app:carla}
CARLA~\cite{carla} is an open source driving simulator with diverse environments. 
We use 5/1/1 towns as train/val/test split with randomly placed static vehicles and run 10 drives for each town to obtain the scans.
For each town, we run 10 drives and randomly select 3500/30/180 center poses used for train/val/test split, respectively.
For each center pose, we create accumulated 5/10 scans for input by accumulating the scans from the center pose and 4/9 other nearby poses, respectively.
In CARLA, obtaining ground truth mesh is non-trivial.
Therefore, we leverage extra LiDAR sensors other than the LiDAR sensor to collect input scans, to obtain the ground truth geometry.
We additionally place 5 LiDAR sensors with high elevation angle, having relative offsets of (0, 0, 0), (0, -9.6, 3), (0, 9.6, 3), (0, -19.2, 3), (0, 19.2, 3) meters from the position of LiDAR sensor placed on simulated ego-vehicle to collect input scans, where x-axis and z-axis refer to the front and up direction of the ego-vehicle.
We obtain ground truth surface points by accumulating points from scans acquired from additional LiDAR sensor, visualized in top of Fig.~\ref{fig:carla_vis}.
For obtaining implicit function, we sample 4,000,000 points from ground truth surface points with random noise variance 0.03 and 0.1 in meter scale, total of 8,000,000 point-distance pairs.
The distance for each points are computed with nearest neighbor against the ground truth surface, since ground truth mesh is not available.

\subsection{Waymo-Open}
Waymo-Open~\cite{Waymo-Open} is a real world dataset for autonomous driving containing sequence of LiDAR scans from a ego-vehicle drive. 
We randomly sampled 202 scenes and selected 3 center poses for each scene to evaluate the quantitative metrics. 
For each center pose, we create accumulated scans that serve as input by accumulating the scans from the center pose and 4 other poses within 50 meters.
We remove dynamic objects from the point clouds using annotated bounding box tracks for both input and accumulated point clouds.
Due to noisy dynamic label annotations, some sparse dynamic points lie after bounding box filtering.
Thus, we further perform erosion to the accumulated points in voxel resolution of $\text{10cm}^3$.

\subsection{Lidar Simulation} \label{app_sec:lidar_sim}
For all the experiments, we simulate synthetic LiDAR using the LiDAR beam angle and rotation-speed parameters from the Waymo-Open dataset~\cite{Waymo-Open}. 
The Waymo LiDAR captures full 360 degrees and results in range image dimension of 64 $\times$ 2650 pixels. 
For details on the sensor specification see~\cite{https://doi.org/10.48550/arxiv.1912.04838}. 
To simulate LiDAR on Karton City, we generate a curve by interpolating the ego-vehicle poses and simulate a rotating beam along this curve. 
This simulation, thus correctly captures rolling shutter effects. 
To obtain ground truth points in CARLA, we use 512 channels LiDAR with field of view (-30\textdegree, 30\textdegree) and range of 75 meters.

\section{Implementation Details}

\subsection{hierarchical Generative Cellular Automata}

\begin{figure}
    \centering
    \includegraphics[width=0.8\linewidth]{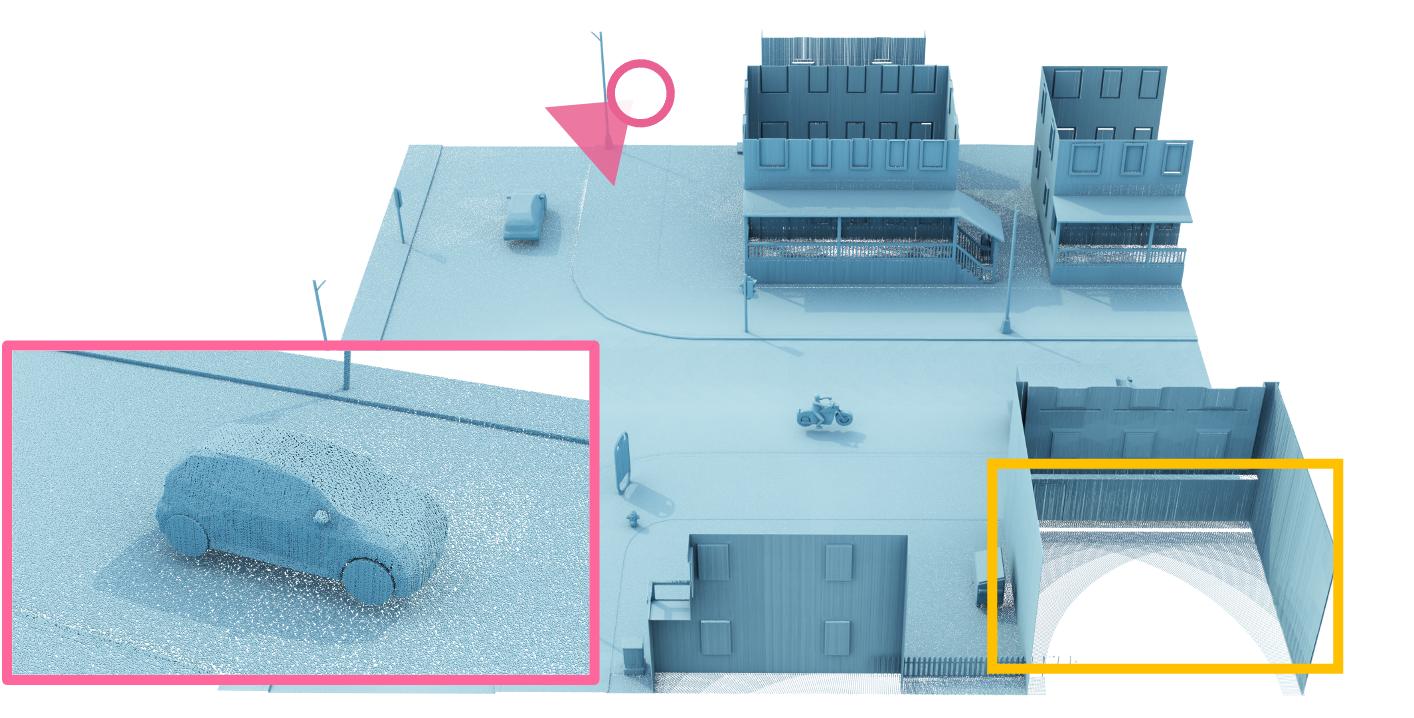}\\
    \includegraphics[width=0.8\linewidth]{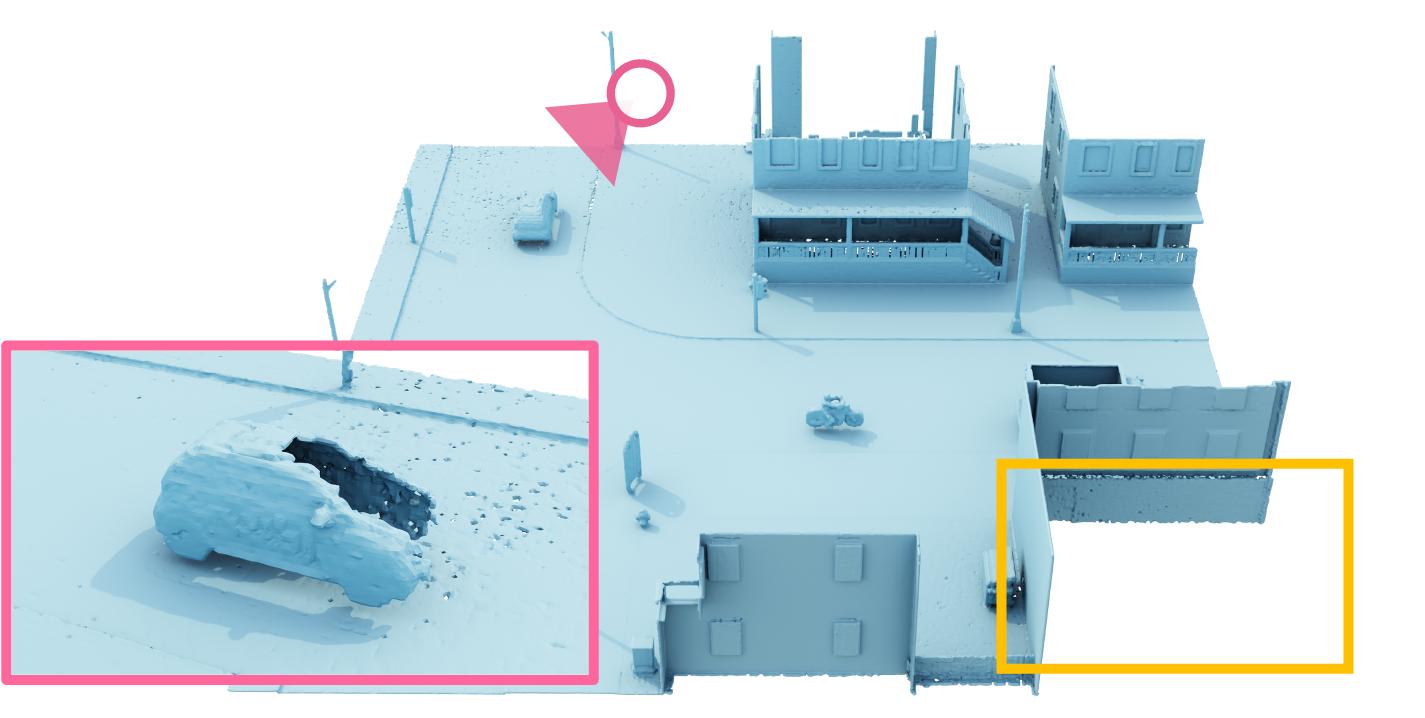}
    \caption{
        Visualization of CARLA dataset. 
        Top: Ground truth surface points obtained from additional sensors. 
        For visualization of density, we intentionally render with low density.
        Bottom: Decoded implicit function for supervising upsampling module.
        Inconsistent, sparse regions (yellow) lead upsampling module to unstable training. 
        Thus, to remove sparse regions for supervision, we train only on regions visible from the street, which may be incomplete (pink), but provide dense and accurate geometry for supervision.
        However, upsampling is a local operation and upsampling stage of hGCA can be trained with incomplete data.
    }
    \label{fig:carla_vis}
\end{figure}

\textbf{Training Upsampling Module.}
We train the upsampling module by minimizing the log-likelihood of the data distribution, which we defer to cGCA~\cite{zhang2022probabilistic} for further details.
We train on synthetic data by combining CARLA and Karton City like other methods.
However, as mentioned in Sec.~\ref{app:carla}, obtaining ground truth mesh is difficult for CARLA, and while the ground truth points obtained from additional sensors may be dense enough for $\text{10cm}^3$ voxel resolution, we observe sparse surface points in occluded regions from the street, such as interiors of the building~(Fig.~\ref{fig:carla_vis}).
Naive training of upsampling module with sparse surface points led to unstable training of local implicit latent feature, where upsampling results varied inconsistently depending on training step.
Therefore, for training the upsampling module on CARLA, we supervise with ground truth augmented state $x$ on regions visible from the road, as visualized in bottom of Fig.~\ref{fig:carla_vis}, which tend to be dense.
We observed that the upsampling module can generalize to complete scenes even with training on incomplete ground truth, since upsampling is a local operation.
During training, we train on combined CARLA and Karton City dataset with rate 15\% and 85\%, which led to stable training.

\textbf{Neural Network Architecture.}
For sparse convolution network of our coarse completion and upsampling module, we employ the same MinkowskiUNet~\cite{choy20194d, ronneberger2015unet} as in GCA~\cite{zhang2021learning} and cGCA~\cite{zhang2022probabilistic}, respectively.
We additionally append 3D positional encoding with 128 dimension to the features of the input sparse tensor.
For planner module, local point net consists of a fully connected layer that transforms the normalized coordinates to 32-dimension feature followed by 3 fully connected residual block and another fully connected layer with a 32-dimension feature output.
The residual block consists of 2 fully connected layers with 32 hidden dimensions.
After the local pointnet, we add 2D positional encoding to the features and pass it through a 2D UNet\footnote{\href{https://github.com/milesial/Pytorch-UNet}{https://github.com/milesial/Pytorch-UNet}}~\cite{ronneberger2015unet} to obtain a 2D feature of dimension 128.
Lastly, we employ 5 convolutional blocks, which consists of two convolutions of kernel size 3, for obtaining 4 SPADE~\cite{park2019SPADE} features that compute the mean and variance per pillar for denormalization and one rough occupancy prediction.

\textbf{Other Details.}
We use MinkowskiEngine~\cite{choy20194d} for sparse convolutions.
For all GCAs we use the infusion scheduler of $\alpha^t=0.15 + 0.005t$, and obtain the last state with additional maximum likelihood estimation instead of randomly sampling.
We use Adam~\cite{kingma2015adam} optimizer with constant learning rate 5e-4 and clip gradient with maximum norm of 0.5.
For our coarse completion model, we use batch size of 6 and for the upsampling module, we crop a scene into quarters and use batch size of 3.
We train the low-resolution GCA attached with planner for 400k steps and upsampling cGCA for 300k steps which takes roughly 5/4 days, respectively, with a single 3090 GPU.
Note that the two models can be trained independently.

\subsection{Baselines} \label{app:baselines}
\textbf{Generative Celluar Automata~\cite{zhang2021learning, zhang2022probabilistic}}.
We use the official implementation released from the authors\footnote{\href{https://github.com/96lives/gca}{https://github.com/96lives/gca}}.
For fair comparison, we use the same hyperparameters as our model.
We use cGCA of voxel size $\text{20cm}^3$.

\textbf{Convolutional Occupancy Networks~\cite{Peng2020ECCV}}
We use the 3D grid resolution of 64 version from the official implementation released from the authors\footnote{\href{https://github.com/autonomousvision/convolutional_occupancy_networks}{https://github.com/autonomousvision/convolutional\_occupancy\_networks}}.
For obtaining occupancy representation, since we cannot obtain watertight mesh for neither Karton City nor CARLA, we make occupancy for point in the sampled point-distance pairs that have distance to surface below 5cm.
We additionally sample 100,000 points in the block range uniformly to create unoccupied points.

\textbf{SG-NN~\cite{dai2020sgnn}}.
We compare with the state-of-the-art indoor scene completion network.
We use the official implementation released from the authors\footnote{\href{https://github.com/angeladai/sgnn}{https://github.com/angeladai/sgnn}}.
We use the SG-NN to predict the occupancy in $\text{10cm}^3$ voxel resolution.

\textbf{JS3CNet~\cite{JS3CNet} and SCPNet~\cite{scpnet}}
We compare with JS3CNet~\cite{JS3CNet} and SCPNet~\cite{scpnet}, state-of-the-art outdoor semantic scene completion methods. 
We use the official implementation released from the authors\footnote{\href{https://github.com/yanx27/JS3C-Net}{https://github.com/yanx27/JS3C-Net}}\footnote{\href{https://github.com/SCPNet/Codes-for-SCPNet}{https://github.com/SCPNet/Codes-for-SCPNet}}.
We adapt the method to our setting by changing the semantic class output to binary variable representing occupancy. 
We observe a class imbalance problem during training, where there are much more empty voxels than the occupied ones.
We find that weighing the loss 3:1 for occupied to empty cells performs best for both models.
While the original semantic scene completion works uses $20\text{cm}^3$ voxel resolution, we additionally train on $10\text{cm}^3$ voxel resolution for JS3CNet.
For $10\text{cm}^3$ model, we modify the output resolution to $10\text{cm}^3$ to match our voxel resolution by adding an extra upsampling layer. 
For SCPNet, we omit training on $10\text{cm}^3$ resolution since it did not fit in a single 24GB GPU.

\subsection{Other Implementation Details} \label{app:other_implementation_details}
All of our methods are implemented using PyTorch~\cite{paszke2019pytorch}.
For all point cloud to voxel conversion, we first round point cloud into $\text{10cm}^3$ voxels and use floor operation on the coordinates of voxels to create $\text{20cm}^3$ voxels.
For any models except the upsampling stage of hGCA, we train on a dataset combined with CARLA and Karton city having a rate of 50\% for each dataset.
For methods that utilize unsigned distance fields (cGCA, hGCA), we create mesh in $\text{5cm}^3$ voxel resolution with marching cubes~\cite{lorensen1987mcubes} using the unsigned distance values of the voxels. 
For IoU and street CD evaluation, we obtain points close to surface by sampling voxels in $\text{5cm}^3$ resolution that have implicit distance below 0.5.
For ConvOcc~\cite{Peng2020ECCV}, we evaluate IoU and street CD by sampling from the points created from the mesh, which represents the surface of the completed shape in occupancy representation.
We use blender~\cite{blender} for visualization.
For visualizations overlayed with input, such as Fig.~\ref{fig:waymo_big}, we render input points if a point is either in front of a mesh or is less than 0.5 meters back from the mesh in the rendering view.

\begin{figure*}[]
    \centering
    \begin{tabular}{cccc}
        \includegraphics[width=0.3\textwidth]{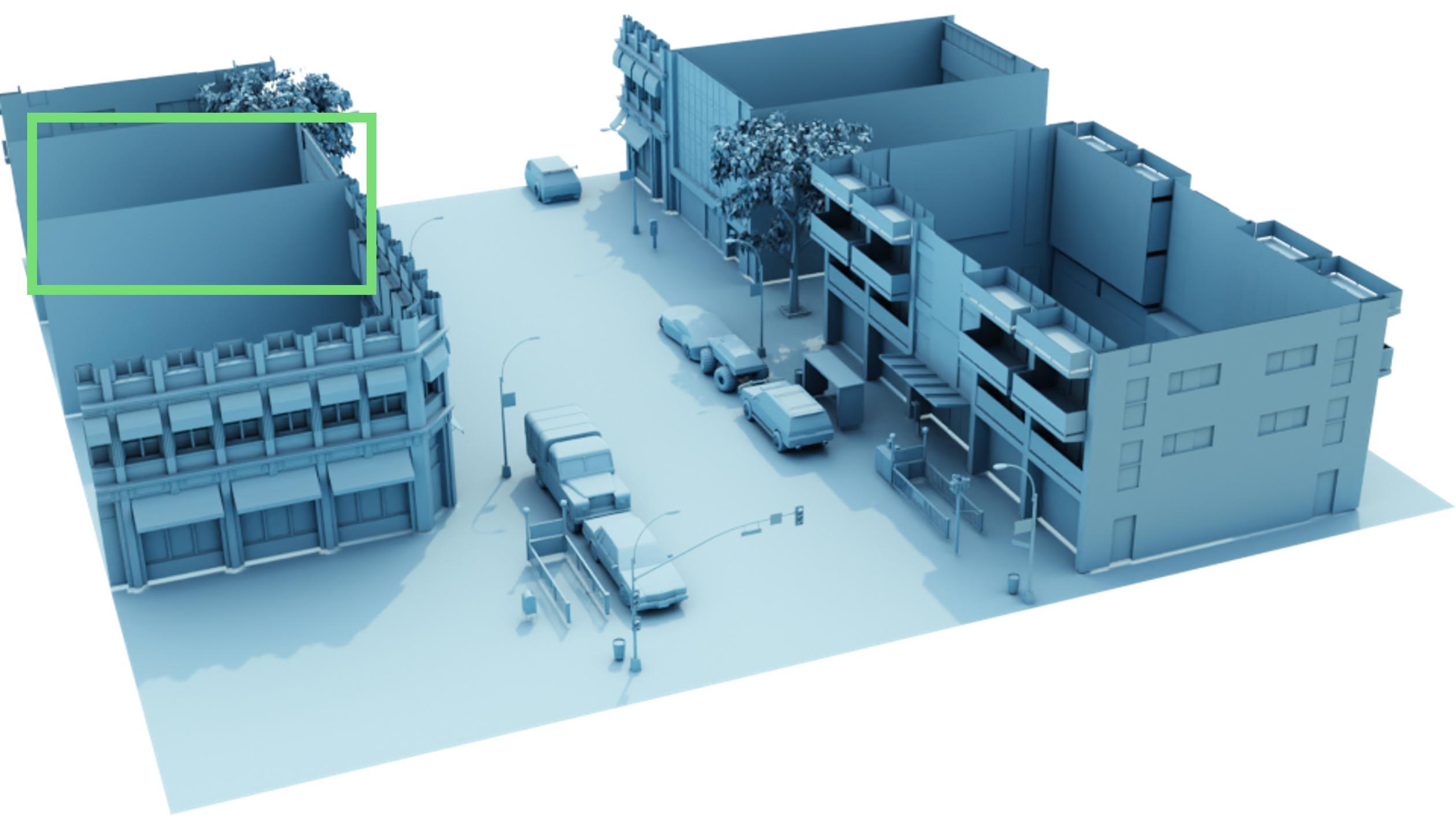} &
        \includegraphics[width=0.3\textwidth]{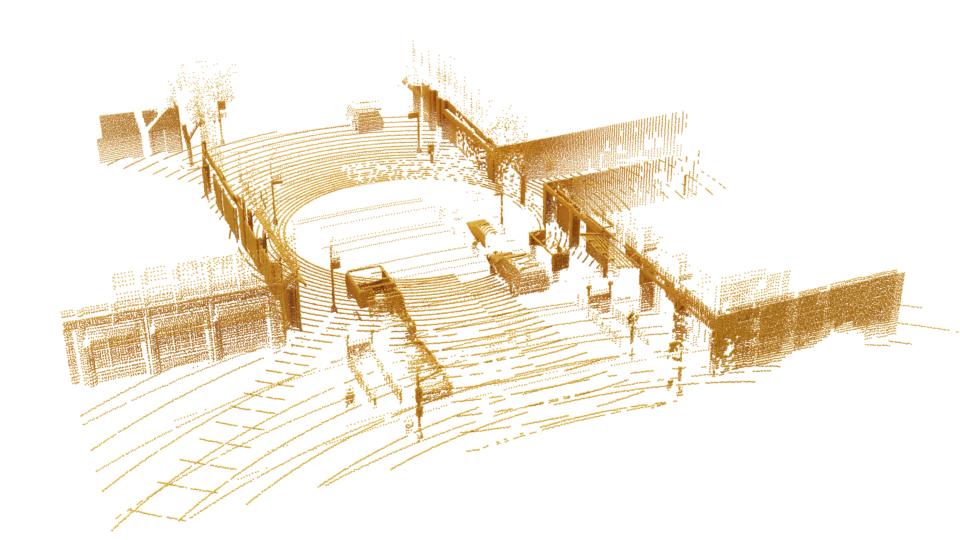} &
        \includegraphics[width=0.3\textwidth]{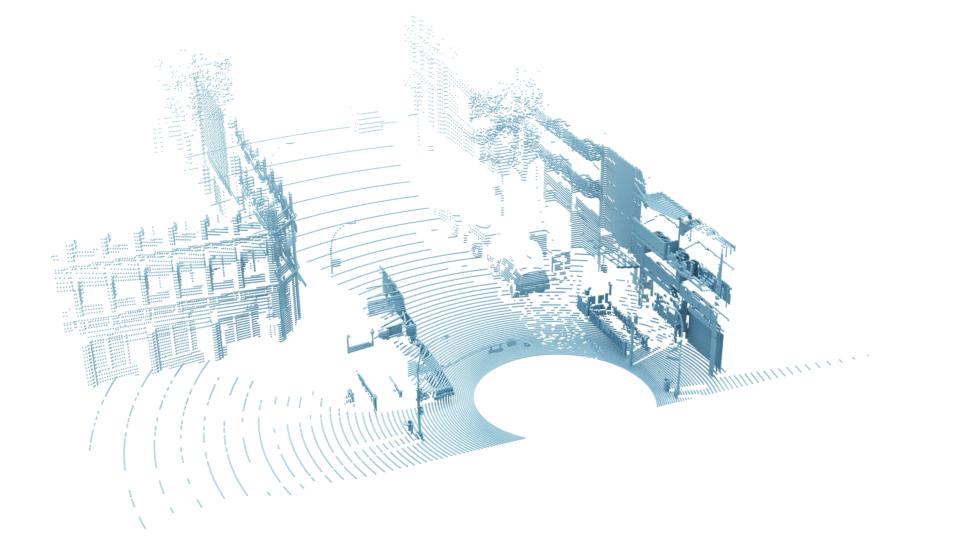} \\
        (a) GT  & (b) Input scans & (c) High LiDAR ReSim  \\[6pt]
        \end{tabular}
        \begin{tabular}{cccc}
        \includegraphics[width=0.3\textwidth]{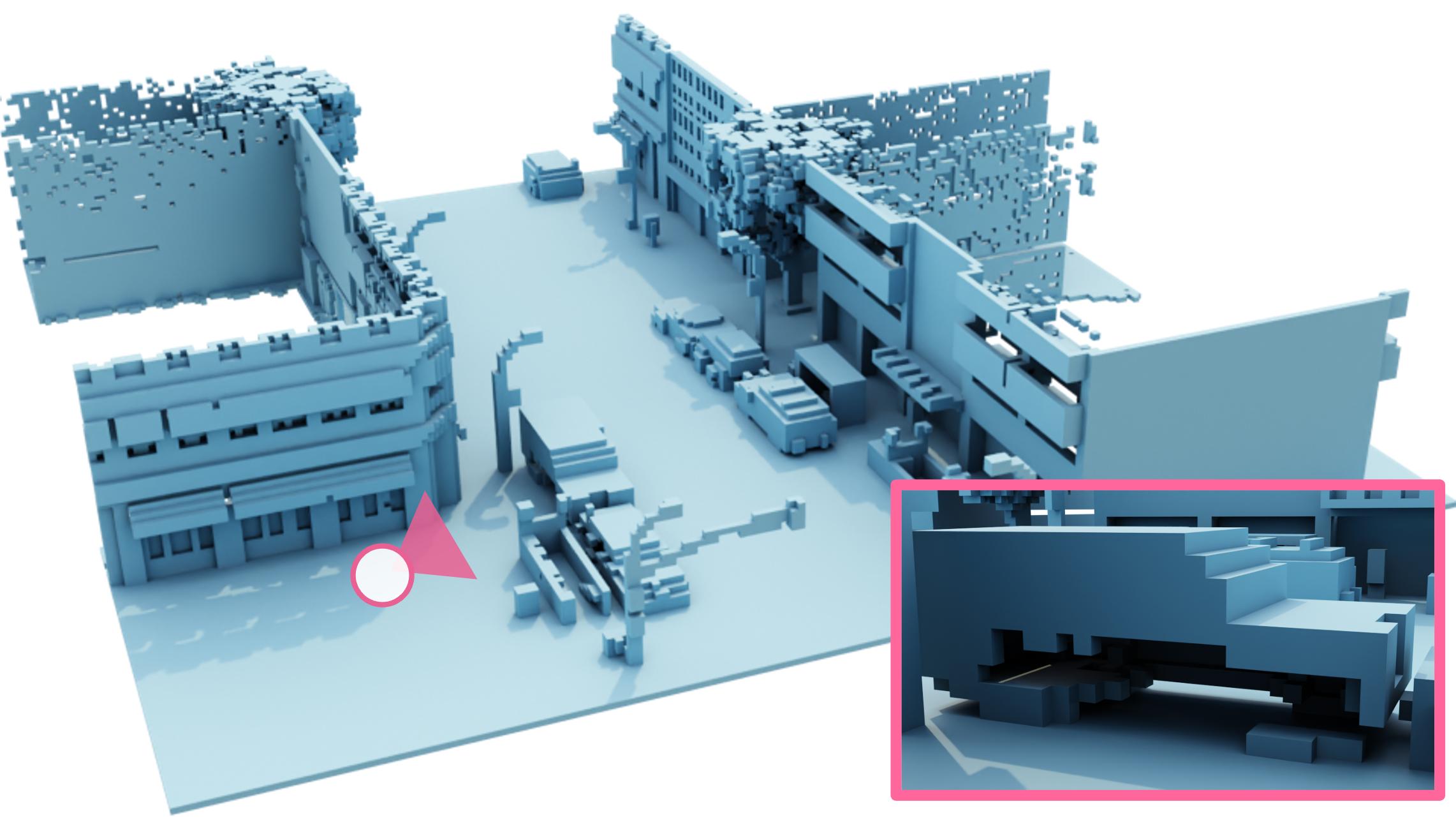} &
        \includegraphics[width=0.3\textwidth]{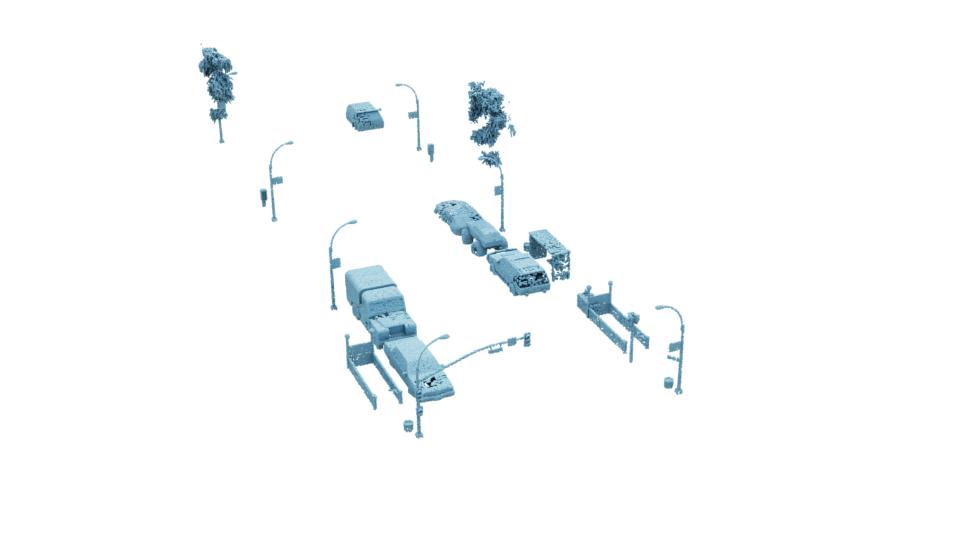} \\
        (d) IoU Mask & (e) Regions for Street CD  \\[6pt]
    \end{tabular}
    \caption{ 
        Evaluation visualization.
        (a) Ground truth geometry in Karton City. (b) Input scans. (c) High LiDAR Resim of GT from a novel pose. (d). Ground truth occupied regions for IoU evaluation. (e) Regions for evaluating street CD.
        High LiDAR ReSim and IoU captures geometry above the input LiDAR range, while it does not capture inconsistent building interiors (green).
        Street CD is the only metric that can evaluate completion of occluded geometry from road, such as side-walk side of the car.
    }
    \label{fig:eval}
\end{figure*}

\section{Evaluation Metric} \label{app:eval_metric}

\textbf{High LiDAR ReSim}\label{app:high_lidar_resim} evaluates the fidelity of the completion beyond the LiDAR range focusing on regions visible from the street, visualized in Fig.~\ref{fig:eval}.
It computes the Chamfer distance between a ground truth LiDAR scan and a re-simulated LiDAR scan from a pose distant from the center after the completion.
The metric avoided evaluating inconsistent geometry in interior walls of the buildings in ground truth geometry (green in Fig.~\ref{fig:eval}).

Given an origin in the ego-vehicle frame, we set a region of interest $R$ to be a box of 38.4 $\times$ 38.4 meters.
We take the scanned input point cloud $X$ within $R$ and generate completion $Y$.
We select two poses $p_1, p_2 \in P$ from the ego vehicle trajectory that enters and leaves the region of interest $R$.
The selected poses are distant from the center, which and inside $R$, making the re-simulated LiDAR scan far from the input scan taken from the center while making the LiDAR ReSim free of occlusions occurring outside of $R$.
We found only 3 out of 1440 (180 scenes $\times$ 2 poses per scene $\times$ 4 test configurations with Karton City/CARLA dataset with 5/10 input scans) poses overlapped with the input pose, indicating that the selected poses for evaluation are mostly unique.
We perform lidar simulation from poses $p_t$ one the completion $Y$ in mesh representation.
If the completion is in voxel representation, we convert it into mesh using marching cubes~\cite{lorensen1987mcubes} and average the Chamfer Distance between the GT and re-simulated LiDAR scan.
For high range LiDAR used from evaluation, we use a 128-beam LiDAR with elevation angle (-30\textdegree, 30\textdegree), which acquires denser scans than the input captured with 64-beam LiDAR, to cover the range above the input scan, visualized in (c) of Fig~\ref{fig:eval}.
Thus the Chamfer distance (CD) for input $X$ is defined by
\begin{equation*}
    \text{CD}_{ReSim}(X, Y)=\frac{1}{|P|} \Sigma_{p_t \in P}{\text{CD}(X_t, Y_t)},
\end{equation*}
where $X_t$ is the lidar scan from pose $p_t$ in the region of interest and $Y_t$ is the simulated lidar scan of generation $Y$ from pose $p_t$.
We also evaluate diversity TMD as in~\cite{wu_2020_ECCV, zhang2021learning, zhang2022probabilistic} for input $X$ is defined as 
\begin{multline*}
    TMD_{ReSim}(X, Y) = \\ \frac{1}{|P|} \Sigma_{p_t \in P}{TMD(\{X_t\}, \cup_{1 \le k \le K} \{Y_{t, k}\})},
\end{multline*}
where $Y_{t, k}$ is the $k$-th completion for input $X_t$ and TMD is defined as the following:
\begin{multline*}
    \text{TMD}(S_p, S_c) = \\ 
        \frac{1}{|S_p|}
        \sum_{P \in S_c}{
            \frac{2}{k(k - 1)} \sum_{1 \leq i < k}{\sum_{i < j \leq k}}{\text{CD}(C^P_i, C^P_j)},
        }
\end{multline*}
where $P \in S_p$ denotes the partial input and $S_c = C^P_{1:k}$ is the set of completions $c^P_i$ for partial input $P$.

\subsection{IoU}
\textbf{IoU} is evaluated on the visible regions in $\text{20cm}^3$ voxel resolution from the street following the previous semantic scene completion (SSC) works~\cite{SemanticKITTI, JS3CNet, scpnet}.
In contrast to SSC that computes IoU against accumulated LiDAR scans, we compute IoU against ground truth geometry 
from the visible regions are obtained using high elevation LiDAR, used for High LiDAR Resim, and covers regions beyond the input LiDAR range, visualized in (d) of Fig.~\ref{fig:eval}.
To obtain the visibility mask, we perform TSDF fusion~\cite{newcombe2011kinect} from all the poses in a single drive for each block of interest and obtain the TSDF values for the block with grid of voxel resolution $\text{10cm}^3$.
We set the grid to visible only if the TSDF value is bigger than -0.3 and convert the mask into $\text{20cm}^3$ voxel resolution.

\textbf{Street CD} includes evaluation on geometry completely occluded from the ego-trajectory, such as the sidewalk side of parked cars, visualized in Fig.~\ref{fig:eval}, which neither High LiDAR ReSim nor IoU (pink) can evaluate.
On Karton City dataset, where the scene is a simple crossroad junction, we compute Chamfer distance between the generated geometry against GT, only on the objects on the main street.
To evaluate objects above the ground, we remove it for both the completion and ground truth by simply thresholding the z-axis with 20cm.

\section{Additional Visualizations on Synthetic Scenes}
We provide additional visualizations on synthetic scenes in Fig.~\ref{fig:synthetic_1},~\ref{fig:synthetic_5},~\ref{fig:synthetic_10},~\ref{fig:synthetic_all}.
\def \qualwidth{0.23}
\begin{figure*}
        \rotatebox[origin=lB]{90}{\small{\;\;\;\;\;\;\;\;Input}}
        \includegraphics[width=\qualwidth\linewidth]{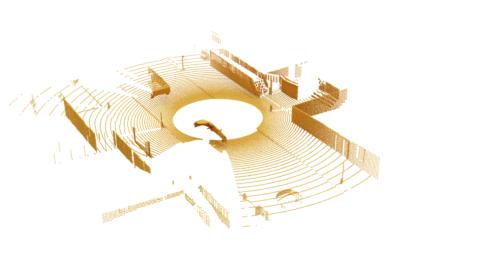}
        \hfill
        \includegraphics[width=\qualwidth\linewidth]{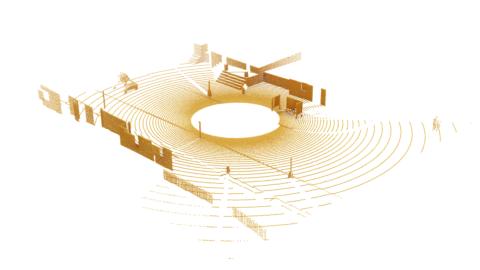}
        \includegraphics[width=\qualwidth\linewidth]{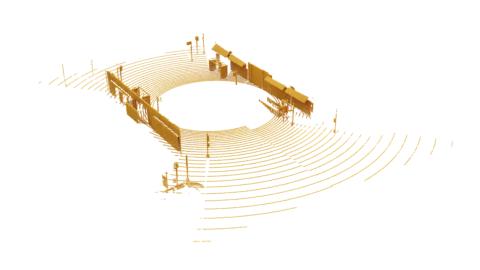}
        \hfill
        \includegraphics[width=\qualwidth\linewidth]{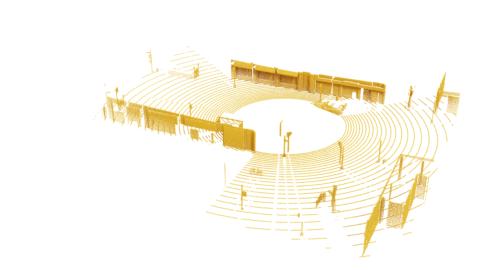}
        \\
        \rotatebox[origin=lB]{90}{\small{\;\;\;\;\;\;\;\;\;\;GT}}
        \includegraphics[width=\qualwidth\linewidth]{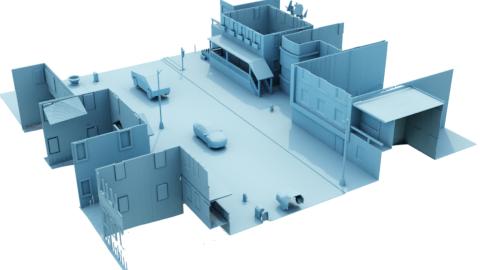}
        \hfill
        \includegraphics[width=\qualwidth\linewidth]{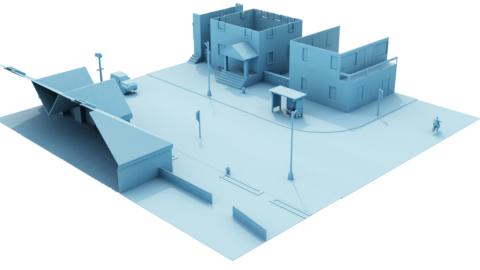}
        \hfill
        \includegraphics[width=\qualwidth\linewidth]{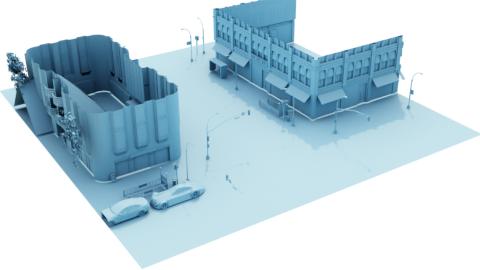}
        \hfill
        \includegraphics[width=\qualwidth\linewidth]{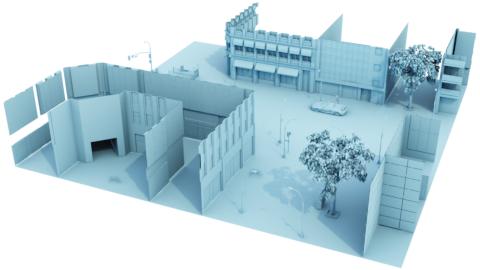}
        \\        
        \rotatebox[origin=lB]{90}{\small{\;\;\;\;\;ConvOcc}}
        \includegraphics[width=\qualwidth\linewidth]{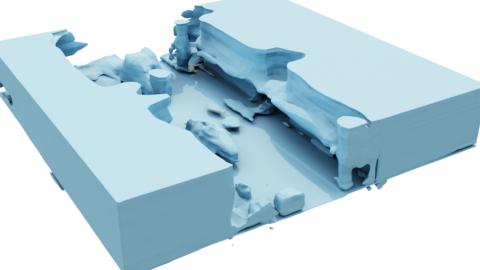}
        \hfill
        \includegraphics[width=\qualwidth\linewidth]{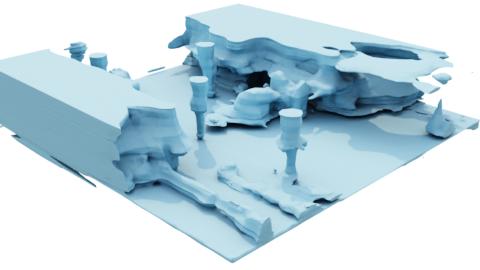}
        \hfill
        \includegraphics[width=\qualwidth\linewidth]{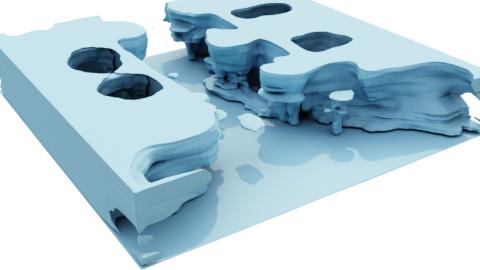}
        \hfill
        \includegraphics[width=\qualwidth\linewidth]{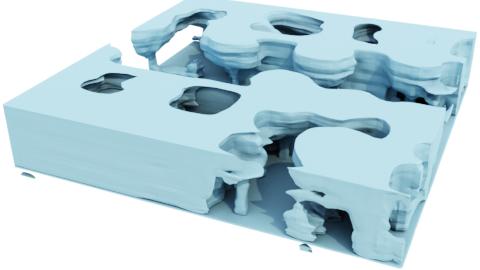}
        \\
        \rotatebox[origin=lB]{90}{\small{\;\;\;\;\;\;\;SCPNet}}
        \includegraphics[width=\qualwidth\linewidth]{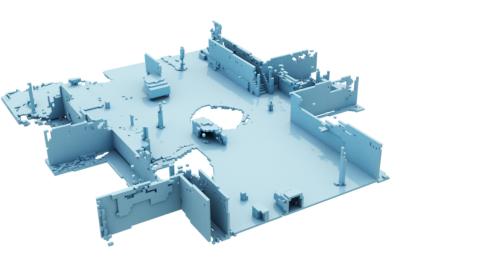}
        \hfill
        \includegraphics[width=\qualwidth\linewidth]{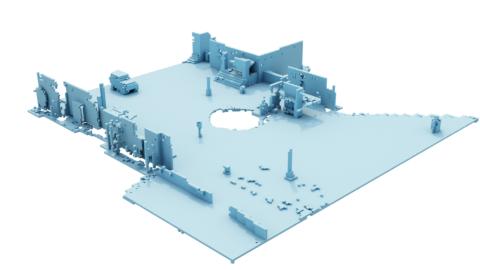}
        \hfill
        \includegraphics[width=\qualwidth\linewidth]{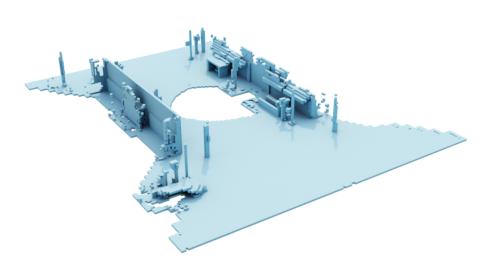}
        \hfill
        \includegraphics[width=\qualwidth\linewidth]{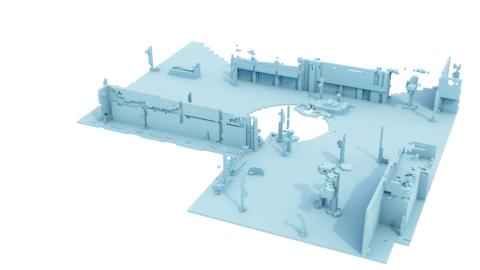}
        \\
        \rotatebox[origin=lB]{90}{\small{\;\;JS3CNet } \scriptsize{($\text{20cm}$)}}
        \includegraphics[width=\qualwidth\linewidth]{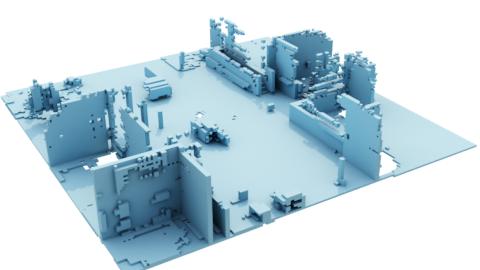}
        \hfill
        \includegraphics[width=\qualwidth\linewidth]{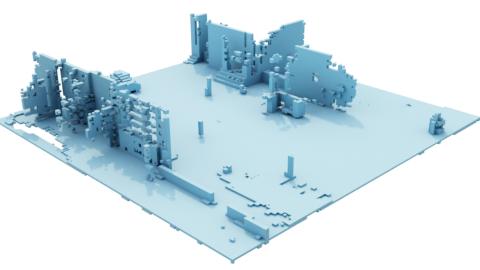}
        \hfill
        \includegraphics[width=\qualwidth\linewidth]{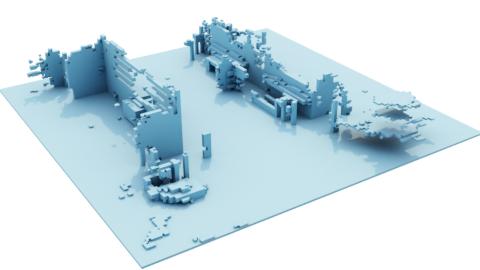}
        \hfill
        \includegraphics[width=\qualwidth\linewidth]{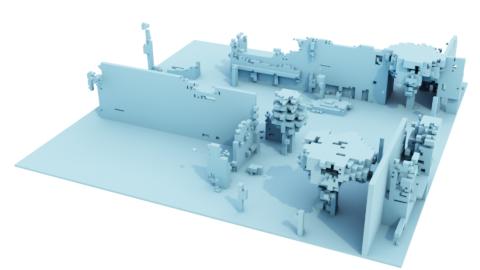}
        \\
        \rotatebox[origin=lB]{90}{\small{\;\;\;\;\;\;\;SG-NN}}
        \includegraphics[width=\qualwidth\linewidth]{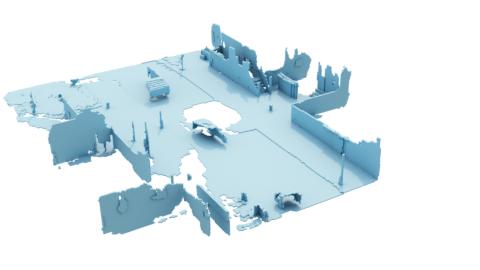}
        \hfill
        \includegraphics[width=\qualwidth\linewidth]{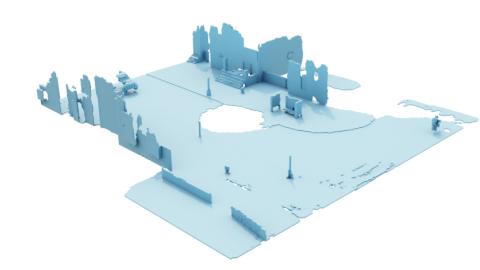}
        \hfill
        \includegraphics[width=\qualwidth\linewidth]{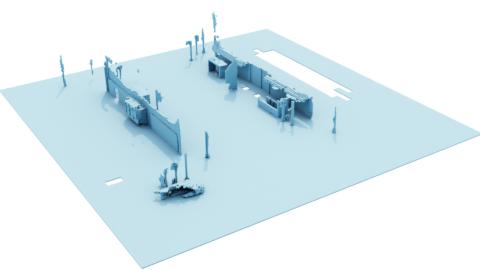}
        \hfill
        \includegraphics[width=\qualwidth\linewidth]{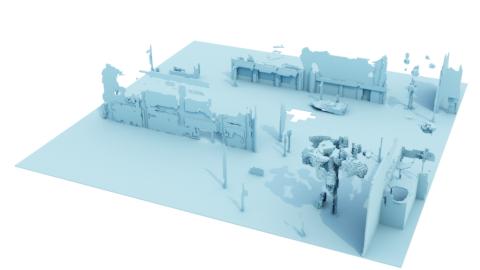}
        \\
        \rotatebox[origin=lB]{90}{\small{\;\;\;\;GCA }\scriptsize{($\text{20cm}$)}}
        \includegraphics[width=\qualwidth\linewidth]{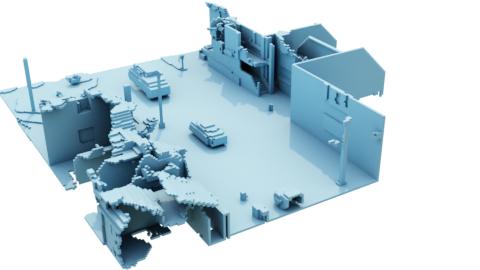}
        \hfill
        \includegraphics[width=\qualwidth\linewidth]{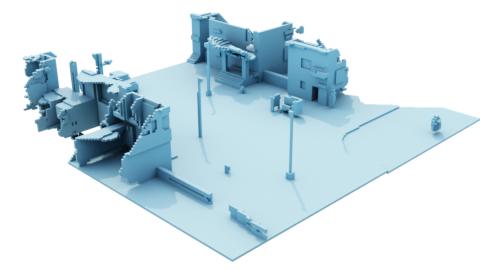}
        \hfill
        \includegraphics[width=\qualwidth\linewidth]{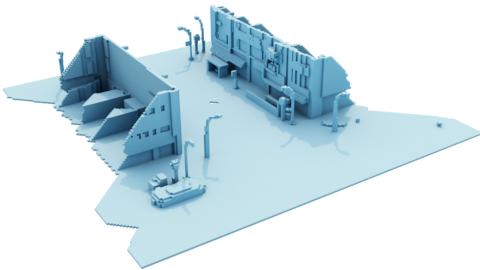}
        \hfill
        \includegraphics[width=\qualwidth\linewidth]{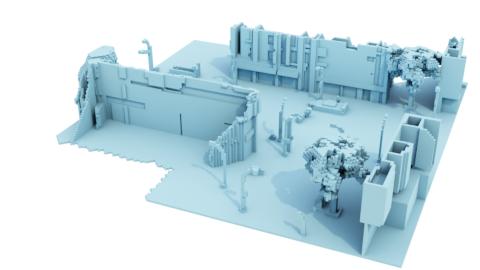}
        \\
        \rotatebox[origin=lB]{90}{\small{\;\;\;\;\;\;\;\;cGCA}}
        \includegraphics[width=\qualwidth\linewidth]{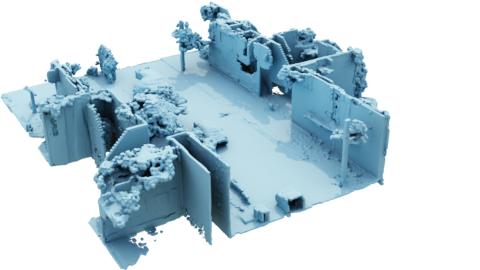}
        \hfill
        \includegraphics[width=\qualwidth\linewidth]{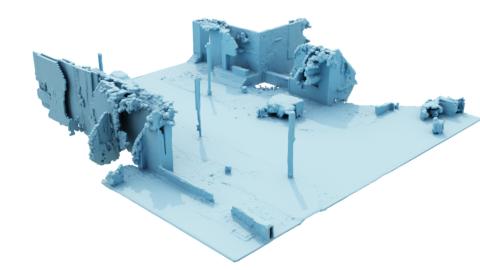}
        \hfill
        \includegraphics[width=\qualwidth\linewidth]{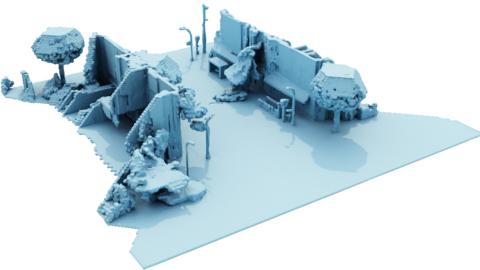}
        \hfill
        \includegraphics[width=\qualwidth\linewidth]{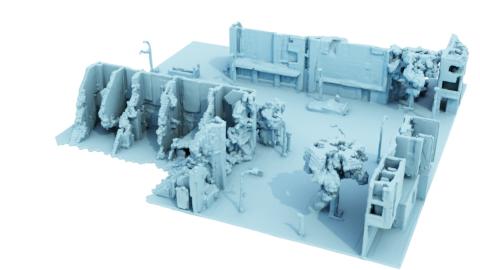}
        \\
        \rotatebox[origin=lB]{90}{\small{\;\;\;\;\;\;\;\;hGCA}}
        \includegraphics[width=\qualwidth\linewidth]{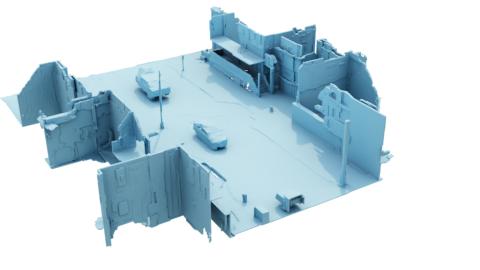}
        \hfill
        \includegraphics[width=\qualwidth\linewidth]{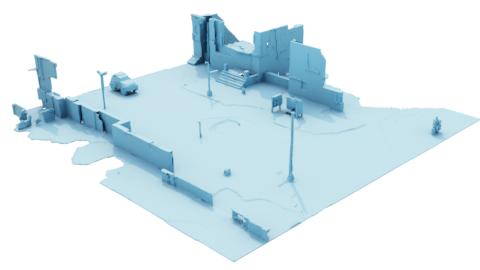}
        \hfill
        \includegraphics[width=\qualwidth\linewidth]{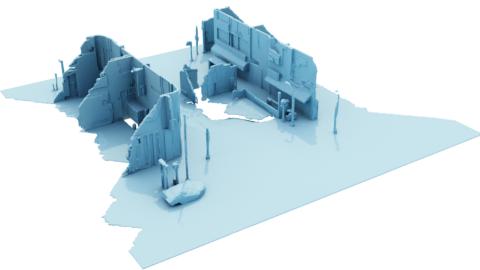}
        \hfill
        \includegraphics[width=\qualwidth\linewidth]{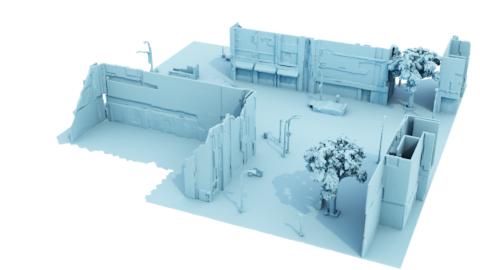}

    \caption{
    Visualizations on CARLA (first 2 columns) and Karton City (last 2 columns) from a single scan. Scenes were randomly chosen.
    }
    \vspace{-1em}
    \label{fig:synthetic_1}
\end{figure*}

\def \qualwidth{0.23}

\begin{figure*}
        \rotatebox[origin=lB]{90}{\small{\;\;\;\;\;\;\;\;Input}}
        \includegraphics[width=\qualwidth\linewidth]{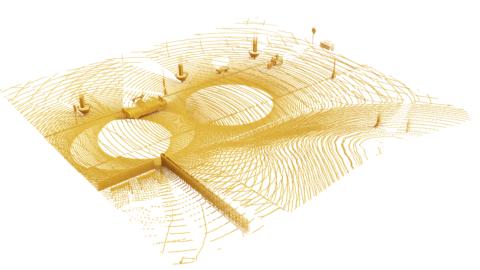}
        \hfill
        \includegraphics[width=\qualwidth\linewidth]{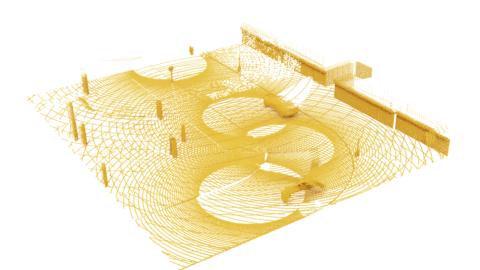}
        \includegraphics[width=\qualwidth\linewidth]{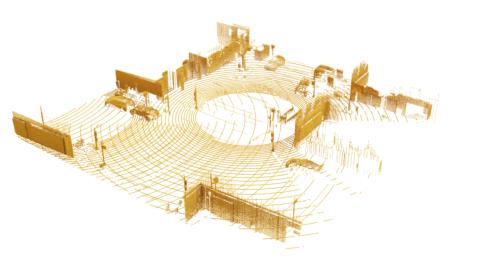}
        \hfill
        \includegraphics[width=\qualwidth\linewidth]{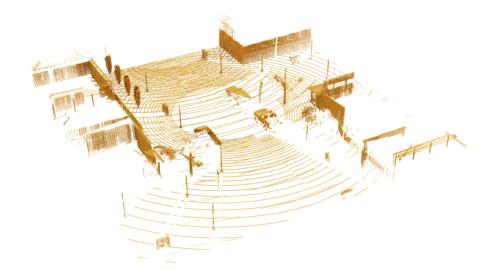}
        \hfill
        \\
        \rotatebox[origin=lB]{90}{\small{\;\;\;\;\;\;\;\;\;\;GT}}
        \includegraphics[width=\qualwidth\linewidth]{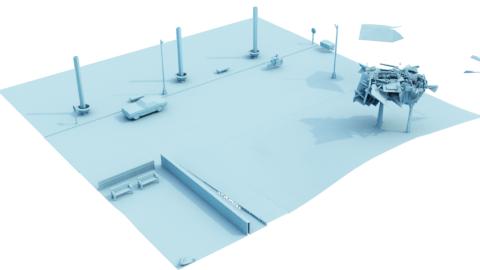}
        \hfill
        \includegraphics[width=\qualwidth\linewidth]{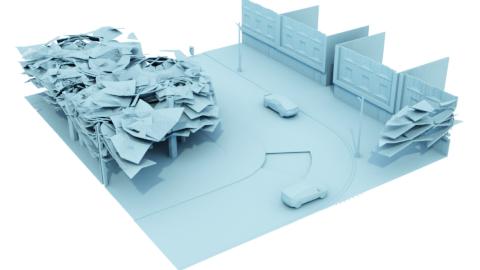}
        \includegraphics[width=\qualwidth\linewidth]{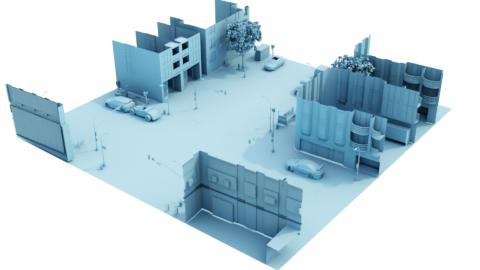}
        \hfill
        \includegraphics[width=\qualwidth\linewidth]{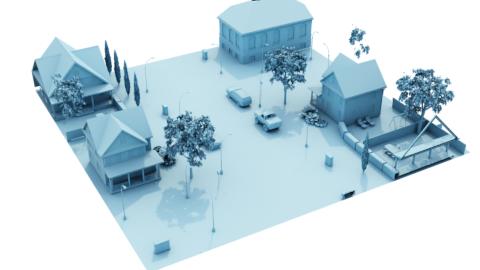}
        \hfill
        \\        
        \rotatebox[origin=lB]{90}{\small{\;\;\;\;\;ConvOcc}}
        \includegraphics[width=\qualwidth\linewidth]{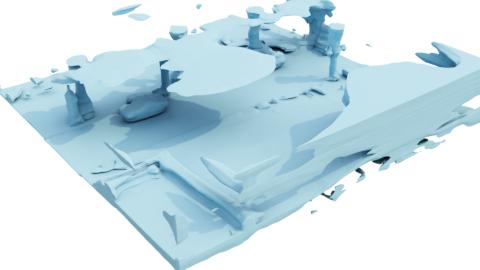}
        \hfill
        \includegraphics[width=\qualwidth\linewidth]{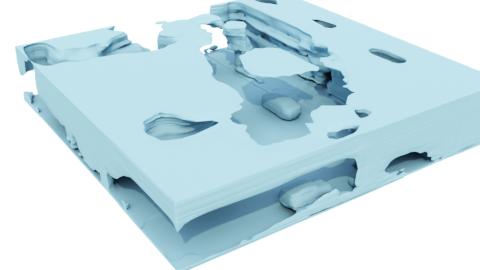}
        \includegraphics[width=\qualwidth\linewidth]{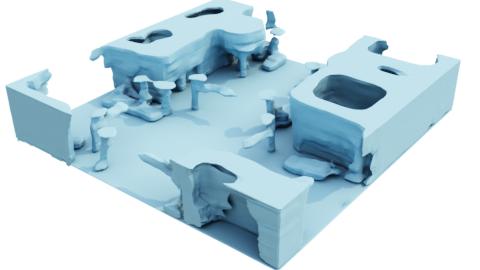}
        \hfill
        \includegraphics[width=\qualwidth\linewidth]{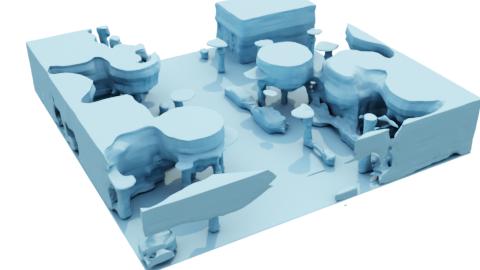}
        \hfill
        \\
        \rotatebox[origin=lB]{90}{\small{\;\;\;\;\;\;\;SCPNet}}
        \includegraphics[width=\qualwidth\linewidth]{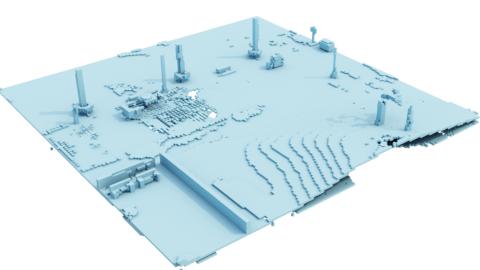}
        \hfill
        \includegraphics[width=\qualwidth\linewidth]{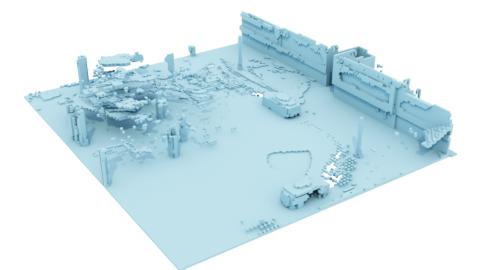}
        \includegraphics[width=\qualwidth\linewidth]{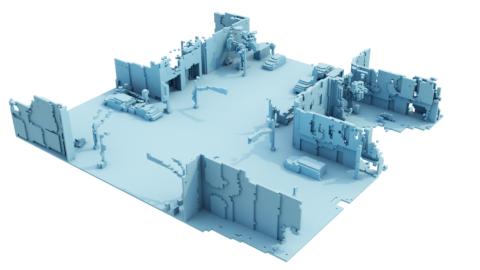}
        \hfill
        \includegraphics[width=\qualwidth\linewidth]{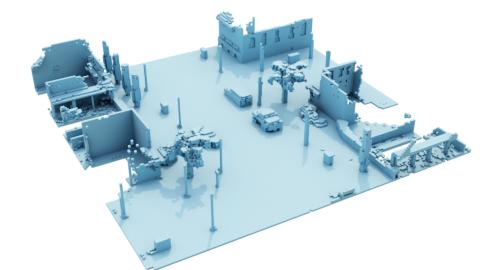}
        \hfill
        \\
        \rotatebox[origin=lB]{90}{\small{\;\;JS3CNet } \scriptsize{($\text{20cm}$)}}
        \includegraphics[width=\qualwidth\linewidth]{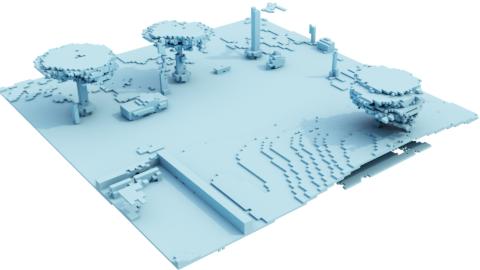}
        \hfill
        \includegraphics[width=\qualwidth\linewidth]{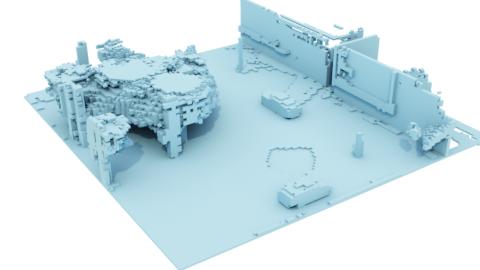}
        \includegraphics[width=\qualwidth\linewidth]{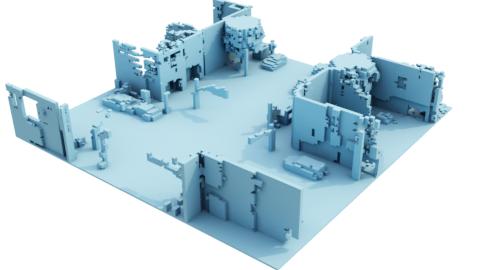}
        \hfill
        \includegraphics[width=\qualwidth\linewidth]{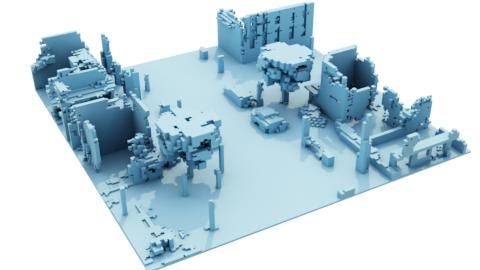}
        \hfill
        \\
        \rotatebox[origin=lB]{90}{\small{\;\;\;\;\;\;\;SG-NN}}
        \includegraphics[width=\qualwidth\linewidth]{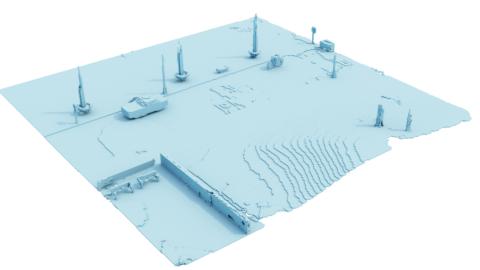}
        \hfill
        \includegraphics[width=\qualwidth\linewidth]{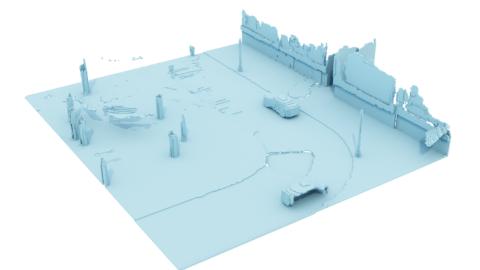}
        \includegraphics[width=\qualwidth\linewidth]{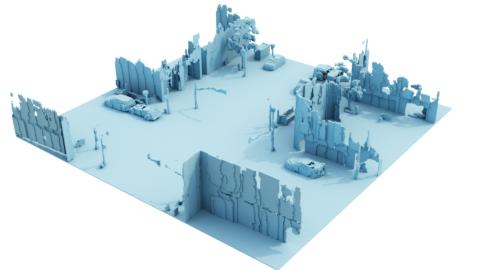}
        \hfill
        \includegraphics[width=\qualwidth\linewidth]{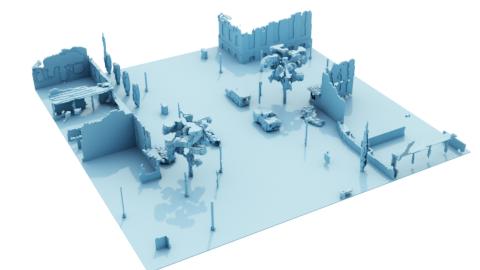}
        \hfill
        \\
        \rotatebox[origin=lB]{90}{\small{\;\;\;\;GCA }\scriptsize{($\text{20cm}$)}}
        \includegraphics[width=\qualwidth\linewidth]{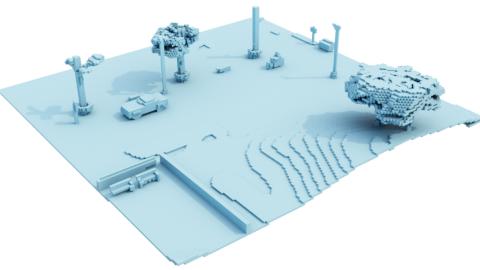}
        \hfill
        \includegraphics[width=\qualwidth\linewidth]{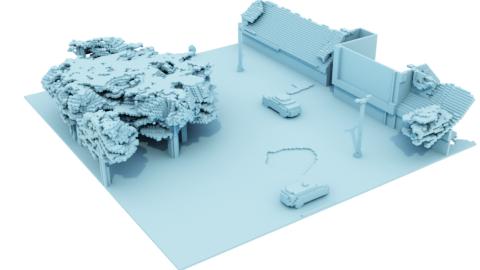}
        \includegraphics[width=\qualwidth\linewidth]{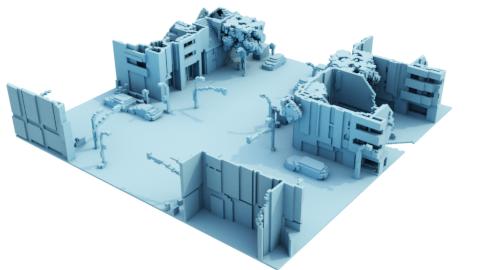}
        \hfill
        \includegraphics[width=\qualwidth\linewidth]{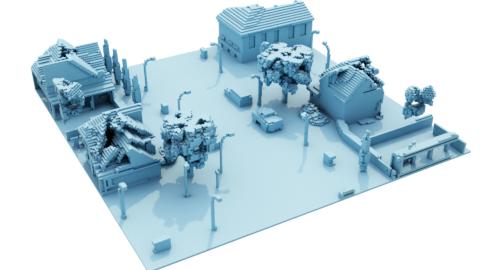}
        \hfill
        \\
        \rotatebox[origin=lB]{90}{\small{\;\;\;\;\;\;\;\;cGCA}}
        \includegraphics[width=\qualwidth\linewidth]{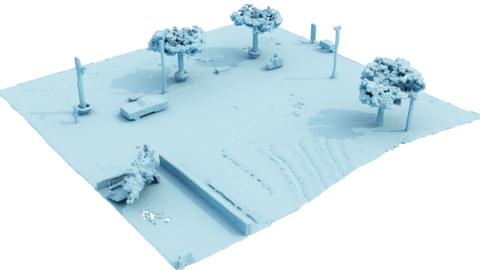}
        \hfill
        \includegraphics[width=\qualwidth\linewidth]{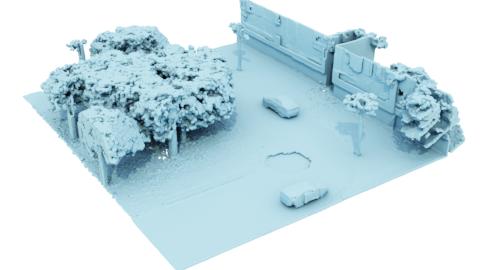}
        \includegraphics[width=\qualwidth\linewidth]{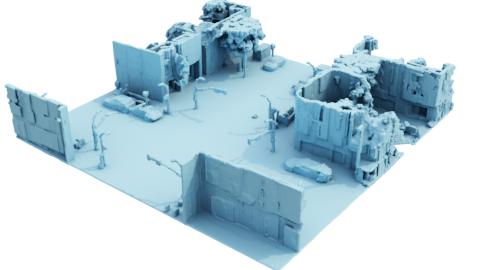}
        \hfill
        \includegraphics[width=\qualwidth\linewidth]{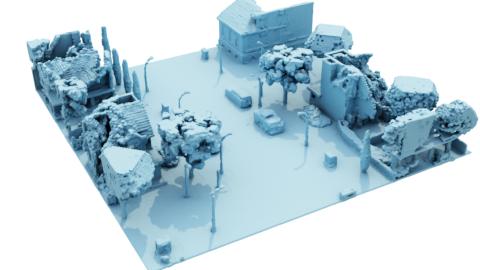}
        \hfill
        \\
        \rotatebox[origin=lB]{90}{\small{\;\;\;\;\;\;\;\;hGCA}}
        \includegraphics[width=\qualwidth\linewidth]{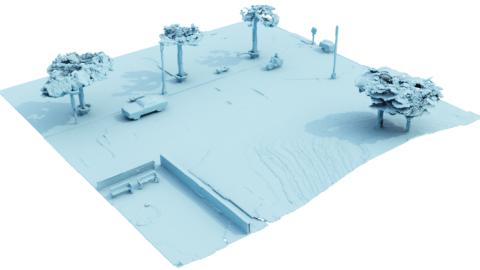}
        \hfill
        \includegraphics[width=\qualwidth\linewidth]{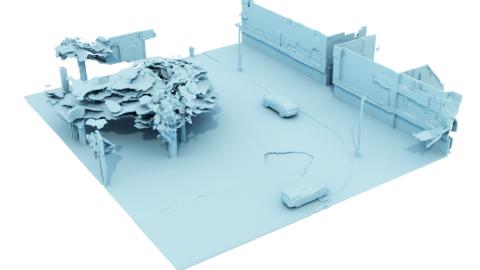}
        \includegraphics[width=\qualwidth\linewidth]{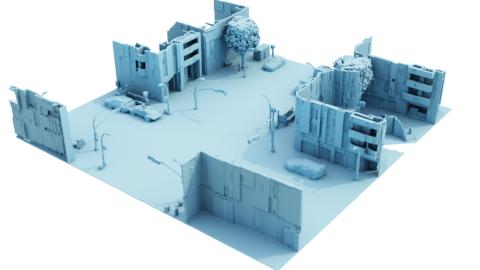}
        \hfill
        \includegraphics[width=\qualwidth\linewidth]{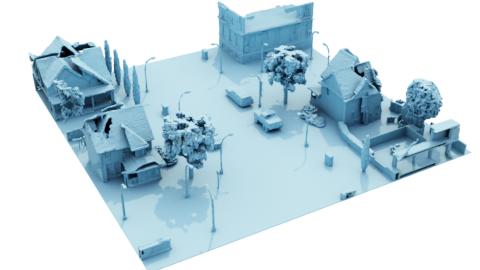}
        \hfill
        \\

    \caption{
    Visualizations on CARLA (first 2 columns) and Karton City (last 2 columns) from 5 scans. Scenes were randomly chosen.
    }
    \vspace{-1em}
    \label{fig:synthetic_5}
\end{figure*}

\def \qualwidth{0.23}
\begin{figure*}
        \rotatebox[origin=lB]{90}{\small{\;\;\;\;\;\;\;\;Input}}
        \includegraphics[width=\qualwidth\linewidth]{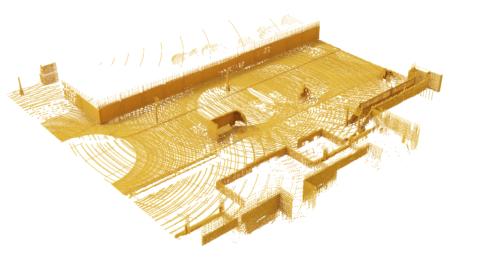}
        \hfill
        \includegraphics[width=\qualwidth\linewidth]{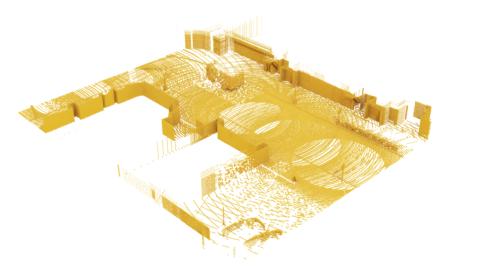}
        \hfill
        \includegraphics[width=\qualwidth\linewidth]{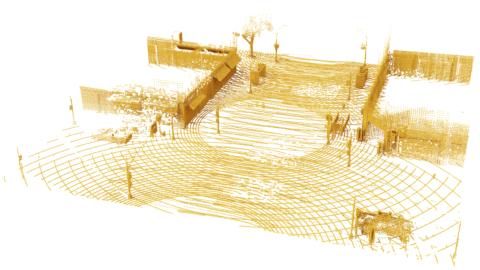}
        \hfill
        \includegraphics[width=\qualwidth\linewidth]{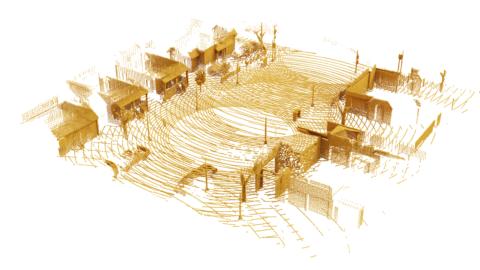}
        \\
        \rotatebox[origin=lB]{90}{\small{\;\;\;\;\;\;\;\;\;\;GT}}
        \includegraphics[width=\qualwidth\linewidth]{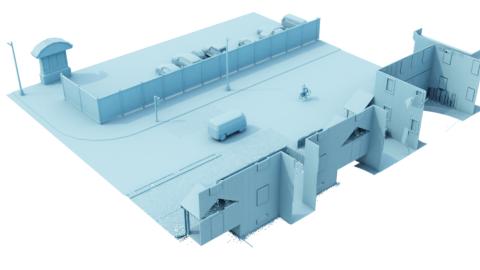}
        \hfill
        \includegraphics[width=\qualwidth\linewidth]{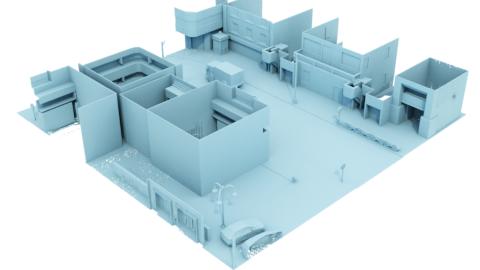}
        \hfill
        \includegraphics[width=\qualwidth\linewidth]{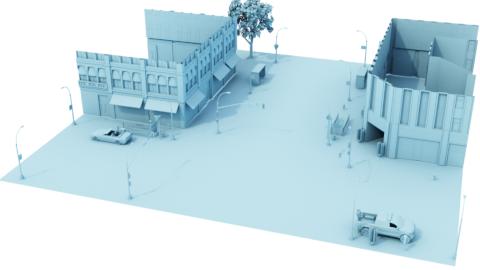}
        \hfill
        \includegraphics[width=\qualwidth\linewidth]{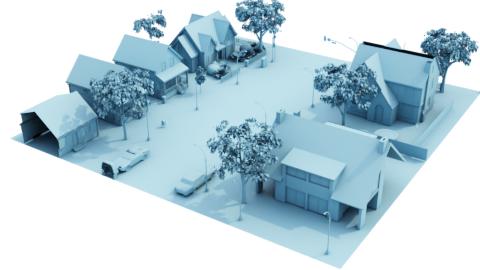}
        \\        
        \rotatebox[origin=lB]{90}{\small{\;\;\;\;\;ConvOcc}}
        \includegraphics[width=\qualwidth\linewidth]{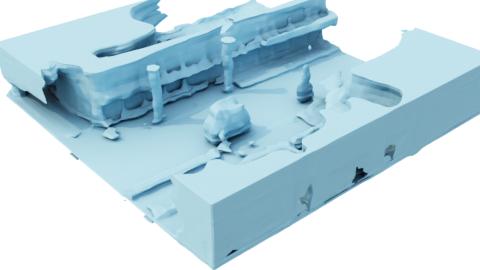}
        \hfill
        \includegraphics[width=\qualwidth\linewidth]{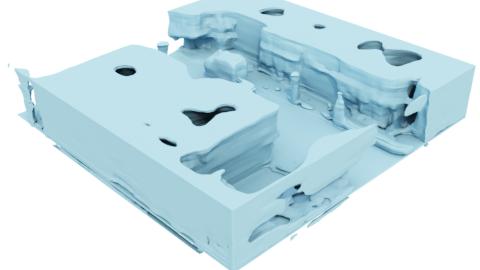}
        \hfill
        \includegraphics[width=\qualwidth\linewidth]{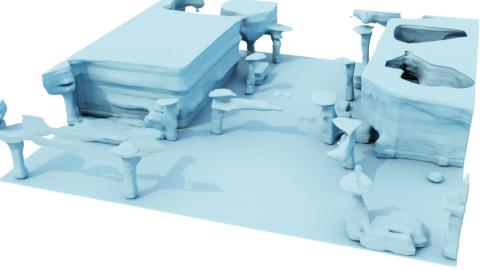}
        \hfill
        \includegraphics[width=\qualwidth\linewidth]{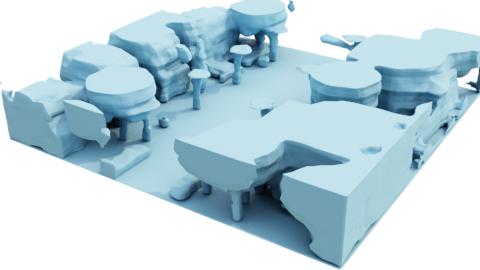}
        \\
        \rotatebox[origin=lB]{90}{\small{\;\;\;\;\;\;\;SCPNet}}
        \includegraphics[width=\qualwidth\linewidth]{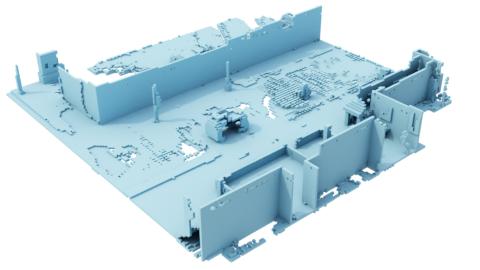}
        \hfill
        \includegraphics[width=\qualwidth\linewidth]{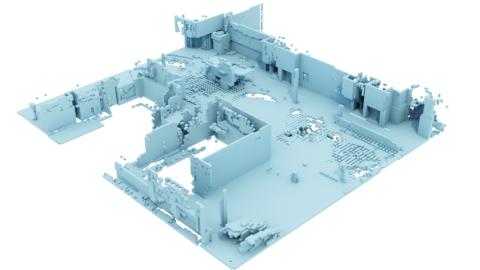}
        \hfill
        \includegraphics[width=\qualwidth\linewidth]{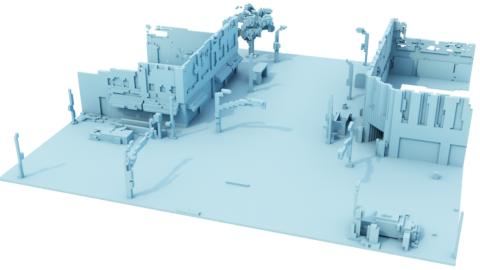}
        \hfill
        \includegraphics[width=\qualwidth\linewidth]{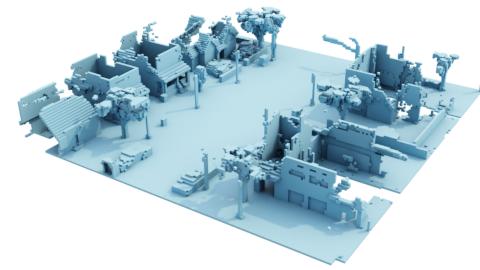}
        \\
        \rotatebox[origin=lB]{90}{\small{\;\;JS3CNet } \scriptsize{($\text{20cm}$)}}
        \includegraphics[width=\qualwidth\linewidth]{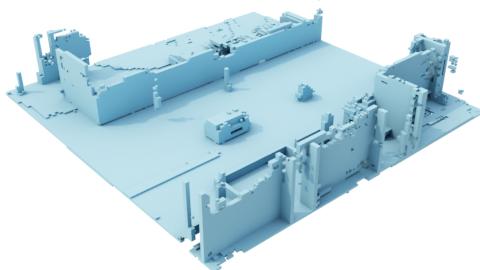}
        \hfill
        \includegraphics[width=\qualwidth\linewidth]{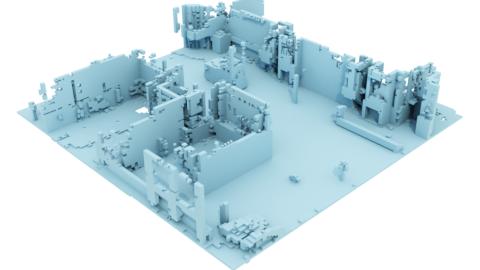}
        \hfill
        \includegraphics[width=\qualwidth\linewidth]{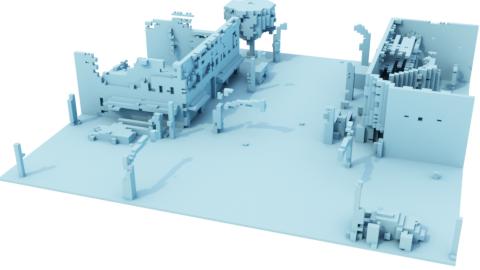}
        \hfill
        \includegraphics[width=\qualwidth\linewidth]{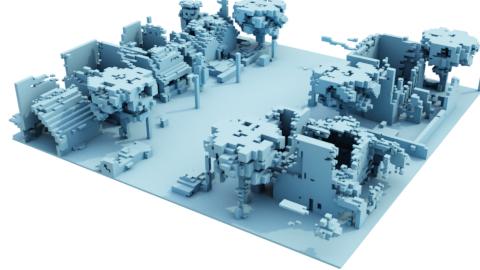}
        \\
        \rotatebox[origin=lB]{90}{\small{\;\;\;\;\;\;\;SG-NN}}
        \includegraphics[width=\qualwidth\linewidth]{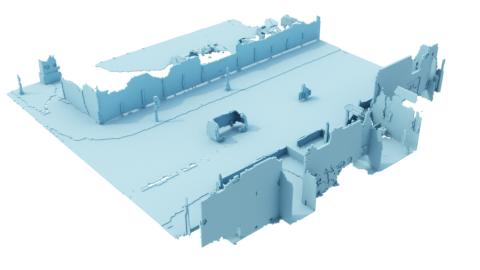}
        \hfill
        \includegraphics[width=\qualwidth\linewidth]{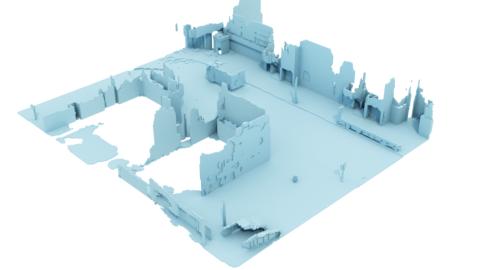}
        \hfill
        \includegraphics[width=\qualwidth\linewidth]{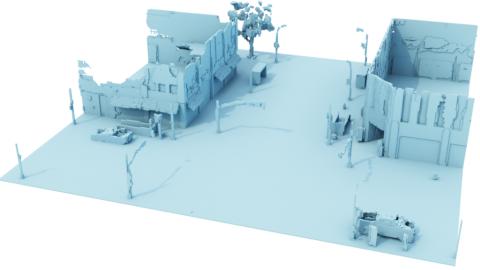}
        \hfill
        \includegraphics[width=\qualwidth\linewidth]{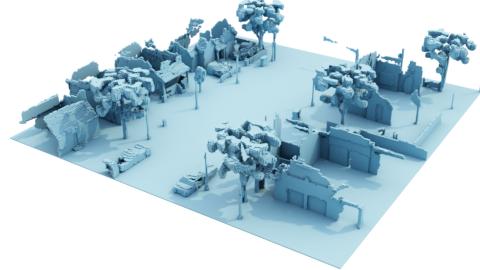}
        \\
        \rotatebox[origin=lB]{90}{\small{\;\;\;\;GCA }\scriptsize{($\text{20cm}$)}}
        \includegraphics[width=\qualwidth\linewidth]{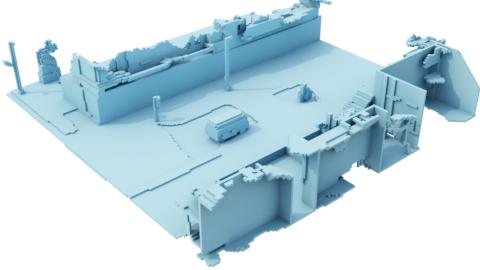}
        \hfill
        \includegraphics[width=\qualwidth\linewidth]{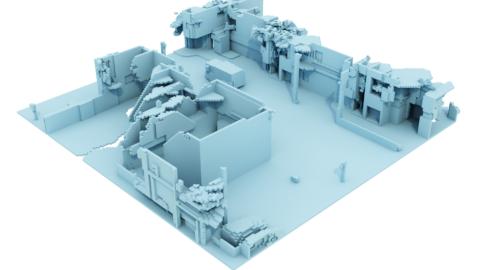}
        \hfill
        \includegraphics[width=\qualwidth\linewidth]{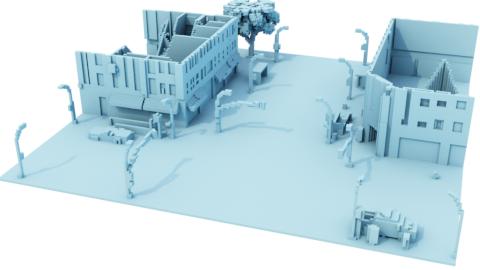}
        \hfill
        \includegraphics[width=\qualwidth\linewidth]{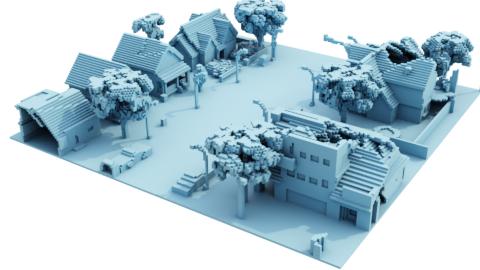}
        \\
        \rotatebox[origin=lB]{90}{\small{\;\;\;\;\;\;\;\;cGCA}}
        \includegraphics[width=\qualwidth\linewidth]{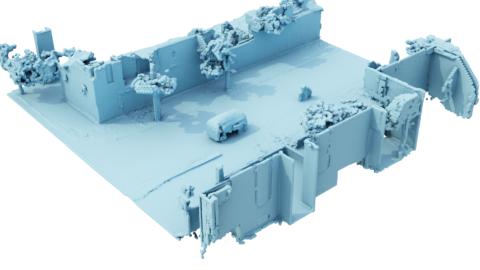}
        \hfill
        \includegraphics[width=\qualwidth\linewidth]{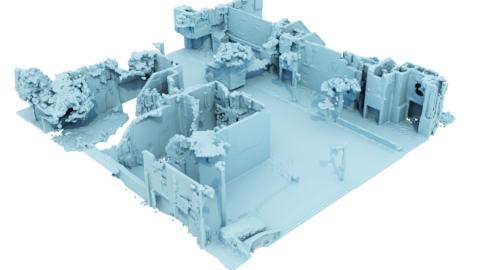}
        \hfill
        \includegraphics[width=\qualwidth\linewidth]{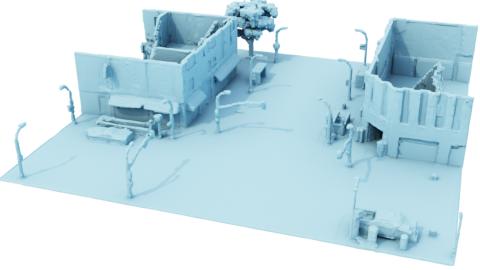}
        \hfill
        \includegraphics[width=\qualwidth\linewidth]{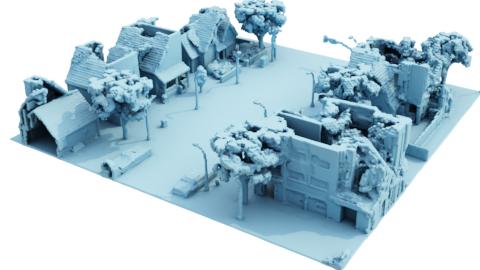}
        \\
        \rotatebox[origin=lB]{90}{\small{\;\;\;\;\;\;\;\;hGCA}}
        \includegraphics[width=\qualwidth\linewidth]{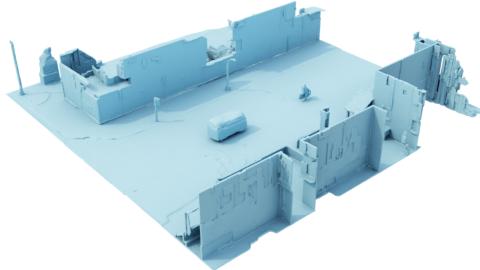}
        \hfill
        \includegraphics[width=\qualwidth\linewidth]{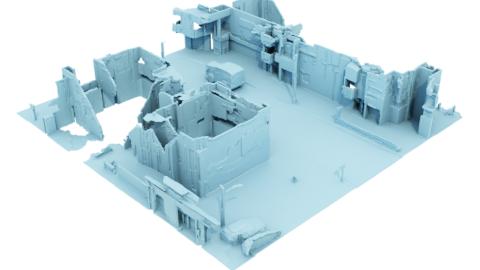}
        \hfill
        \includegraphics[width=\qualwidth\linewidth]{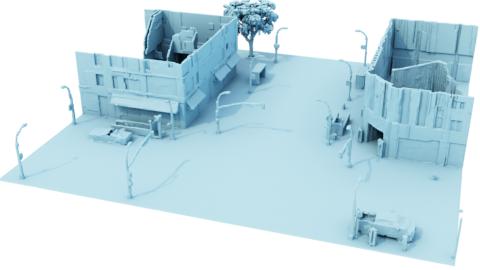}
        \hfill
        \includegraphics[width=\qualwidth\linewidth]{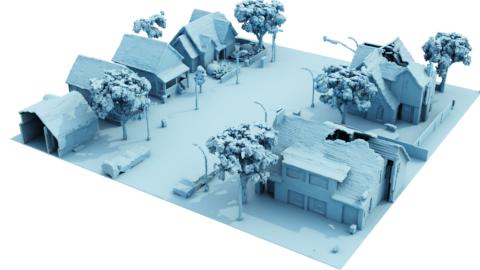}

    \caption{
    Visualizations on CARLA (first 2 columns) and Karton City (last 2 columns) from 10 scans. Scenes were randomly chosen.
    }
    \vspace{-1em}
    \label{fig:synthetic_10}
\end{figure*}

\def \qualwidth{0.23}
\begin{figure*}
        \rotatebox[origin=lB]{90}{\small{\;\;\;\;\;\;\;\;Input}}
        \includegraphics[width=\qualwidth\linewidth]{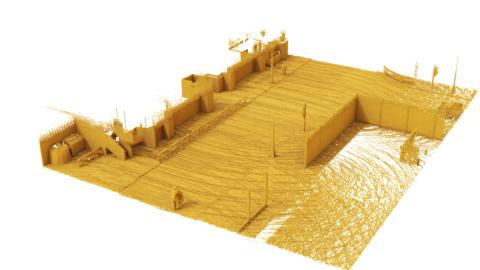}
        \hfill
        \includegraphics[width=\qualwidth\linewidth]{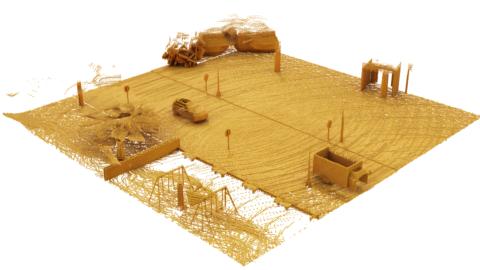}
        \hfill
        \includegraphics[width=\qualwidth\linewidth]{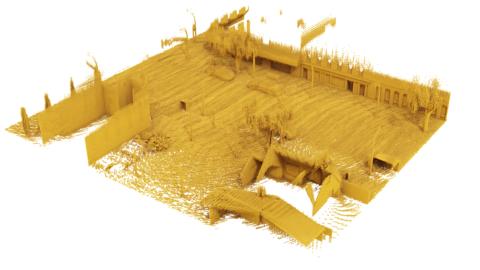}
        \hfill
        \includegraphics[width=\qualwidth\linewidth]{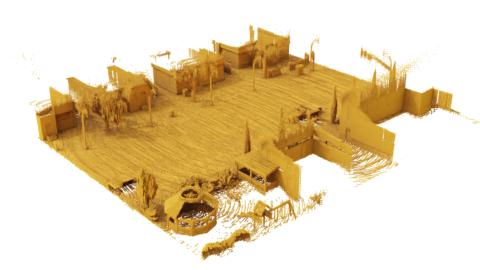}
        \\
        \rotatebox[origin=lB]{90}{\small{\;\;\;\;\;\;\;\;\;\;GT}}
        \includegraphics[width=\qualwidth\linewidth]{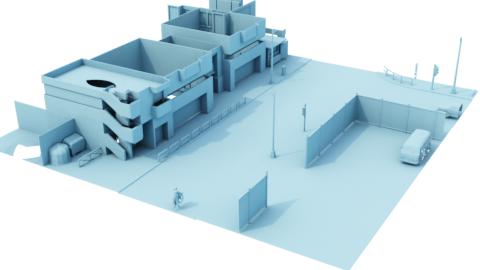}
        \hfill
        \includegraphics[width=\qualwidth\linewidth]{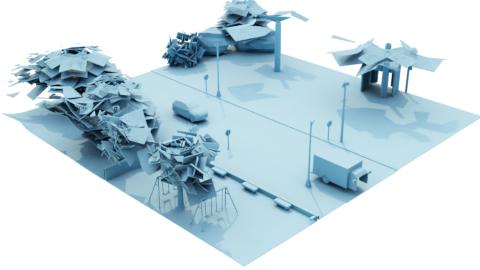}
        \hfill
        \includegraphics[width=\qualwidth\linewidth]{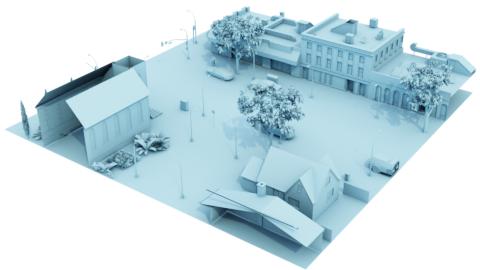}
        \hfill
        \includegraphics[width=\qualwidth\linewidth]{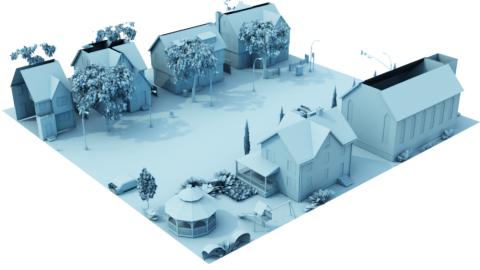}
        \\        
        \rotatebox[origin=lB]{90}{\small{\;\;\;\;\;ConvOcc}}
        \includegraphics[width=\qualwidth\linewidth]{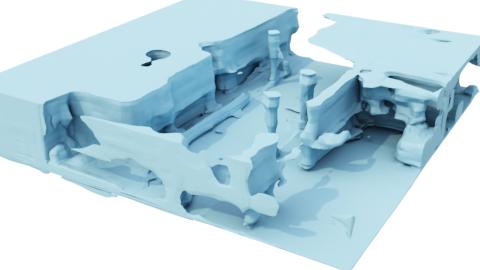}
        \hfill
        \includegraphics[width=\qualwidth\linewidth]{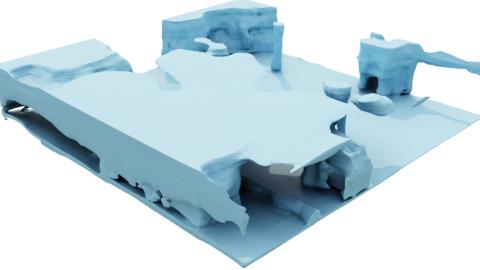}
        \hfill
        \includegraphics[width=\qualwidth\linewidth]{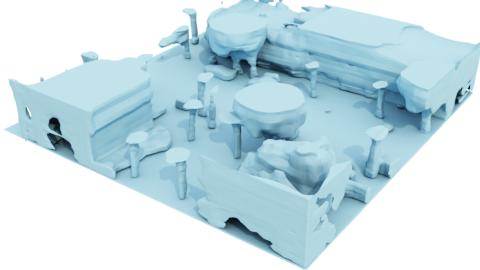}
        \hfill
        \includegraphics[width=\qualwidth\linewidth]{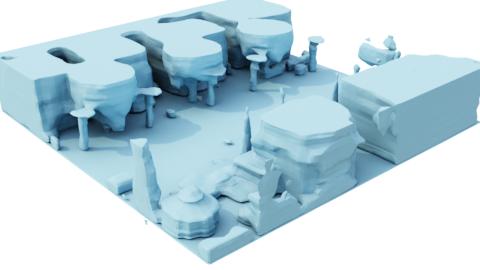}
        \\
        \rotatebox[origin=lB]{90}{\small{\;\;\;\;\;\;\;SCPNet}}
        \includegraphics[width=\qualwidth\linewidth]{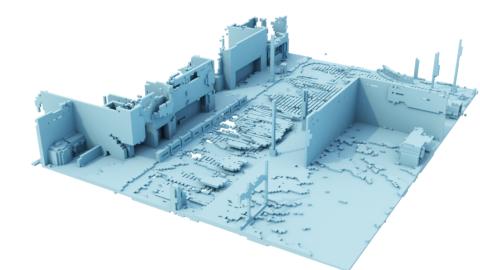}
        \hfill
        \includegraphics[width=\qualwidth\linewidth]{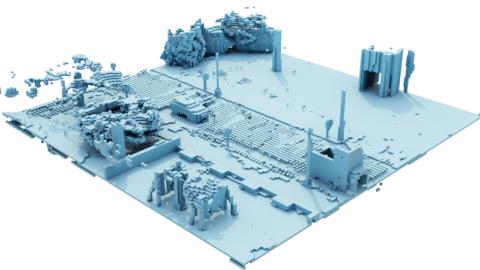}
        \hfill
        \includegraphics[width=\qualwidth\linewidth]{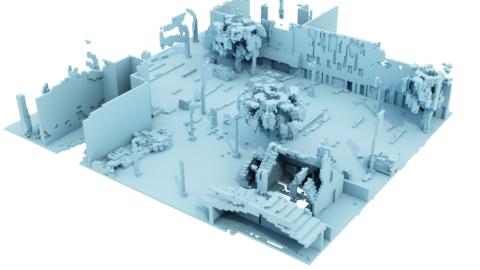}
        \hfill
        \includegraphics[width=\qualwidth\linewidth]{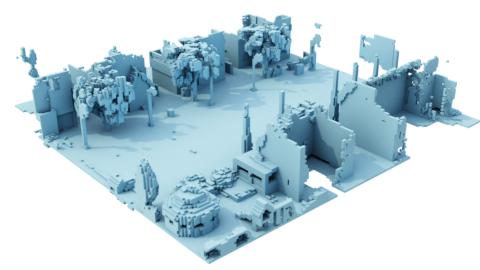}
        \\
        \rotatebox[origin=lB]{90}{\small{\;\;JS3CNet } \scriptsize{($\text{20cm}$)}}
        \includegraphics[width=\qualwidth\linewidth]{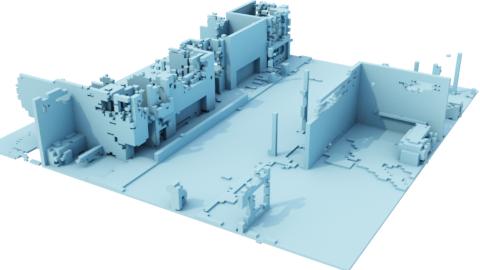}
        \hfill
        \includegraphics[width=\qualwidth\linewidth]{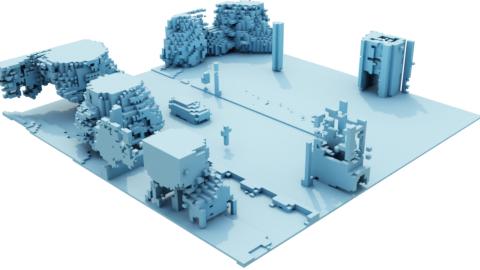}
        \hfill
        \includegraphics[width=\qualwidth\linewidth]{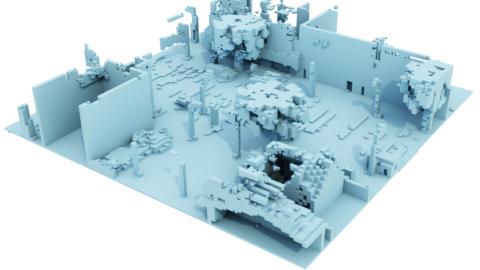}
        \hfill
        \includegraphics[width=\qualwidth\linewidth]{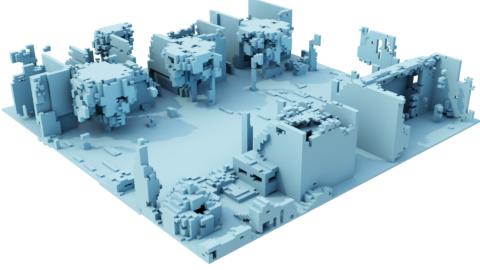}
        \\
        \rotatebox[origin=lB]{90}{\small{\;\;\;\;\;\;\;SG-NN}}
        \includegraphics[width=\qualwidth\linewidth]{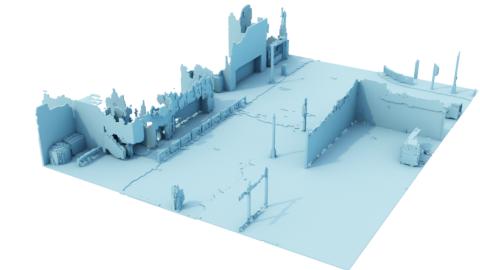}
        \hfill
        \includegraphics[width=\qualwidth\linewidth]{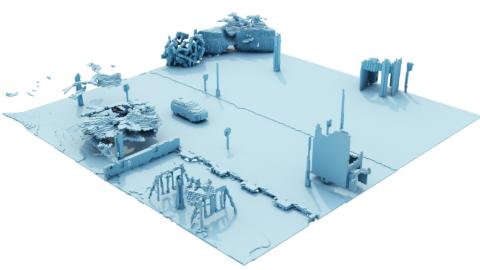}
        \hfill
        \includegraphics[width=\qualwidth\linewidth]{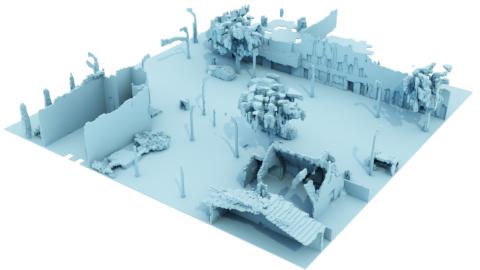}
        \hfill
        \includegraphics[width=\qualwidth\linewidth]{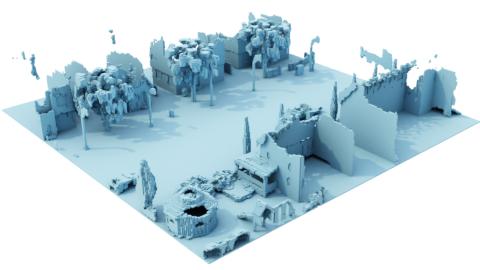}
        \\
        \rotatebox[origin=lB]{90}{\small{\;\;\;\;GCA }\scriptsize{($\text{20cm}$)}}
        \includegraphics[width=\qualwidth\linewidth]{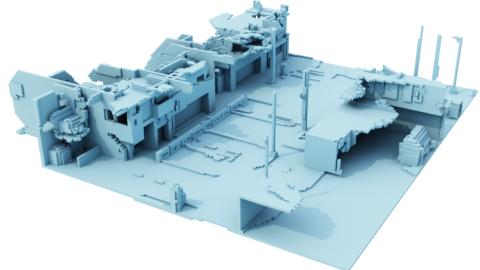}
        \hfill
        \includegraphics[width=\qualwidth\linewidth]{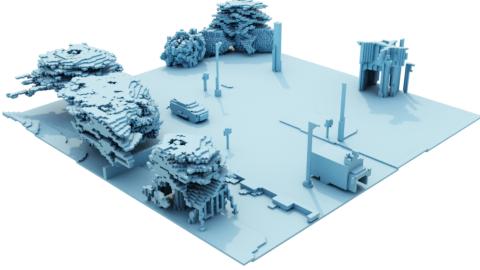}
        \hfill
        \includegraphics[width=\qualwidth\linewidth]{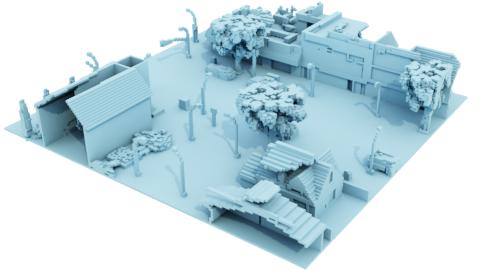}
        \hfill
        \includegraphics[width=\qualwidth\linewidth]{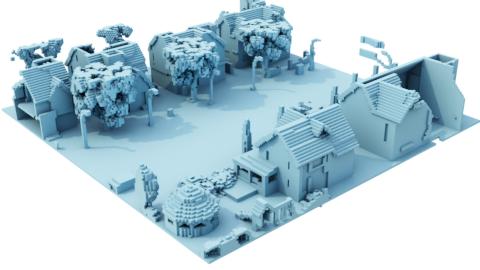}
        \\
        \rotatebox[origin=lB]{90}{\small{\;\;\;\;\;\;\;\;cGCA}}
        \includegraphics[width=\qualwidth\linewidth]{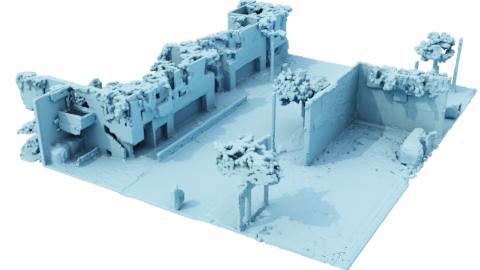}
        \hfill
        \includegraphics[width=\qualwidth\linewidth]{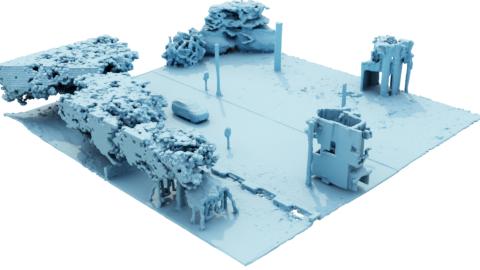}
        \hfill
        \includegraphics[width=\qualwidth\linewidth]{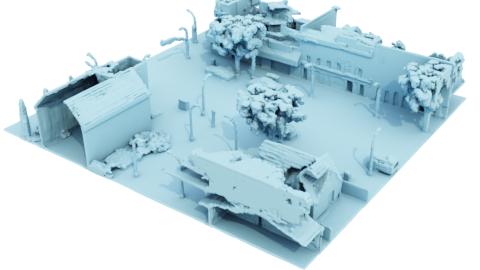}
        \hfill
        \includegraphics[width=\qualwidth\linewidth]{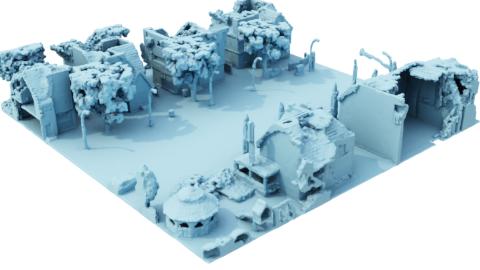}
        \\
        \rotatebox[origin=lB]{90}{\small{\;\;\;\;\;\;\;\;hGCA}}
        \includegraphics[width=\qualwidth\linewidth]{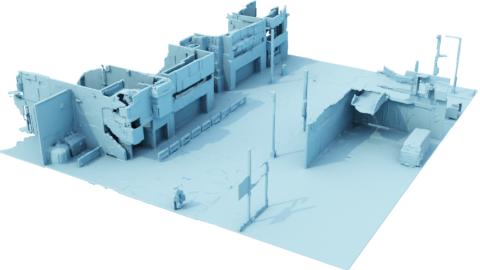}
        \hfill
        \includegraphics[width=\qualwidth\linewidth]{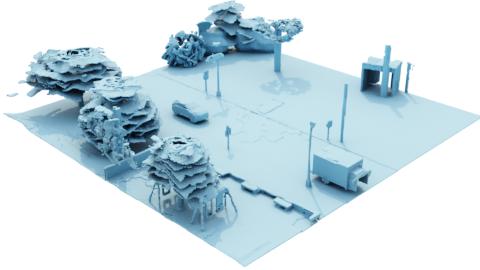}
        \hfill
        \includegraphics[width=\qualwidth\linewidth]{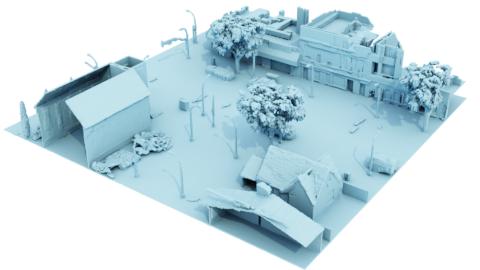}
        \hfill
        \includegraphics[width=\qualwidth\linewidth]{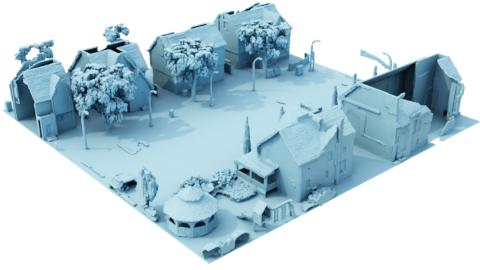}

    \caption{
    Visualizations on CARLA (first 2 columns) and Karton City (last 2 columns) from many accumulated scans. Scenes were randomly chosen.
    }
    \vspace{-1em}
    \label{fig:synthetic_all}
\end{figure*}


\end{document}